# GURU: A Hierarchical Neuro-Symbolic Multi-Agent System for Verification-Aware Automation of Mission-Critical Digital Engineering and Operations

Allan Grosvenor, Anton Zemlyansky, Abdul Wahab, Kyrylo Bohachov,
Aras Dogan, Ryland Adams, Dwyer Deighan
*MSBAI, Los Angeles, California, United States*
*Correspondence: Allan Grosvenor (allan@msb.ai)*[1]

## Abstract

An AI can find a safe action faster than any human, but it must never decide what counts as safe. We present GURU, a hierarchical neuro-symbolic multi-agent system for mission-critical digital engineering and operations, built on one partition of authority: learned models propose, rank, and accelerate; a symbolic rules engine checks every consequential action against hard limits before it executes; and validated physics confirms and decides. GURU runs a workflow as a closed control loop, with Navigation, Action, and Planning agents sharing one Common Internal Language and one training factory. Nuclear, aviation, and national-security guidance now requires this pattern; a harnessed language model keeps the model at the center, GURU puts a validated reference there. In computational fluid dynamics, the workflow raised the converged-simulation rate on unseen aircraft geometries from under 1 percent to 74 percent over 100 single-attempt cases, ran 54 live cases to Mach 6 on three facilities, and an ablation put the symbolic layer at 13 points of boundary-layer capture. Knowledge-graph retrieval and role adapters raised ParaView control success from 55 to 75 percent. In space domain awareness, where a language model cannot be trusted to decide, a frozen self-supervised world model under validated propagation scored highest of five methods at conjunction risk, including fully fine-tuned language models, with several-times-lower seed variance: the trustworthy design gives up nothing in accuracy, and the recipe transferred to reactor anomaly detection. A radar-to-avoidance debris pipeline was demonstrated in stages on a 2026 reentry; the hierarchy coordinates simulated teams of 200 aircraft. Training scaled near-linearly past 1,000 Aurora nodes and ran at 8,000. The approach extends to a control-grade microreactor digital twin with validated physics, not a learned model, as the safety basis.

## 1. Introduction

*An AI can find the safe move faster than any human, in places no human can reach. But it never decides what counts as safe. The validated physics does.*

Artificial intelligence is now asked to operate in settings where a mistake can endanger lives or destroy expensive hardware. Space operators must watch tens of thousands of objects in orbit and notice, within minutes, the few that have maneuvered or begun to behave abnormally. Air traffic managers must keep aircraft clear of launch and reentry hazards while national policy drives the cadence up: the National Space Transportation Policy of August 2026 sets a goal of more than 1,000 launches and reentries a year by 2030 and directs the Department of Transportation to integrate space launch and reentry management into airspace and traffic control modernization efforts (National Security Presidential Memorandum 17, 2026). Aerospace engineers must search large design spaces with high-fidelity simulation before a vehicle is ever built. This is digital engineering: an integrated approach that uses authoritative models and data as a continuum across a system's life cycle (Department of Defense, 2023). Operators of advanced energy systems must keep a plant inside a safe envelope under faults and transients. These tasks share a structure. They are stateful, they are bound to specialized software and instruments, and they are governed by hard physical and procedural constraints.

---

[1] DISTRIBUTION STATEMENT A. Approved for public release: distribution is unlimited. Portions of Section 6 were presented at the 2026 Department of the Air Force Modeling, Simulation and Analytics Summit, Colorado Springs, Colorado, May 6, 2026 (Grosvenor, 2026).

A useful automation for these settings must satisfy four requirements at once. It must be reliable, so that decisions hold under rapidly changing conditions. It must be explainable, so that an operator understands why a recommendation was made. It must be adaptable, so that it generalizes to new environments without manual retuning. And it must be scalable, so that the same system runs from a laptop to a leadership-class supercomputer. Neither a purely symbolic pipeline nor a purely data-driven model meets all four. Symbolic systems give formal guarantees but become brittle in open worlds. Deep networks excel at perception and pattern recognition but act as opaque components whose failures are hard to anticipate. A vision system trained on summer imagery can behave differently once the leaves change color, not because anything that matters has changed but because the cues it learned were never the ones that mattered (Geirhos et al., 2020).

A single large language model does not close this gap either, even when it is wrapped in a capable harness of tools, memory, and retries (Section 2.3). An agent that configures a hypersonic simulation, screens an orbital catalog for collisions, or helps operate a reactor must do more than retrieve and recombine text. It must hold a model of the system it acts on, predict what its actions will do, and plan. Analyses of language-model planning reach the same conclusion from the planning side: these models are often unreliable as standalone planners where correctness must be guaranteed, and they are most effective as one component inside a structured framework that checks their proposals (Kambhampati et al., 2024).

**The requirement bar.** The bar for trustworthy AI in these domains is no longer a matter of opinion, because several independent authorities have now written it down and converged on the same requirements. In the nuclear domain, the Nuclear Regulatory Commission's 2024 gap assessment expects physics-based models in safety-significant roles and names the failure modes that matter, including distributional shift, opacity, rare inputs, and the absence of a test oracle, while its proposed 2026 Part 57 microreactor rule anticipates autonomous operation, and June 2026 national-security guidance requires fielded AI to be reliable, robust, steerable, and controllable under rigorous test, evaluation, validation, and verification (U.S. Nuclear Regulatory Commission, 2024, 2026; National Security Presidential Memorandum 11, 2026).

The same bar has been set independently in civil aviation and in space. Civil-aviation learning-assurance frameworks (European Union Aviation Safety Agency, 2024, 2026; Federal Aviation Administration, 2024), the joint process standard for airborne AI (SAE International & EUROCAE, n.d.), and the space run-time-assurance literature (Schierman et al., 2020; Dunlap et al., 2024; Hobbs et al., 2025) all place a learned component behind an independently designed monitor, bound it against hard constraints, and keep a human on the loop. The operator of the national airspace has now applied the same condition to its own first use of artificial intelligence: the airspace management system the Federal Aviation Administration fielded in September 2026 is described as enhancing existing air traffic management systems while air traffic controllers remain responsible for keeping aircraft safely separated (Federal Aviation Administration, 2026b).

Across nuclear, aviation, and space, these sources differ in status: a regulatory gap assessment, a proposed rule, a presidential policy memorandum, regulator concept papers and roadmaps, a process standard in development, and the research literature on runtime assurance. They are not one certification regime, but they converge on one assurance pattern with five requirements for AI in safety-critical operation: (R1) a validated reference that decides, (R2) an independent runtime check on every consequential action, (R3) explicit handling of out-of-distribution inputs and uncertainty, (R4) a human who retains override authority, and (R5) a documented lifecycle. A learned model's own confidence is useful evidence under R3, but it cannot satisfy R1 or R2: on conditions it has never seen it can be confidently wrong, and it shares every blind spot of the model it is judging. In our view, large language models therefore have no place automating consequential safety decisions, although they can help people state intent and review evidence. The question this paper answers is therefore concrete: can an autonomous system operate mission-critical physical infrastructure while its own self-assessment is never the final word on safety or physical validity?

**This paper presents GURU,** a hierarchical neuro-symbolic multi-agent system built by MSBAI for exactly these conditions (Grosvenor et al., 2022; Grosvenor, Zemlyansky, et al., 2025; Grosvenor, Wahab, et al., 2025). The one-sentence contribution is this: GURU is a hierarchical neuro-symbolic multi-agent system in which learned models propose, rank, compress, and accelerate, an external symbolic layer checks every consequential action against hard physical limits before it can reach a tool or an actuator, and validated solvers confirm and decide. GURU treats a workflow as a closed control loop, in which the system observes, predicts, acts, checks, and revises, rather than as a single prompt and response. Data from sensors, simulations, files, and software interfaces enters a hierarchy of agents and is turned into structured representations and learned embeddings. Low-level Navigation agents stay in contact with the environment. Action agents turn state into bounded, constraint-aware decisions. Planning agents reason over goals, predict the future with learned world models, and revise plans when observations diverge from prediction. A symbolic verification layer runs alongside the whole hierarchy and checks every consequential action against explicit constraints before it is allowed to execute.

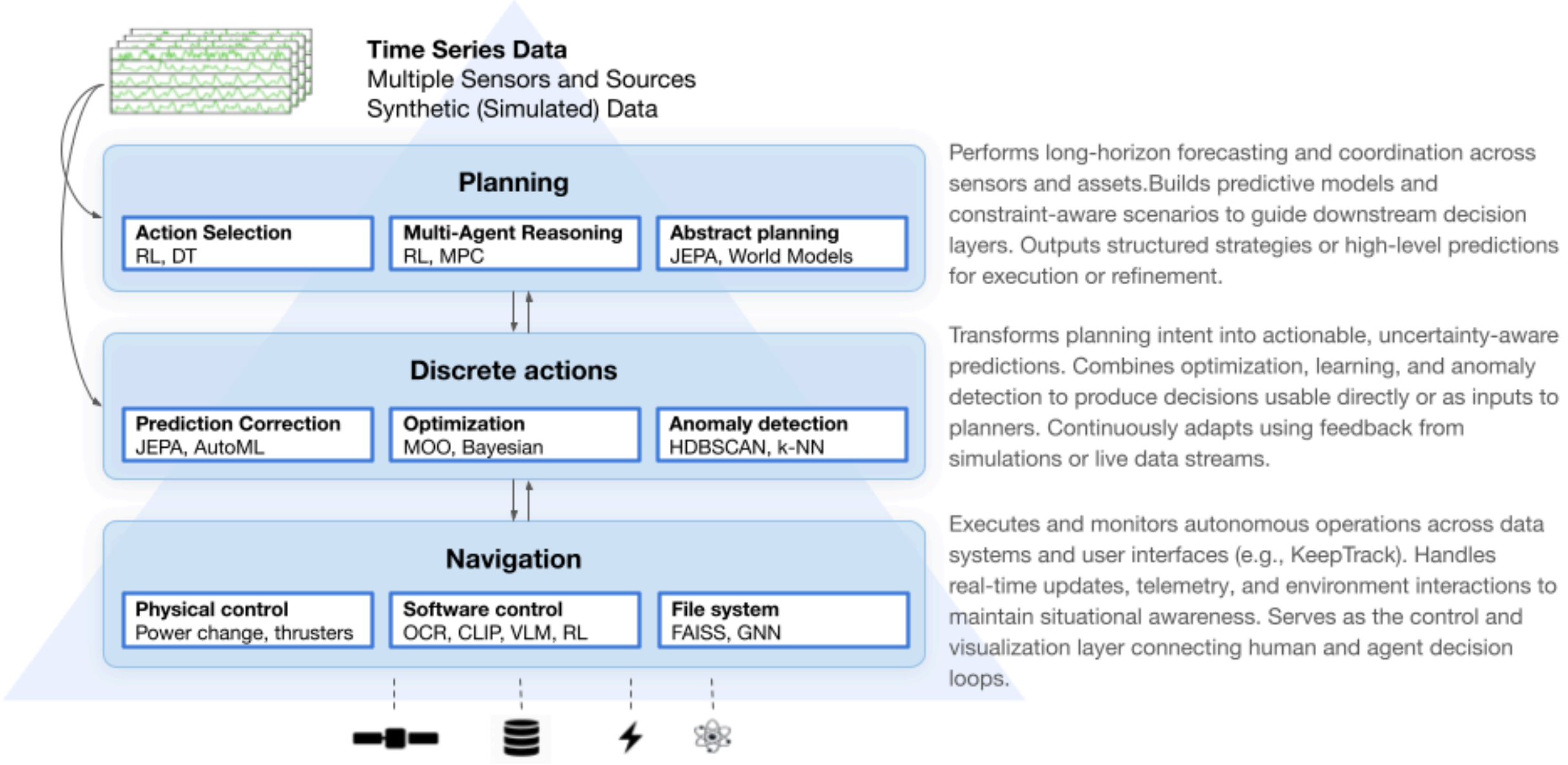

***Figure 1:*** *The GURU hierarchy. The Planning layer performs long-horizon forecasting and planning with JEPA world models and model predictive control. The Action layer turns plans into bounded decisions through optimization, parameter prediction, and anomaly detection. The Navigation layer executes and monitors operations across software and physical systems. Feedback flows bidirectionally, so plans constrain low-level action while observations update the planner. The middle layer, labeled Discrete actions in the figure, is the Action layer.*

The system is built so that one platform and one training factory serve many applications. New capabilities are added by collecting or generating data, training and selecting a model for a specialized agent, validating the agent, and registering it, without disturbing the agents that already work. Applications therefore differ in their data, agents, ontologies, and software wrappers, while the platform stays fixed. This is what lets a system proven on aerodynamics and orbital mechanics carry into a new domain with the same assurances.

**The hypothesis and how we test it.** This paper advances and evaluates one systems hypothesis: a hierarchical hybrid-intelligence architecture can provide scalable autonomy for high-consequence engineering and operations if authority is deliberately partitioned, so that learned components provide adaptation, prediction, search, and acceleration; symbolic runtime mechanisms enforce explicit constraints; and a domain-appropriate validated reference retains final authority over consequential actions. We test the hypothesis through deliberately heterogeneous applications rather than repeated variants of one task. The applications span digital engineering and software control, orbital operations, airspace decision support, distributed vehicle control, and reactor engineering, and they differ in physics, software, data, failure modes, and maturity. That heterogeneity is intentional: it tests whether the architectural partition survives changes in domain, not whether one model transfers unchanged. Architectural transfer does not mean model transfer. The learned model, physical solver, rules, and domain ontology may all change. What remains invariant is the division of responsibility: specialized learned components propose or accelerate; explicit symbolic mechanisms constrain; a domain-specific validated reference retains decision authority; provenance is preserved; and the human retains supervisory authority.

The applications are not offered as equally mature certification demonstrations. Computational fluid dynamics and space operations provide the strongest measured end-to-end evidence; software control, airspace decision support, and multi-aircraft autonomy test transfer of the same decomposition and verification discipline under different software, physics, and operational constraints; and reactor control tests whether the architecture remains traceable to the most demanding assurance case while several integrated mechanisms remain research targets. Each application answers three questions. Instantiation: can the same architectural roles be instantiated without changing the platform? Utility: do the learned components materially improve performance, scalability, search, prediction, or operator capability? Assurance: does consequential authority remain outside those learned components, and how completely have R1 to R5 been realized? Across the applications evaluated to date, the evidence is consistent with one central result: learned components are most useful when they provide adaptation, prediction, and search without holding final authority over physical validity.

**Contributions.** This paper makes seven contributions. First, we describe the GURU platform: a three-layer hierarchy of cooperating agents over a symbolic verification layer and a neuro-symbolic Common Internal Language (Figure 1), and we state precisely how authority is divided between learned and validated components. Second, we describe the training factory that builds, verifies, and versions agents, including a provenance-preserving knowledge graph extracted from manuals, tutorials, and autonomous software exploration. Third, we present an autonomous computational fluid dynamics application, from image-derived geometry through meshing, solving, adaptation, and post-processing in ParaView, with measured gains on unseen geometries. Fourth, we present a space domain awareness application built on a Joint Embedding Predictive Architecture (JEPA) world model (LeCun, 2022; Assran et al., 2023), including a method for irregularly sampled orbital prediction and a prediction-error channel, meaning the gap between the world model's forecast and what is then observed, used as an anomaly signal. Fifth, we show transfer of the same platform to airspace and reentry decision support and to supervised control of simulated multi-aircraft teams. Sixth, we report near-linear scaling of training and simulation on the Frontier and Aurora supercomputers. Seventh, we outline how the approach extends to a control-grade microreactor digital twin in which validated physics, not a learned model, remains the safety basis, and we trace every application against requirements R1 to R5 (Section 12).

# 2. Background and Related Work

## 2.1 Hybrid intelligence

Two families of AI have complementary failures. Symbolic systems state their rules explicitly and can be verified against them, but they are brittle once the world departs from what their rules anticipate. Learned, connectionist systems adapt to messy, high-dimensional data, but they do not reliably abide by hard rules, and their failures are hard to anticipate. The complementarity was stated plainly before deep learning: symbolic and connectionist methods each succeed where the other fails, and capable systems need both (Minsky, 1991). Modern neuro-symbolic computing pursues the same goal, keeping the adaptability of learning while recovering the verifiability of symbolic reasoning (Garcez et al., 2019; Garnelo and Shanahan, 2019). Our earlier work frames the same need as hybrid intelligence for reliable automation, with the human as a partner whose judgment the system supports rather than replaces (Grosvenor, Zemlyansky, et al., 2025).

The coordination model GURU uses is older still. The Society of Mind argues that complex behavior emerges from many narrow processes rather than one monolithic controller (Minsky, 1986). Global Workspace Theory describes how a shared workspace broadcasts salient information to specialized processes that then self-activate (Baars, 1988; Shanahan, 2006). GURU implements these ideas directly: agents are narrow and specialized, most information flows through event-driven message pools and a shared blackboard, and a small number of critical transitions use deterministic channels. For constraint checking, the symbolic layer uses a satisfiability-modulo-theories solver in the tradition of Z3 (de Moura and Bjørner, 2008). GURU's specific commitment, developed throughout this paper, is that learned and symbolic components are not blended into a single decision; they are given different authority.

## 2.2 Runtime assurance and the assurance of learned components

The pattern of letting an unverified, high-performance component act only under the supervision of a simpler verified one is the Simplex architecture: a decision module switches control to a verified fallback whenever the system approaches the boundary of its safe envelope (Sha, 2001). Runtime assurance carries the pattern into aerospace flight control (Schierman et al., 2020) and spacecraft autonomy (Dunlap et al., 2024; Hobbs et al., 2025). Aviation learning-assurance guidance and the joint SAE and EUROCAE process standard in development place learned components inside a lifecycle with independent verification (European Union Aviation Safety Agency, 2024, 2026; Federal Aviation Administration, 2024; SAE International & EUROCAE, n.d.). In the nuclear domain, the NRC's gap assessment expects physics-based models in safety-significant roles (U.S. Nuclear Regulatory Commission, 2024), and operating experience with digital instrumentation and control identifies systematic software faults, common to every identical copy, as the dominant common-cause concern, with design diversity as the established defense (Liu and Ma, 2026). Calibrated uncertainty is a necessary part of this picture, but it is not sufficient on its own: predictive uncertainty degrades under dataset shift, exactly where an independent check is needed most (Ovadia et al., 2019). GURU applies the Simplex pattern across a hierarchy rather than a single loop. Learned agents are the high-performance tier, a validated solver, propagator, or detect-and-avoid algorithm is the verified tier, and the rules engine is the decision module between them.

## 2.3 Language-model agents, harnesses, and where authority lives

Much recent progress in applied AI comes from engineering around a general-purpose language model rather than from the model alone. Tool-using agents interleave reasoning with actions (Yao et al., 2023) and learn when to call external tools (Schick et al., 2023), and practitioners now describe the loop, tools, retrieved context, memory, retries, and guardrails around a model as its harness. Harnesses make models far more useful, and frameworks that pair a generative proposer with external critics show how much checking from outside the model can add (Kambhampati et al., 2024). Harness engineering and GURU share a premise, that capability comes from composition, a premise that runs from the Society of Mind through global-workspace architectures.

They differ in where authority lives (Table 1). In a harness, the model sits at the center and the scaffolding serves it; the model proposes, and in many deployments the same model, or a second model, also judges. A harness can make a proposal more useful; it does not change what is allowed to decide. In GURU, a validated reference sits at the center and the learned models serve it. Language models translate human intent into the Common Internal Language and help operators review evidence. Narrow learned specialists propose, rank, and accelerate. A symbolic verification layer checks every consequential action against typed, unit-aware hard limits, and a validated solver, propagator, or detect-and-avoid algorithm decides. For the applications in this paper, controllability has to mean a check that can refuse an action, not an instruction asking a model to be careful.

**Table 1:** A language model with a harness, compared with GURU.

| Dimension | Language model with a harness | GURU |
|---|---|---|
| Center of authority | The model; tools, memory, context, and guardrails serve it | A validated reference; learned models serve it |
| Role of the language model | Planner, actor, and often judge of its own work | Translates intent into the Common Internal Language and aids human review; makes no consequential decision |
| Learned components | One general model, sometimes in several prompted roles | Many narrow specialists, each trained, validated, versioned, and registered |
| How actions are checked | Self-critique, second-model review, output filters, external tools where available | Rules engine on typed, unit-aware records; a validated solver, propagator, or algorithm confirms |
| Independence of checks | Often shares the model's training lineage and blind spots | Designed for functional and implementation diversity from the proposer |
| Hard limits | Usually prompts or filters | Non-compensatory veto at the actuation gate |
| Human role | A participant in the conversation | Sets intent and constraints, approves consequential actions, retains override |

The same distinction is now being drawn from inside the community that built the strongest search-based systems. Graepel (2026), a core member of the AlphaGo team, argues that chain-of-thought output does not make a language model a reasoner, because the intermediate steps come from the same next-token process as the answer, and he identifies three missing properties: an explicit, persistent, and inspectable record of what the system holds as settled, doubted, and open; a separation between what the system knows and the machinery that manipulates that knowledge; and an account of how a conclusion was reached that is the actual route rather than a story composed afterward. His prescription is an independent part of the system that evaluates each step against evidence, as AlphaGo's search evaluated the moves its policy network proposed. GURU is built to those properties structurally rather than by prompting. The Common Internal Language is the explicit state: typed, unit-aware, provenance-carrying, and separate from the weights of any model (Section 3.3). The knowledge graph holds what the system knows apart from the agents that act on it (Section 4.4). The symbolic rules engine and the validated reference are the independent evaluators, and what a reviewer audits is the sequence of typed records, rule decisions, and solver confirmations, not a language model's account of itself (Section 3.5). We do not claim that this makes GURU a general reasoner in Graepel's sense; the evidence in this paper concerns workflow reliability. The narrower claim is that the properties he names as prerequisites for trustworthy machine reasoning are the same properties R2 and R5 require of a safety-significant system, and GURU satisfies them by construction.

A good harness can call external checkers, and the best do. The distinction this paper draws is narrower and, for licensing, decisive: whether an independent validated reference holds non-compensatory veto authority by design, and whether the system's failure modes are diverse from the checks that guard them (Liu and Ma, 2026). GURU itself uses harness-style engineering inside its translation layer and its software-control agents (Section 5.5). What it does not do is let that layer, or any learned component, decide what counts as safe.

### 2.4 World models and joint-embedding prediction

A world model predicts how the system an agent acts on will evolve, given its state and an action. Generative world models predict future observations in full detail; joint-embedding predictive architectures (JEPA) instead predict in a learned representation space, on the argument that predicting every pixel spends capacity on detail that does not matter for planning (LeCun, 2022). The approach has been demonstrated for images, video, and robotic planning (Assran et al., 2023; Bardes et al., 2024; Assran et al., 2025), and recent theory identifies when such models recover the latent variables of the underlying system (Balestriero and LeCun, 2025; Klindt et al., 2026). Section 4.3 develops the details GURU relies on.

### 2.5 Related work by GURU sub-module

No prior system, to our knowledge, combines all of GURU's parts, so we position each sub-module against its own literature. For agentic computational fluid dynamics, the CFD Vision 2030 roadmap and its 2026 update describe an agent architecture of planner, executor, monitor, and checker that leaves the solver in charge (Slotnick et al., 2014; Wissink et al., 2026); GURU is a leadership-scale realization of that pattern, building on earlier GURU simulation workflows (Grosvenor et al., 2022). For space domain awareness, conjunction screening has a long operational lineage (Healy, 1995; Kelso and Alfano, 2005) and recent deep-learning benchmarks for all-versus-all screening (Stevenson et al., 2023); GURU adds a self-supervised world model whose embeddings serve maneuver detection, out-of-family classification, and risk ranking at once (Grosvenor, Wahab, et al., 2025). For airspace decision support, NASA's DAIDALUS provides validated detect-and-avoid alerting (Muñoz et al., 2015), which GURU orchestrates rather than replaces. Detection of falling material in weather-radar data has a lineage in meteorite recovery (Fries & Fries, 2010; Smeresky et al., 2021), and the reentry work in Section 7 builds on it. For software control, multimodal agents that operate real applications are evaluated on open-ended computer tasks (Xie et al., 2024); GURU couples such agents to an explicit, provenance-preserving knowledge graph and role-specific adapters (Section 4.4). For multi-agent control, centralized-training, decentralized-execution methods and sampling-based model-predictive control are established (Lowe et al., 2017; Williams et al., 2017), and the multi-aircraft application in Section 8 composes them inside GURU's hierarchy. For reactor digital twins, exascale coupled multiphysics is now tractable (Merzari et al., 2021, 2023); GURU's contribution is the separation of authority between learned acceleration and validated physics in a control loop.

## 3. The GURU Platform Architecture

GURU is a platform for autonomously driving digital engineering and operations workflows across software, data, simulation, and physical-control environments. It has three parts. The first is a lightweight interface that runs in a browser on the user's own device and captures intent, context, and environment state. The second is a deployment and orchestration backend that places containerized environments on the compute the user chooses, from a local machine to a cloud cluster or, for training, a leadership-class supercomputer. The third is a multi-agent system coordinated through a global workspace, in which agents are built and trained by the Learning Engine, communicate in a shared Common Internal Language, and are checked by a rules engine before any consequential action executes (Figure 2). The platform is designed to be software-agnostic: its interface, orchestration, Learning Engine, and Common Internal Language do not depend on any one solver or application, while individual agents are trained for specific software and domains. Any specific application, such as a CFD solver or a visualization package, is one workload among a growing set of digital engineering and operations workflows that agents are trained, validated, and registered to drive, not the purpose of the platform. The sections below describe each part and the way they fit together.

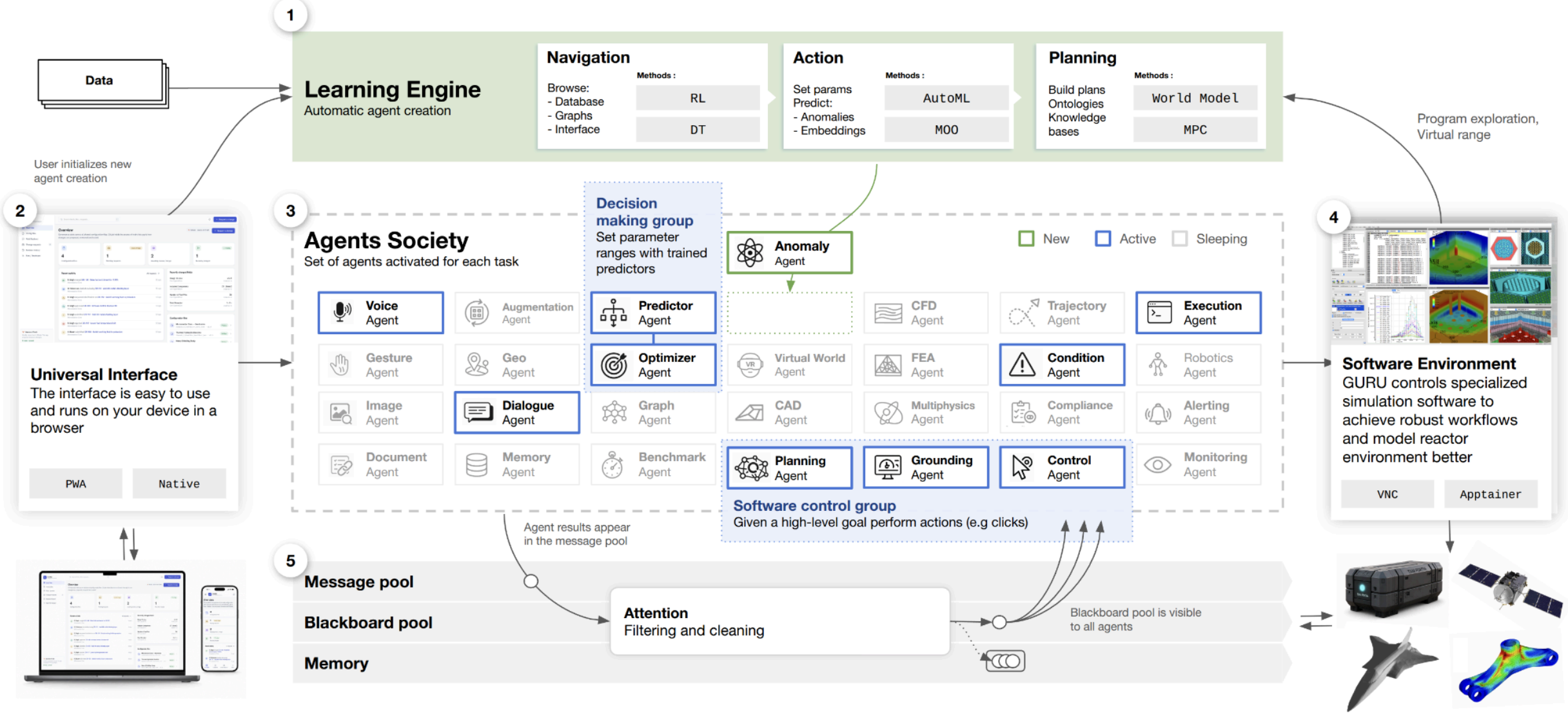

***Figure 2:*** *The GURU platform. The Learning Engine trains Navigation, Action, and Planning agents. The rules engine validates schemas, ontologies, constraints, and execution state. The agent society coordinates specialized agents through message pools and a shared workspace for auditable execution. Inputs span sensor streams, simulation outputs, documents, and direct user interaction, and the same platform drives applications in digital engineering, space operations, and reactor operations.*

## 3.1 On-device interface and remote backend

The interface is a browser-based application that runs on phones, tablets, desktops, and immersive displays. It accepts speech, text, images, and direct interaction with on-screen controls, and it presents plots, job status, remote views, and workflow controls. It does not perform the heavy reasoning or training locally. Instead, it captures intent, context, and environment state and routes them to remote backends that handle planning, learning, agent coordination, model execution, and long-running autonomous workflows. The device is the control surface, and the backend is the computational substrate (Figure 3). For specialized software that cannot be fully driven through an interface, GURU runs the application on a remote system and streams its display, so that an operator can watch, verify, and intervene. Control agents can act on an application through its programming interface or through the streamed display; the software-control results in Section 5.5 use structured application state, and control through the remote display alone remains under evaluation.

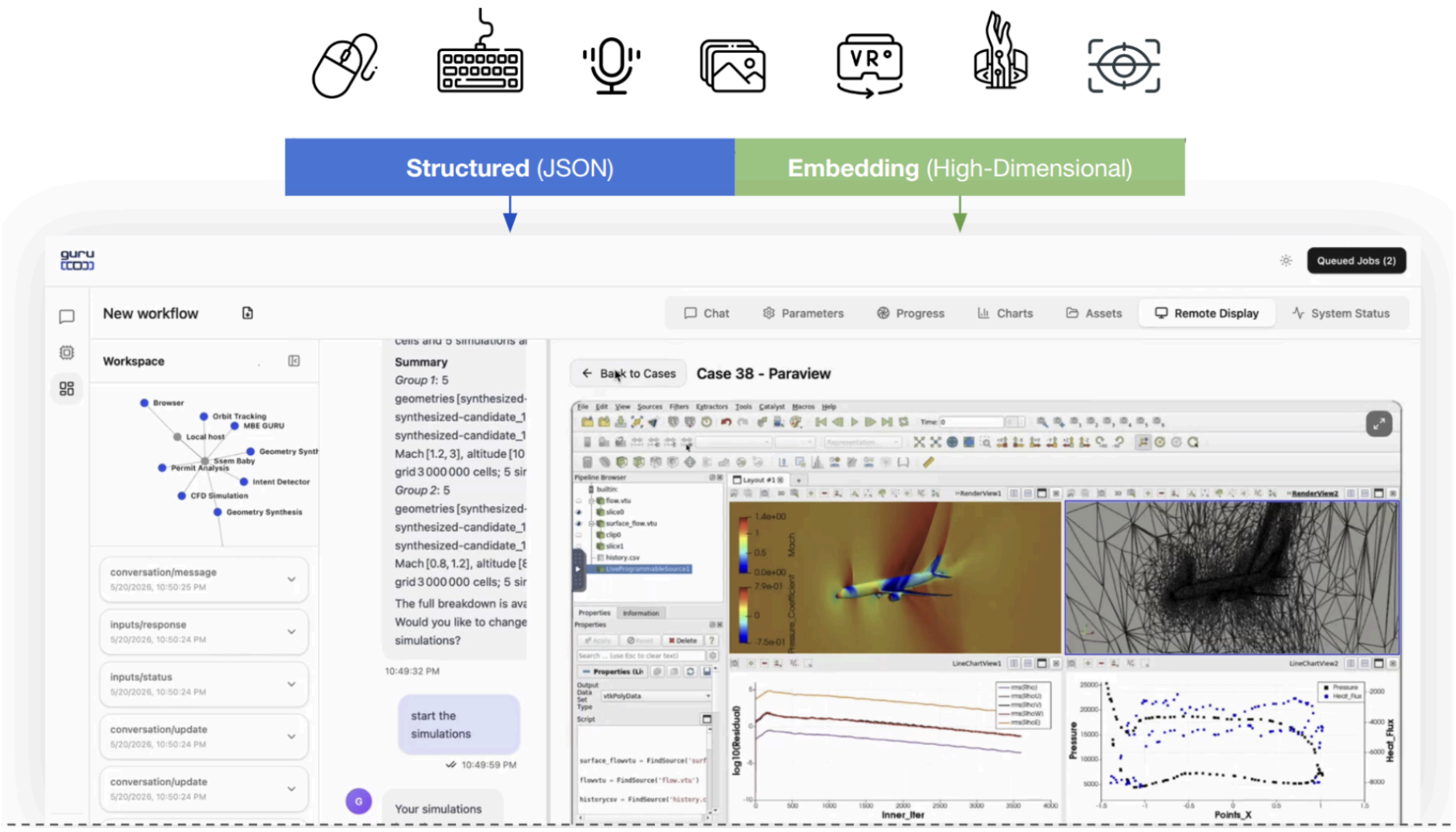

*Figure 3: Lightweight clients capture multimodal input and workflow intent. Structured representations and high-dimensional embeddings are routed to backend services for reasoning, orchestration, and execution across local, cloud, and high-performance computing environments. Remote applications can be driven and viewed through a streamed display.*

## 3.2 The agent society and global-workspace coordination

GURU organizes its capabilities as a society of agents. Throughout this paper, an agent is a software component with a narrow function, defined inputs and outputs in the Common Internal Language, and a bounded degree of autonomy. An agent may wrap a learned model, a deterministic tool, or both. The term is broader than its reinforcement-learning sense: some GURU agents contain reinforcement-learning policies, but most do not. Each agent encapsulates a narrow function and a degree of autonomy: perception, representation, prediction, optimization, anomaly detection, constraint checking, software navigation, execution monitoring, or domain-specific reasoning.

**Table 2:** The GURU layers, representative agents and methods, and the failure each layer is designed to control.

| Layer | Typical agents | Representative methods | Primary failure controlled |
|---|---|---|---|
| **Navigation** | Interface control, file-system and database navigation, remote-display control | Graph methods, computer vision, OCR and vision-language perception, reinforcement learning, state graphs | Loss of contact with live external state |
| **Action** | Anomaly detection, parameter prediction, grid-quality scoring, classification, optimization | AutoML, JEPA world models, PatchTST, graph neural networks, multi-objective optimization | An invalid bounded decision or a weak uncertainty estimate |
| **Planning** | Workflow planning, multi-step execution, scenario generation, policy search, control-response selection | Decision transformers, reinforcement learning, model predictive control, JEPA world models, language and vision-language planners | Open-loop plan drift or an unverified action |
| **Symbolic verification** | Rules, schemas, units, preconditions, solver contracts, safety limits, provenance checks | Common Internal Language, deterministic rules, satisfiability-modulo-theories checks | An unbounded learned output or a silent assumption mismatch |

For a given task, only the relevant agents activate and form a temporary coalition. System-level capability comes from their composition, not from a single controller. Coordination uses a global-workspace design.

Low-priority and exploratory information flows through event-driven message pools. Salient information is promoted to a shared blackboard, where relevant agents observe shared context and self-activate. Critical execution paths, such as constraint violations, safety checks, job failures, and required handoffs, use deterministic peer-to-peer channels rather than opportunistic broadcast. The result combines event-driven adaptability with controlled orchestration, so that the system can respond asynchronously to change while workflow-critical transitions remain explicit, validated, and auditable.

The three operational layers and the symbolic verification layer can be summarized by the function each performs and the failure each is designed to prevent. Table 2 lists representative agents and methods for each.

## 3.3 The Common Internal Language

The agents share one representation, the Common Internal Language, which is what allows agents, models, tools, and provenance records to operate as a single coherent system (Figure 4). The Common Internal Language is neuro-symbolic. Its symbolic component represents concrete values, rules, procedures, and domain structures explicitly. Its learned component carries embeddings and model-derived representations for complex or ambiguous inputs. It is hierarchical, which preserves control over each transformation, and it records provenance, which supports audit and trust. When a new domain is introduced, knowledge is extracted from domain material and encoded as Common Internal Language data, and agents are trained on that representation. During operation, multimodal user input is also translated into the same representation, so that agents can exchange messages, reason over shared structures, and produce checked outputs. Because the representation states types, units, physical limits, uncertainty, and provenance explicitly, a violation such as an out-of-range value or an inconsistent unit can be detected symbolically rather than remaining hidden inside a network. It is also the explicit, persistent, and inspectable state that Graepel (2026) identifies as missing from language models, whose knowledge and reasoning are interwoven in their weights (Section 2.3). Large language models have a deliberately limited role here. They translate human intent into the Common Internal Language. They do not make consequential engineering or operating decisions.

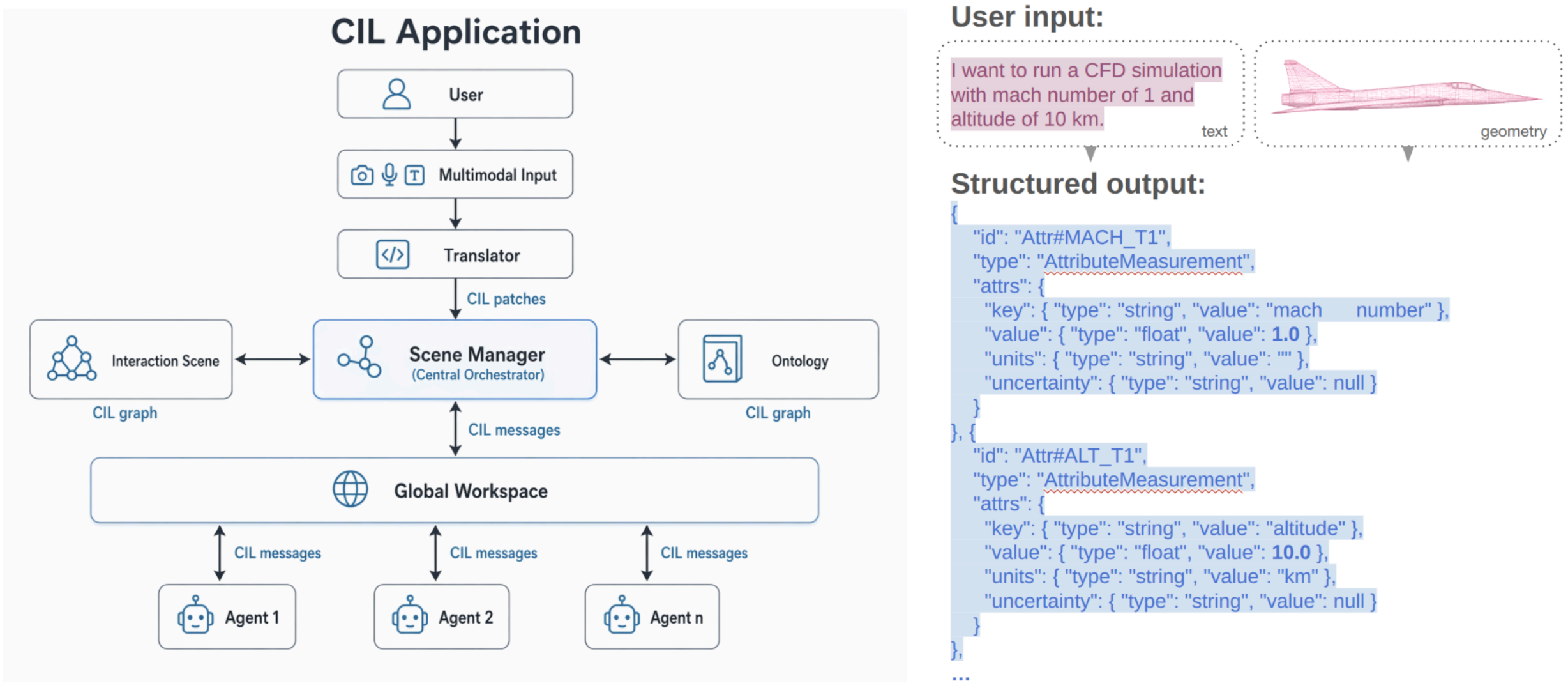

***Figure 4:*** *The Common Internal Language unifies agents through a shared semantic substrate for concepts, values, rules, and procedures, a uniform message format for the global workspace, and hierarchical task structures ready for planning and execution. Inset: a natural-language and geometry request is translated into typed, unit-aware attribute records with explicit provenance and uncertainty fields.*

In implementation, Common Internal Language records are serialized as typed JSON objects (Figure 4); the semantics lie in the schema, the typed references, and the rules that check them, not in the serialization. The knowledge-extraction schema that defines its symbolic vocabulary has ten semantic object families (Table 3). A record may also carry learned fields, such as an embedding of an ambiguous phrase or a geometry, which agents use for similarity and retrieval but which the rules engine never treats as evidence of validity.

**Table 3:** Semantic object families of the knowledge-extraction schema.

| Object type | Meaning |
|---|---|
| Entity | An identifiable object or concept: a component, parameter, application, or API symbol |
| Constraint | A requirement, prohibition, range, or type restriction on another object |
| Relation | A named, non-temporal relationship between two objects |
| Rule | A reusable conditional policy or invariant, with conditions and effects |
| Decision | A choice, its alternatives, the selected option, and the rationale |
| State | A property value associated with an object and an observation phase |
| Action | An observed or instructed operation, with a target, arguments, and condition references |
| Procedure | A goal-directed workflow composed of Action or nested Procedure invocations |
| Temporal | An ordering or timing relationship: before, after, during, overlaps, or repeats |
| Artifact | An input, intermediate, or output work product, such as a dataset, file, or figure |

Objects are connected through typed references. Constraints identify the subject they restrict. Rules reference their conditions, effects, and scope. Relation and Temporal records preserve evidence for semantic and temporal assertions, while accepted connections are also exposed as graph edges. Object identity and the provenance of an assertion remain distinct.

The procedure language uses structured records rather than a flat list of instructions. Each Procedure records a goal, local steps, and transitions. A step has a step identifier for the invocation and a node identifier for the Action or Procedure invoked, so two steps can invoke the same operation without becoming the same step. Transitions reference source and target steps and distinguish success, conditional, retry, alternative, failure, and unknown paths; condition references retain the evidence for guards. Actions carry arguments and precondition and postcondition references. Steps and transitions are supporting control-flow records, not additional members of the ten semantic families.

Evidence references and confidence accompany extracted records. These fields support inspection; confidence is not proof of correctness, and structural validity does not establish successful execution or complete engineering validation.

## 3.4 Adding new capabilities

New capabilities are added through a repeatable training-factory pipeline, described in Section 4. A task is specified through schemas, constraints, and success metrics. Training data is collected or generated from environment exploration, tutorials, manuals, screenshots, configuration files, human demonstrations, simulation traces, and synthetic trajectories. A model for the agent is then trained in a controlled, reproducible way and selected, and the agent is validated and registered. Because every agent is versioned and regression-tested before use, adding a new capability does not silently break a capability that already worked.

## 3.5 The symbolic rules engine and runtime assurance

**The symbolic layer is load-bearing, and an ablation shows it.** Alongside the learned components, a symbolic rules engine, acting as a runtime assurance gate, checks every consequential action against hard limits and execution preconditions before that action can reach a tool or an actuator. This layer is not optional decoration. It is doing measurable work. In the component ablation reported in our prior peer-reviewed work (Grosvenor, Zemlyansky, et al., 2025, Table 5), disabling the symbolic constraints reduced the boundary-layer capture of generated grids (defined in Section 5.4) by 13 percentage points, from 98 percent with the full system to 85 percent, on a corpus of 312 aircraft surfaces with results averaged over three repetitions. In the same ablation, disabling the JEPA embeddings cost 23 points and disabling reinforcement-learning planning cost 18 points, and the corresponding losses in F1 on the space domain awareness task were 0.06, 0.04, and 0.08. Each component contributes measurably on its own, and the symbolic layer's contribution appears in both domains. That ablation used the earlier snappyHexMesh-based grid generator rather than the extrusion-based generator of Section 10.1, so it is evidence about the architecture, not a measurement of the current workflow. The same principle is applied throughout the system. Learned models assist and accelerate state prediction, calibration, control-policy development, and the exploration of candidate actions. They do not replace the validated tools that decide, and they do not become the safety basis. Because a learned model cannot reliably recognize when its prediction has left the region it was trained on, each agent is designed to be wrapped in a protective pattern: an out-of-distribution gate in front, a calibrated uncertainty estimate behind, and a refusal path that hands control to a validated fallback when either fails. The principle is constant: learned models accelerate and rank while validated physics confirms and decides.

**The rules engine.** The symbolic verification layer is implemented by a rules engine. Its rules are Constraint and Rule records in the Common Internal Language (Table 3): typed preconditions, unit and range checks, solver contracts, and hard safety limits, each with provenance. Before a consequential action reaches a tool or an actuator, the rules engine evaluates the action and its predicted effect against the applicable rules, using a satisfiability-modulo-theories check where constraints interact. A violation blocks the action, demotes the proposal, invokes a defined fallback, or escalates to the human, according to the rule's severity. Representative rules include unit and range consistency for user-specified flight conditions (Figure 4) and mesh-readiness thresholds before a solve in computational fluid dynamics; legality of a tactical option, which requires enough active aircraft and communication connectivity, in multi-aircraft control (Section 8); and, by design, rejection of any control-drum command whose predicted reactivity or fuel temperature exceeds its limit in reactor control (Section 9). Throughout, "runtime assurance" refers to this gate at the point of action.

**The global workspace as design rationale.** The global workspace is a coordination design rather than a claimed ablation result. Broadcasting salient state, anomalies, predictions, and constraint violations, so that the relevant specialists act on it is what keeps workflow-critical transitions explicit, validated, and auditable. In the current system this coordination is realized through domain-specific orchestrators, for example for computational fluid dynamics and for space operations, that activate and sequence the relevant modules. We report this as design rationale, kept distinct from the measured thirteen-point compliance drop reported above for removing the symbolic verification layer, which is an ablation result.

**Why this matters for trust.** The distinction we hold throughout this paper is between a learned model that helps decide which validated computations to run and a model whose own output is the final word. GURU uses self-assessment wherever it helps: calibrated uncertainty, perturbation-based confidence, and prediction error all rank, flag, and route work. None of them decides. A learned model's confidence is evidence about its own reliability, and it is weakest in exactly the conditions that matter most, inputs unlike its training data, where it can be confidently wrong while sharing every blind spot of the model it is judging (Ovadia et al., 2019). Safety regulators expect physics-based models in safety-significant roles and independent monitors behind learned components (U.S. Nuclear Regulatory Commission, 2024; Schierman et al., 2020). A proxy that replaces validated physics, serves as the sole judge of its own output, or becomes the safety basis is exactly the design those expectations rule out, and it is not what GURU does. In GURU, the learned model proposes, ranks, compresses, predicts, or accelerates; an independent check confirms; and the validated reference decides. This is the property that lets a population-scale autonomous search remain physically valid, and it is what allows an autonomous controller to be trusted under supervision.

Some parts of the factory are purely deterministic by design, because the task demands deterministic guarantees. One example is the geometry-readiness layer used in the digital engineering application example of computational fluid dynamics. It is a comprehensive grid-quality metrics library that implements established formulations from the grid-generation literature, with more than 30 per-element metrics across surface and volume meshes, configurable readiness thresholds, and support for 15 mesh file formats through a single evaluation call. It is paired with a mesh-repair backend that realizes a variational mesh-optimization framework directly in an automatic-differentiation library, so that only the forward evaluation is written and exact gradients are obtained automatically. This deterministic layer is a permanent, standalone part of the pipeline. Every mesh that passes through GURU is scored and, where necessary, repaired by these tools, whether or not a learned model is available, and this deterministic layer remains the final arbiter of readiness.

# 4. The Training Factory: Data Factory and Learning Engine

The training factory is how GURU converts raw domain material into validated, reusable agents. It has two stages. The Data Factory prepares datasets, and the Learning Engine trains and compares candidate models, selecting the best model for each target for use in a new or existing agent.

## 4.1 Data Factory

The Data Factory transforms raw inputs from simulations, software workflows, interaction logs, and domain records into structured, validated, reusable training assets. It performs the core data-science operations before any model is trained: ingestion, cleaning of missing or invalid values, format normalization, feature extraction, distribution analysis, outlier detection, and metadata preparation. It then creates reproducible training, validation, and test splits, and it can generate alternative views of a dataset for ablation and sensitivity analysis. For parameterized and tabular problems it uses structured sampling, in particular Latin hypercube sampling (McKay et al., 1979), so that a representative spread of input combinations is generated where exhaustive evaluation would be impractical. The same machinery operates at leadership scale. An autonomous geometry-generation pipeline built on an open parametric aircraft modeler produced a large corpus of paired data, including multi-view image embeddings and simulation-quality surface meshes for downstream training.

## 4.2 Learning Engine

The Learning Engine is the model factory behind GURU's agents. Rather than commit to a single model family, it trains and compares several under comparable conditions on high-performance computing resources, then registers the best model for each target. Its library spans automated machine learning for tabular prediction, transformer (Vaswani et al., 2017), vision-language, and vision foundation models (Radford et al., 2021; Oquab et al., 2024) for language and multimodal interpretation, decision transformers (Chen et al., 2021) and reinforcement learning (Schulman et al., 2017) for action and skill acquisition, graph neural networks for structured navigation, and Joint Embedding Predictive Architecture world models for representation learning and state-transition prediction. Two methodologies recur across the applications in this paper, so we describe them here.

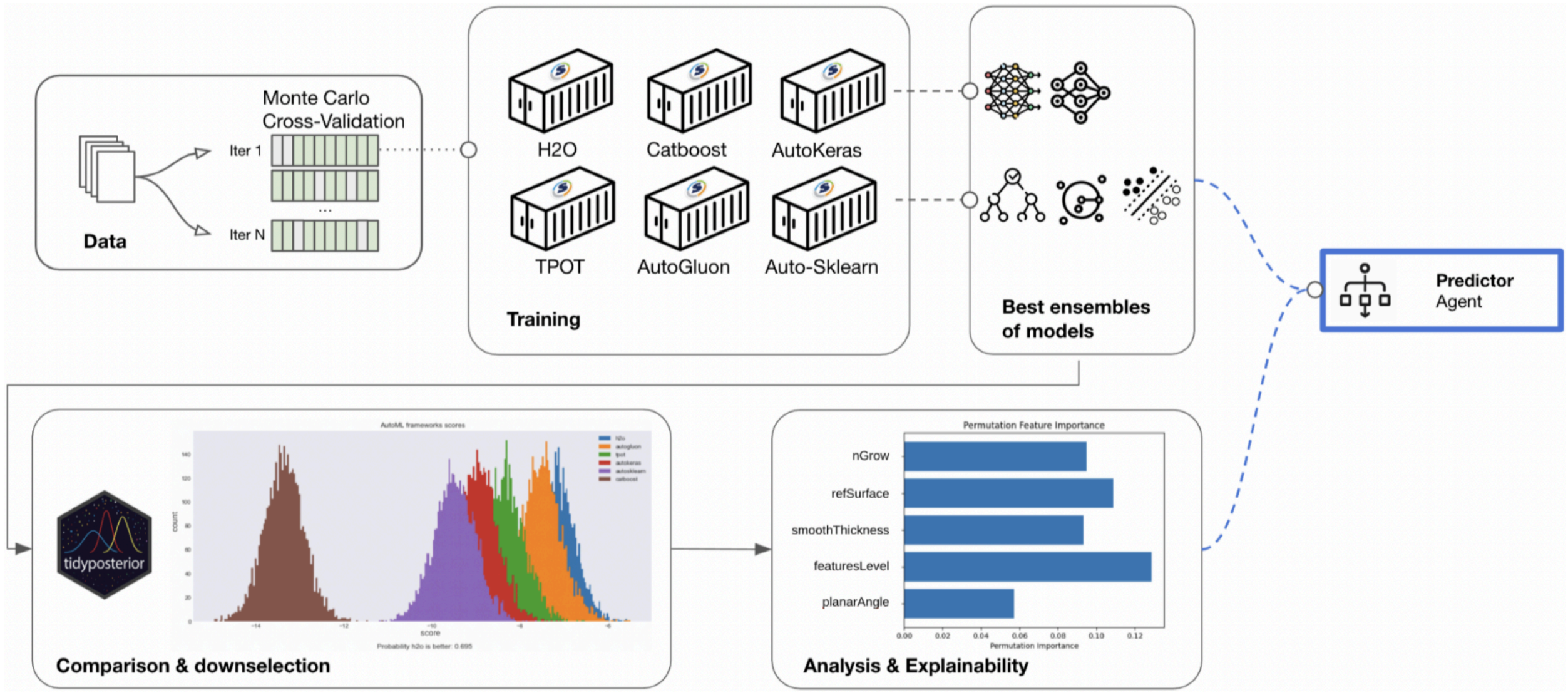

***Figure 5:*** *Data Factory outputs feed multiple automated-machine-learning frameworks trained in parallel. Candidate models are evaluated, compared, and ranked under Monte Carlo cross-validation, and the best predictor for each target is registered for downstream use in the workflow.*

**Automated machine learning for tabular prediction.** Many decisions inside a digital engineering workflow reduce to predicting an outcome from a structured record. The Learning Engine trains several automated-machine-learning frameworks in parallel for these targets, including AutoGluon (Erickson et al., 2020), auto-sklearn (Feurer et al., 2015), AutoKeras (Jin et al., 2019), TPOT (Olson et al., 2016), H2O (LeDell and Poirier, 2020), and CatBoost (Prokhorenkova et al., 2018), because each brings a different strength, such as genetic programming, meta-learning, automatic model selection, or efficient gradient boosting. Latin hypercube sampling generates a diverse spread over both the input features and the framework hyperparameters (Balaprakash et al., 2018), so the search covers data-space and model-space variation without exhaustive enumeration. Each framework runs as a scalable job with configurable node, core, and thread counts. After training, candidate models are evaluated in parallel on the validation set, and their metrics are collected, compared, and ranked so the best model for each target can be selected (Figure 5). These learned predictors are used inside the workflow to evaluate and rank candidate settings. The validated solver still runs and decides.

**World models with JEPA.** The second recurring methodology is a Joint Embedding Predictive Architecture world model. An encoder maps an observed context, such as an interface state, an interaction history, or a sequence of physical observations, into a latent embedding. A predictor then maps the embedding of the current state, together with an action or a prediction horizon, into the embedding of a future state. The model learns the dynamics of state transitions without reconstructing raw observations, which encourages compact, behavior-organized representations. The predictor can be a transformer-based time-series model for scalable temporal modeling or a lightweight network for simpler settings, and the target is computed either by a slow exponential-moving-average copy of the context encoder (Assran et al., 2023) or, in a variant without that copy, by the same encoder under an isotropic-Gaussian regularizer that prevents the representation from collapsing (Balestriero and LeCun, 2025); Section 10.3 compares the two. The JEPA world model learns the signal in a compact latent space and filters out noise. It is a world model, defined operationally as a trained encoder and predictor, and it does not replace the validated physics that decides; it is used for prediction, for anomaly detection through prediction error, and as an informative state for planning and control (Figure 6).

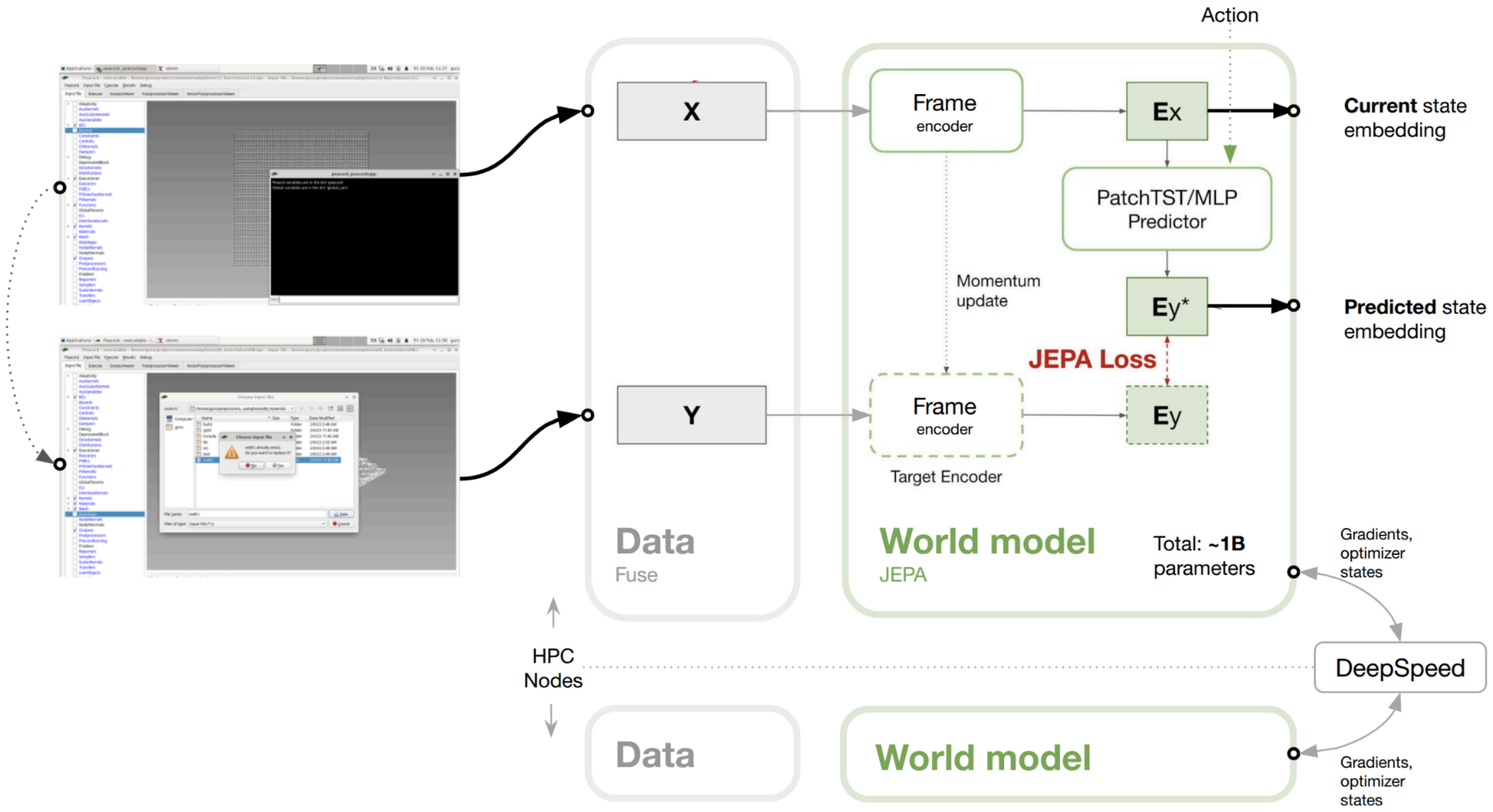

***Figure 6:*** *Current and target states are encoded into latent embeddings. The predictor learns action-conditioned transitions in that latent space, which lets agents forecast the behavior of an interface or environment without reconstructing raw observations.*

## 4.3 World models and the JEPA thesis

A central design choice is how an agent represents and predicts the dynamics of the system it acts on. One option is to generate the world in full detail, for example as predicted video. The Joint Embedding Predictive Architecture takes a different position. It predicts the future in a compact representation space rather than at the level of raw observations, on the argument that predicting every pixel wastes capacity on detail that does not matter for planning (LeCun, 2022). The original architecture was demonstrated for images (Assran et al., 2023), then for video (Bardes et al., 2024), and most recently for planning, where a video world model was used with receding-horizon model predictive control on real robot hardware in a zero-shot setting (Assran et al., 2025).

The 2026 literature has clarified when this family of world models works and why. A systematic study of physical planning isolated the encoder, the predictor architecture, the training objective, the context length, and the planner, and assembled a recipe on simulated and real robot data that outperformed two strong baselines (Terver et al., 2026). Among its findings are that the predictor should be trained on multi-step rollouts rather than single-step teacher forcing, that context length has an intermediate optimum, that data diversity is the most dependable lever, and that sampling-based planning with a simple cost is the most robust default. On the theory side, recent results prove that a Gaussian-regularized world model recovers the latent variables of the underlying system under stationary additive-noise dynamics, and identify exactly where that guarantee breaks (Balestriero and LeCun, 2025; Klindt et al., 2026). The boundary matters for us, because the place where the guarantee fails, an abrupt non-stationary transition such as a maneuver, is precisely where the prediction-error channel becomes the detection asset rather than a liability. Work that stabilizes end-to-end training with a single regularizer (Maes et al., 2026), and work that adapts the architecture to time series and trajectories (Ennadir et al., 2025; Li et al., 2024), complete the context we build on.

**Our position.** We treat the JEPA world model as one carefully bounded component, not as the system. It is used where prediction in representation space is the right tool, and it is wrapped by three safeguards: a predictor and corrector loop, in which the learned predictor forecasts the next state and a validated physics or measurement step corrects that forecast before it is acted on; symbolic verification; and planning under explicit cost. We adopt the demonstrated precedent that a JEPA world model can plan by model predictive control (Assran et al., 2025), and we state our addition plainly: for safety-critical orbital and reactor use, the predictor-corrector loop and symbolic verification are joined by an out-of-distribution gate with a refusal path.

## 4.4 Learning from documents, videos, and exploration

Expert software knowledge is distributed across manuals, API documentation, tutorials, and demonstrations. These sources describe not only which operation to invoke, but also the conditions under which it is appropriate, the objects it acts on, and the result a user should expect. GURU converts this material into reusable procedural knowledge and role-specific training examples. The objective is to make application expertise available to a locally served model through both explicit retrieval and learned behavior.

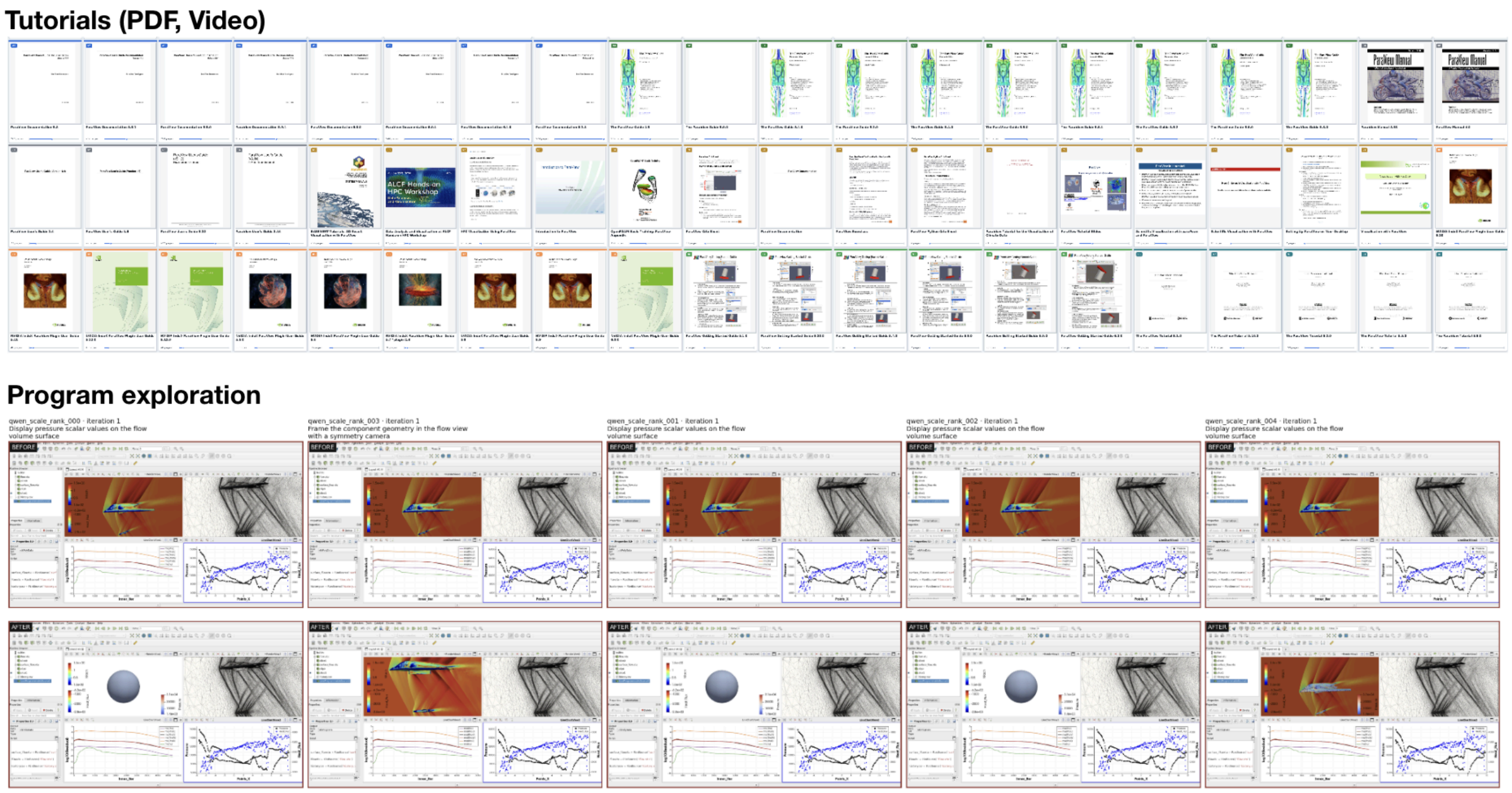

***Figure 7:*** *Multimodal sources for software-knowledge acquisition. Document and video tutorials provide procedural demonstrations and application knowledge, while autonomous ParaView exploration produces state-action-outcome trajectories. The exploration examples show before-and-after application states; several attempts did not achieve the requested goal, which is why execution without an error is never treated as achievement. Both sources are transformed into reusable observations for knowledge extraction and agent training.*

The extraction process starts from bounded source fragments. Document passages and video frames are processed into structured observations, retaining source references and associated textual evidence where available. The observations identify operations, goals, affected entities, conditions, and relationships. Tutorial evidence can therefore contribute both high-level descriptions of a task and concrete steps for carrying it out. API documentation supplies exact symbols and calling contracts that complement the procedural descriptions extracted from demonstrations.

Autonomous exploration contributes another form of evidence: a requested goal, the actions attempted by an agent, and the resulting application observations. These records can describe successful execution, partial progress, or a different outcome from the one requested. Relabeling an achieved outcome is useful only when it remains aligned with the corresponding action span. Execution without an error does not, by itself, prove achievement of the semantic goal. Keeping these distinctions in the source record allows the same evidence to support graph construction and supervised training without equating every recorded interaction with a verified skill. Exploration also scales: in one leadership-scale run on 1,024 nodes, graph-based exploration of ParaView grew the useful coverage of its interface from 1,261 states and 2,948 transitions to 17,823 states and 33,075 transitions at constant compute.

### 4.4.1 Knowledge graph

Structured observations are converted into claims about objects and their relationships. Each claim retains its evidence and source-local references. Schema and reference checks assess whether the records are well formed and whether linked objects can be resolved. These checks establish structural consistency; the supporting material is still needed to assess the meaning and correctness of a claim.

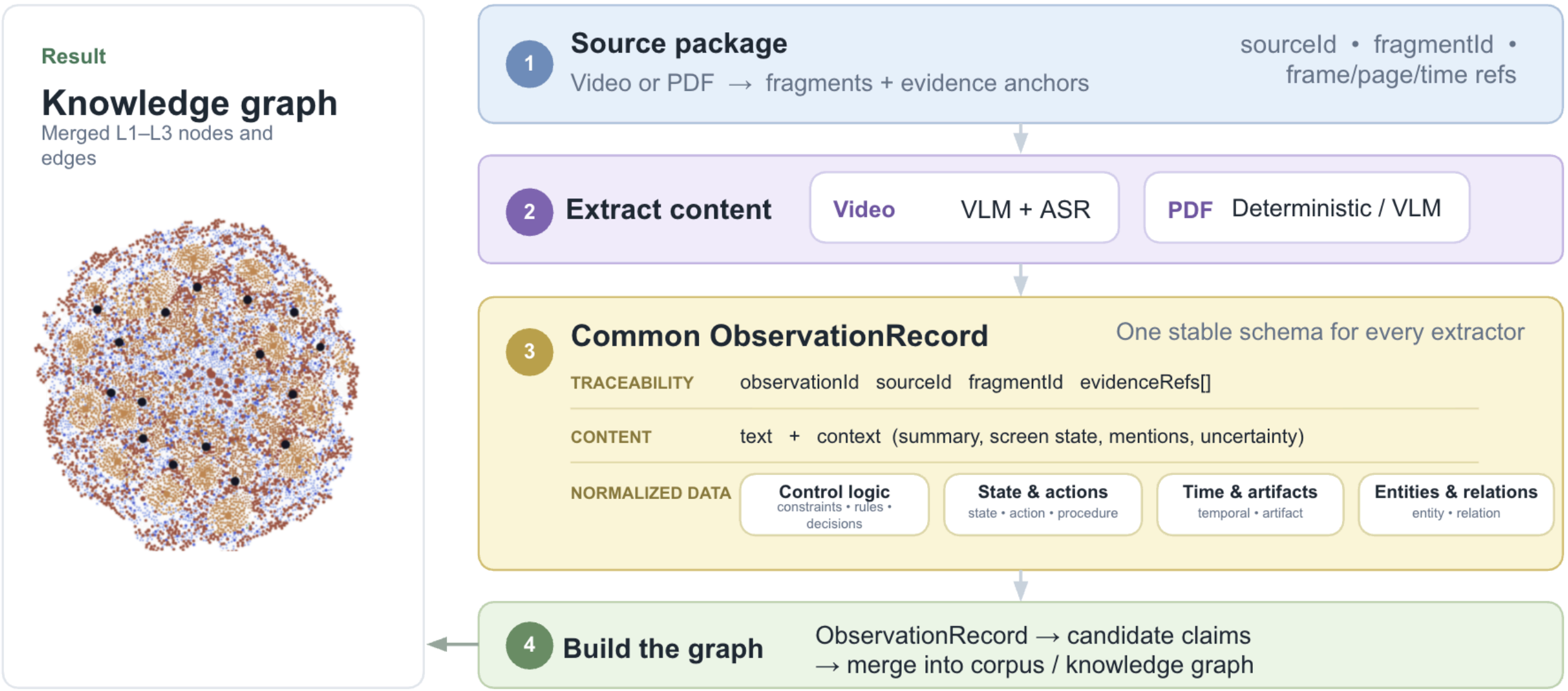

***Figure 8:*** *Knowledge-graph construction pipeline. Heterogeneous software sources are converted into a common, provenance-preserving observation representation, using vision-language models and automatic speech recognition (ASR) for video and deterministic or vision-language extraction for PDF documents, before being transformed into candidate claims and merged into a canonical knowledge graph. The shared intermediate schema allows document and video evidence to contribute to the same reusable representation.*

Claims are assembled into a corpus graph, then materialized into a canonical graph that exposes reusable identities. Identity decisions connect source mentions that refer to the same object while preserving their separate evidence. Relationships connect operations to targets, prerequisites, outcomes, and available implementations. Procedure-local steps and transitions preserve order, nesting, and repeated invocations. Application or version attribution is retained when the source supports it. Unresolved identities and incomplete procedures require explicit treatment rather than being assumed complete.

The resulting graph separates the stored knowledge from the model that consumes it. Supporting records remain available for inspection, and repaired or newly extracted facts can be incorporated without retraining the agent. This makes the graph a reusable source of context across tasks, rather than a collection of task-specific prompts.

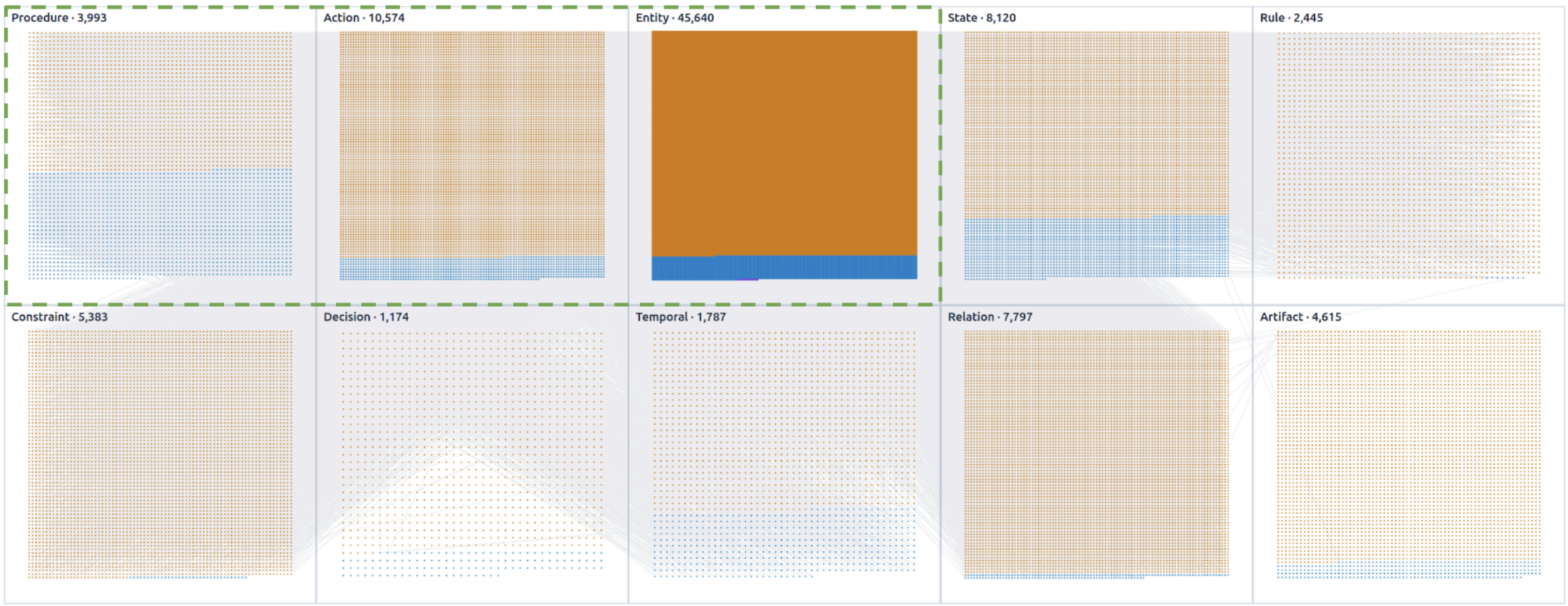

***Figure 9:*** *Knowledge-graph snapshot for ParaView by semantic family. The graph contains 105,125 nodes and 106,202 edges from 64 source packages (42 PDF, 19 video, two API, and one exploration package), including 1,552 exact API definitions, of which 1,478 carry nonempty signatures. The panels count nodes in the ten semantic families; the remaining nodes include procedure steps and transitions, which are stored as supporting control-flow records. These are corpus-wide counts, not counts of facts relevant to any single goal. Automatic semantic matching between documentation and exploration was disabled in this release, so shared storage should not be read as complete automatic linkage.*

### 4.4.2 Retrieval and role-specific fine-tuning

The extracted knowledge is used in two complementary ways: retrieval at inference time and role-specific model adaptation. For a new task, the controller queries the knowledge graph using the goal and available application context. Relevant procedures are expanded into their associated steps, relationships, implementations, and API entities, producing structured context that can be supplied to the agent without modifying the base model. In this way, application-specific knowledge remains explicit, inspectable, and replaceable rather than being encoded only in model weights.

The same evidence can also be converted into supervised examples for the agents that consume this knowledge. In the software-control application described in Section 5.5, the Planning agent learns to map goals and context to structured action sequences, while the Grounding agent learns to translate those actions into executable application commands. We adapt these roles independently using lightweight adapters while retaining the shared base model. Retrieval therefore supplies what is known about the application, whereas fine-tuning teaches each role how to use that information for its part of the control workflow.

## 5. Application I: Autonomous Digital Engineering, Computational Fluid Dynamics

Computational fluid dynamics is a demanding test of the architecture because a single useful result requires coordinated reasoning across geometry generation, mesh preparation, solver setup, convergence monitoring, and post-processing. A failure at any stage wastes the entire computation. This application demonstrates GURU as an autonomous digital-engineering system, developed and exercised on Department of Defense High Performance Computing Modernization Program systems. The aerospace CFD community has converged on the same direction. The 2026 update to the CFD Vision 2030 Roadmap identifies knowledge extraction and agentic workflows as an emerging frontier, describes a useful agentic CFD architecture as a planner, retrieval module, executor, monitor, checker, and reporter, and stresses that such agents are not yet mature enough to replace the solver or the analyst, with the checker comparing results against validation data and acceptance rules (Wissink et al., 2026), a direction the original roadmap anticipated (Slotnick et al., 2014). GURU is a concrete, leadership-scale realization of that pattern in which the validated solver still decides. Its Navigation, Action, and Planning agents and its rules engine fill those roles, and the roadmap's near-term targets, case setup, error recovery, post-processing, and provenance capture, are the stages the workflow below automates.

## 5.1 The end-to-end pipeline

The workflow runs in four connected stages. It begins with geometry synthesis. A 3-dimensional generation model is fine-tuned to produce aerial-vehicle geometries from images (Xiang et al., 2025), trained on synthetic data generated with an open parametric aircraft modeler (McDonald and Gloudemans, 2022) so that the model maps images to valid geometry families rather than to generic meshes. Because the original generation code relied on libraries specific to one vendor, we adapted the flow-matching component to run on Intel and AMD accelerators (Section 11), which matters for deploying generative geometry on heterogeneous supercomputers. The second stage standardizes orientation, because in fluid dynamics the definition of forward and up determines inlet and outlet assignment, lift and drag axes, and boundary-layer assumptions. A direction model trained on videos and paired frames rotates each geometry into a consistent aerodynamic frame. The third stage prepares the mesh. The surface is remeshed for quality and uniformity, then passed to an in-house grid-generation tool that grows a boundary layer of prism elements outward from the surface toward a target near-wall resolution and fills the remainder with tetrahedra, with local limiting where the full boundary-layer height would produce invalid cells. The fourth stage runs the flow solution with the open SU2 solver (Economon et al., 2016), monitors convergence, and then performs solution-adaptive mesh refinement by computing a metric field from a chosen flow variable and refining or coarsening the mesh to match the local flow features, repeating until the grid converges (Figure 10).

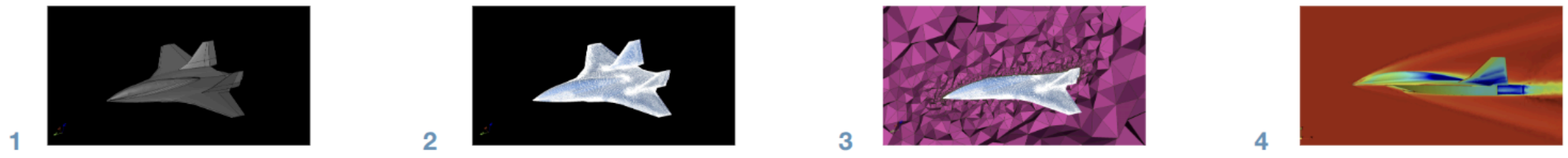

***Figure 10:*** *Geometry synthesis generates candidate vehicles. Surface generation produces watertight, simulation-ready surfaces. Volume generation creates the computational mesh. CFD execution performs solver setup, simulation, adaptation, and post-processing for downstream analysis and optimization.*

## 5.2 A single live run across three facilities

In a single live demonstration the system received 18 images of flight vehicles, synthesized a 3-dimensional geometry for each, and set up a suite of aerodynamic simulations spanning four representative flight regimes: subsonic at Mach 0.3 to 0.5, transonic at Mach 0.8 to 1.2, supersonic at Mach 1.2 to 3, and hypersonic at Mach 4 to 6, with matched altitude and angle-of-attack ranges. With three simulations for each geometry, the run comprised 54 cases. The jobs were submitted in parallel to three computing systems, a Department of Defense system and the Frontier and Aurora supercomputers, and, without advance reservations, each case was dispatched to whichever system accepted it first, the scheduling an autonomous workflow must handle on shared facilities. Cases were visualized as they converged and adapted, and a typical case completed three solution-adaptive refinement cycles in about 19 minutes. The run demonstrated the full path from images to converged, adapted solutions across the entire speed range, with the same agents driving heterogeneous hardware (Figure 11).

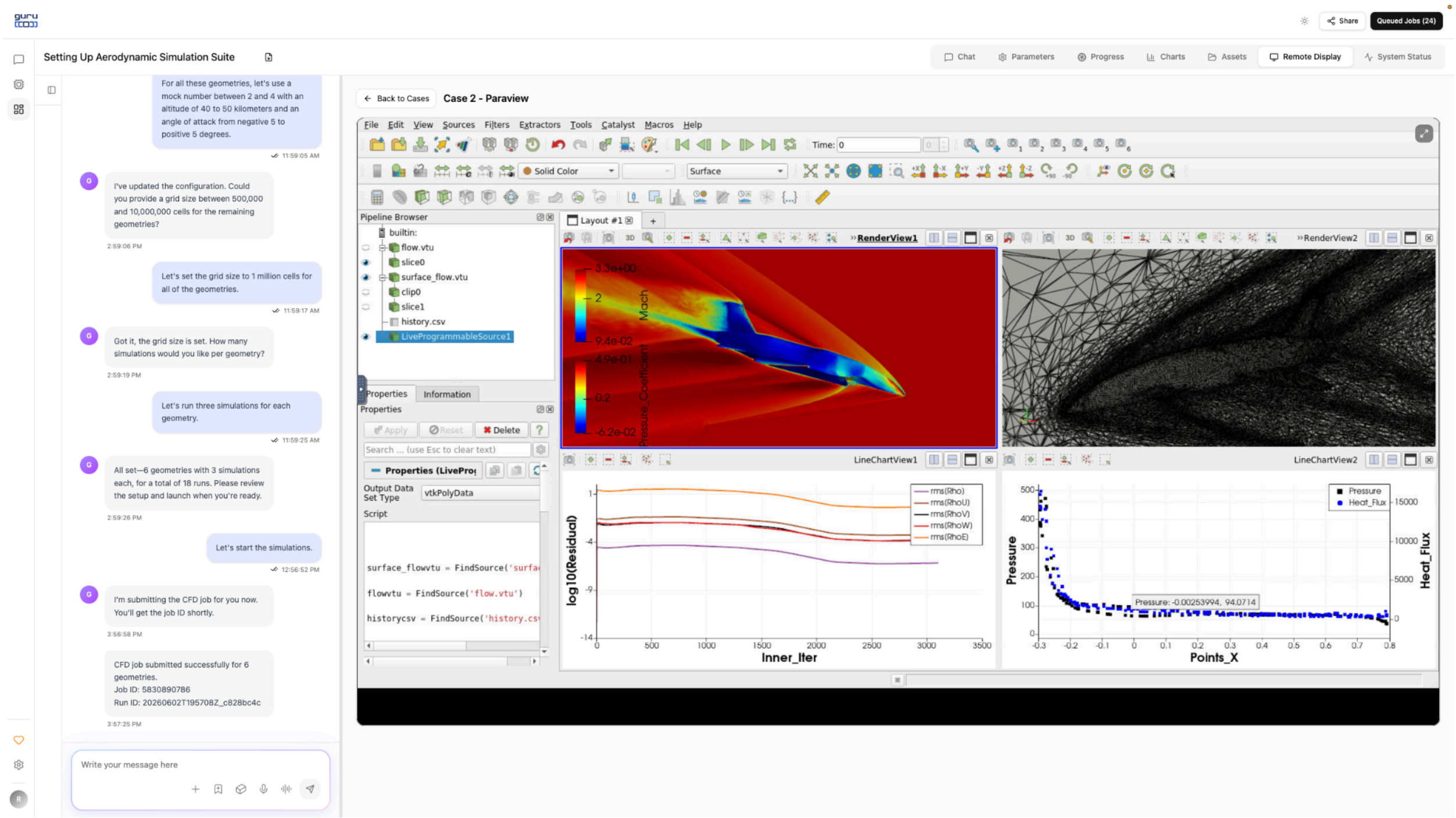

***Figure 11:*** *GURU CFD Suite workflow for high-speed aerial vehicles. The figure shows one representative airframe from the live run described above, the Fury: from a natural-language prompt, GURU configured an aerodynamic sweep at Mach 2 to 4, 40 to 50 km, plus or minus 5 degrees angle of attack, about 1M cells, and autonomously submitted it to HPC. Post-processed output for the Fury airframe shows Mach contours and surface pressure-coefficient distribution, the computational mesh, converged residual histories, and axial surface pressure and heat-flux profiles. Video: [https://youtu.be/GQIJjjVolNY](https://youtu.be/GQIJjjVolNY)*

## 5.3 Turning simulation experience into setup policy

The harder, slower part of digital engineering applications like computational fluid dynamics is not running one case but reliably setting up many (for example: thousands of flight conditions in a flight vehicle aerodynamic database; an entire parametric design study; or a simulation based digital twin of a nuclear reactor for control training and testing). This is the bottleneck the CFD Vision 2030 Roadmap returns to, where automated mesh generation with minimal user interaction remains an open challenge and model creation from image scans is flagged as a future direction (Wissink et al., 2026). It is the part GURU is built to absorb, generating geometry from images and meshing many candidate configurations per case. GURU does not introduce a new meshing algorithm; for unstructured prism-tetrahedral grids on conceptual geometries, it addresses the automation and robustness part of that challenge, selecting parameters, repairing surfaces, and checking readiness so that boundary-layer-resolving meshes are produced without manual intervention.

GURU treats each completed or failed run as a structured example linking geometry features, mesh-quality indicators, flow conditions, solver settings, and outcome, and uses these to train learned predictors for specific setup decisions. Two large data-science campaigns informed the current policies. We examined a large-scale volume-grid campaign spanning 11 aircraft classes, from fixed-wing subsonic through hypersonic to rotorcraft and unconventional configurations, using analysis of variance, mutual information, and a tree-ensemble stack of Random Forest (Breiman, 2001), XGBoost (Chen and Guestrin, 2016), and LightGBM (Ke et al., 2017), with feature attribution by SHAP values (Lundberg and Lee, 2017). Each method's feature importances are normalized before comparison (Figures 12b and 12f), and the cumulative ensemble score in Figures 12a and 12e adds a feature's normalized importance across the three tree models. It identified surface-mesh quality and a small set of grid-growth parameters as the strongest and most stable drivers of grid-generation success (Figures 12a to 12d). A flow-solver campaign of about 15,000 cases showed that the user-specified flight conditions, in particular the combination of Mach number and geometry scale, which together set the Reynolds number, dominate convergence, with solver settings such as the multigrid level decisive inside that envelope (Figures 12e to 12h). These learned predictors evaluate and rank candidate settings and define ranges for the next refinement campaign. The solver still runs and confirms the result.

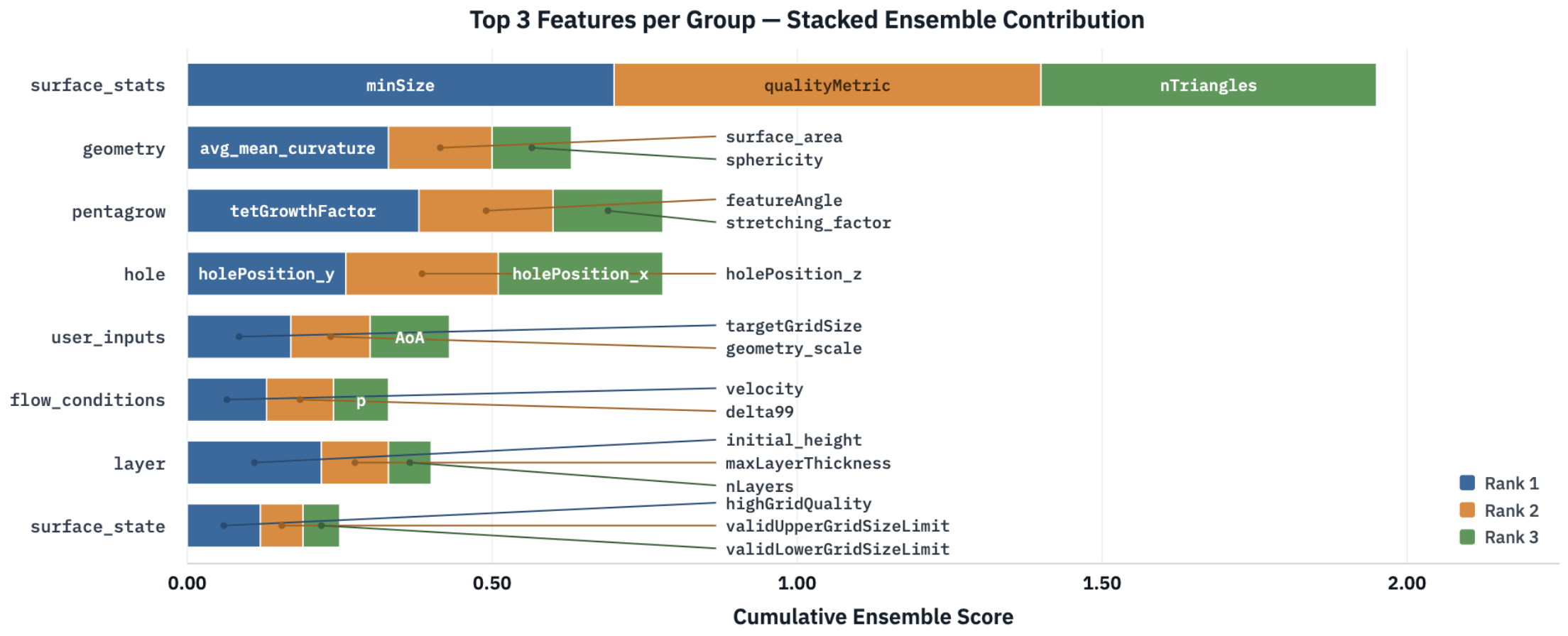

***Figure 12a:*** *Top 3 features per feature group, scored as a stacked contribution from Random Forest, XGBoost, and LightGBM. Surface-mesh statistics (minSize, qualityMetric, nTriangles) dominate, with Pentagrow grid-growth parameters (tetGrowthFactor, featureAngle) second, which is why the next refinement tightens surface-mesh quality upstream.*

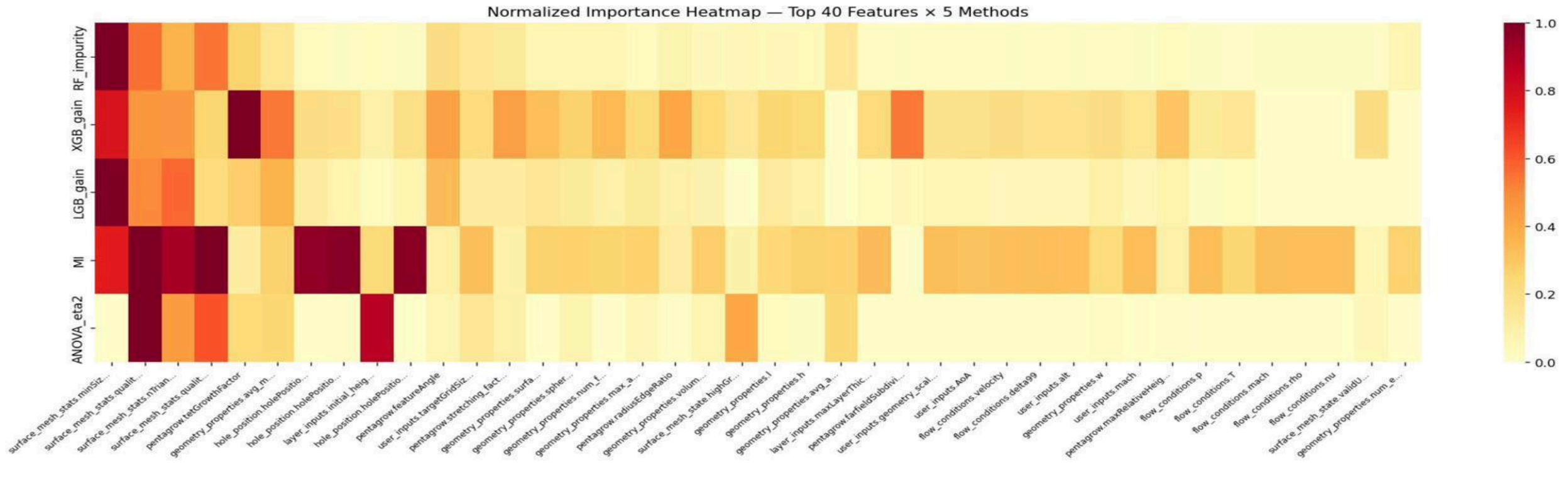

***Figure 12b:*** *Normalized feature importance for the top 40 features across 5 independent methods: Random Forest impurity, XGBoost gain, LightGBM gain, mutual information, and ANOVA eta-squared. Surface-mesh statistics and Pentagrow parameters rank top-tier under every method, while hole-position features carry non-linear signal that appears only under mutual information and ANOVA.*

**What drives successful volume-grid generation.** A data-science study of the full volume-grid campaign across 11 aircraft classes, in which only a small fraction of raw attempts produced a usable grid, identified the levers that the learned setup models then evaluate and rank. The validated grid generator still produces every mesh and the symbolic readiness checks still confirm it.

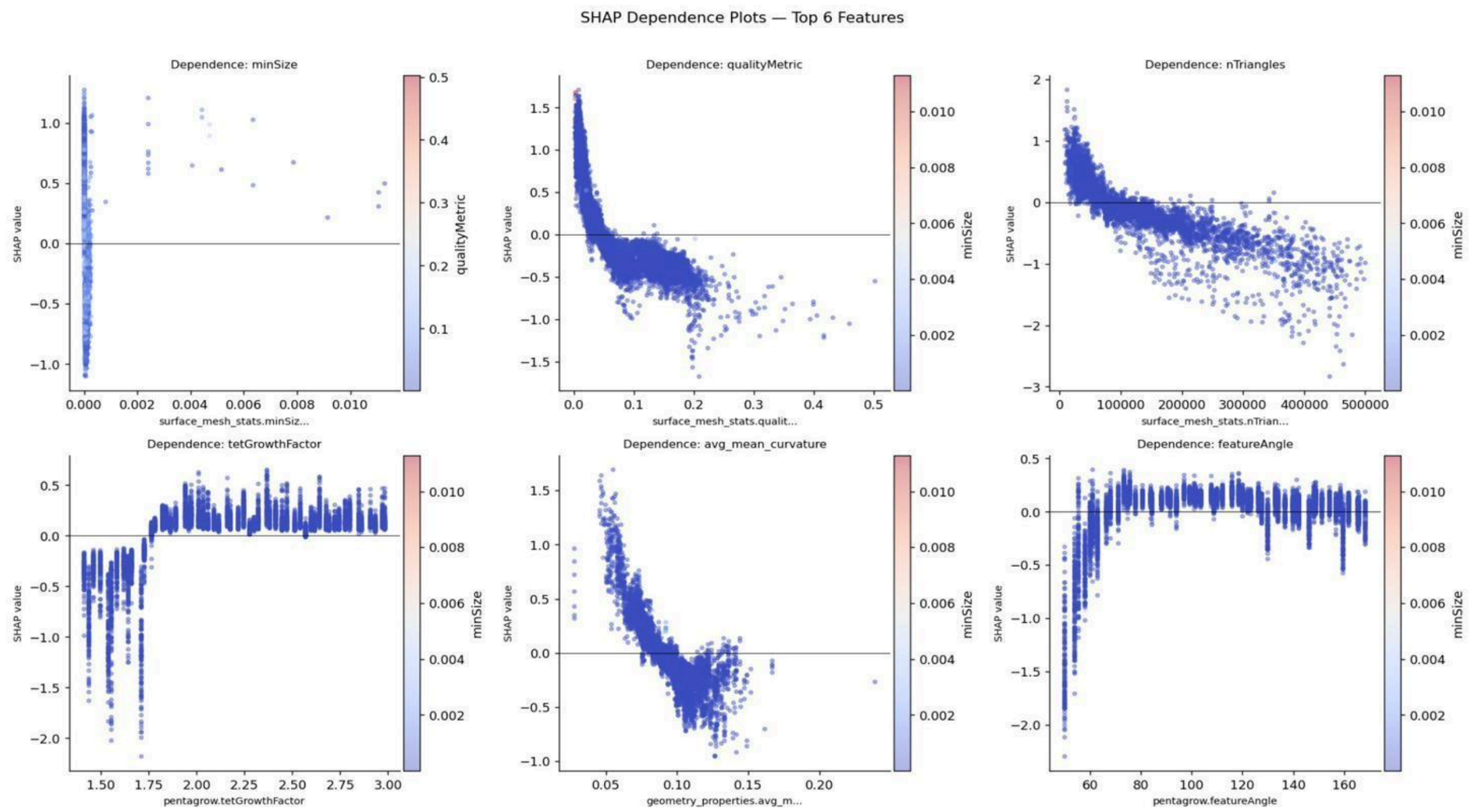

***Figure 12c:*** *SHAP dependence for the 6 most influential features of the grid-success classifier. Very fine surface elements (small minSize) are a precondition, success collapses once surface qualityMetric rises above about 0.05 or the triangle count grows too large, and tetGrowthFactor is stable only above about 1.75.*

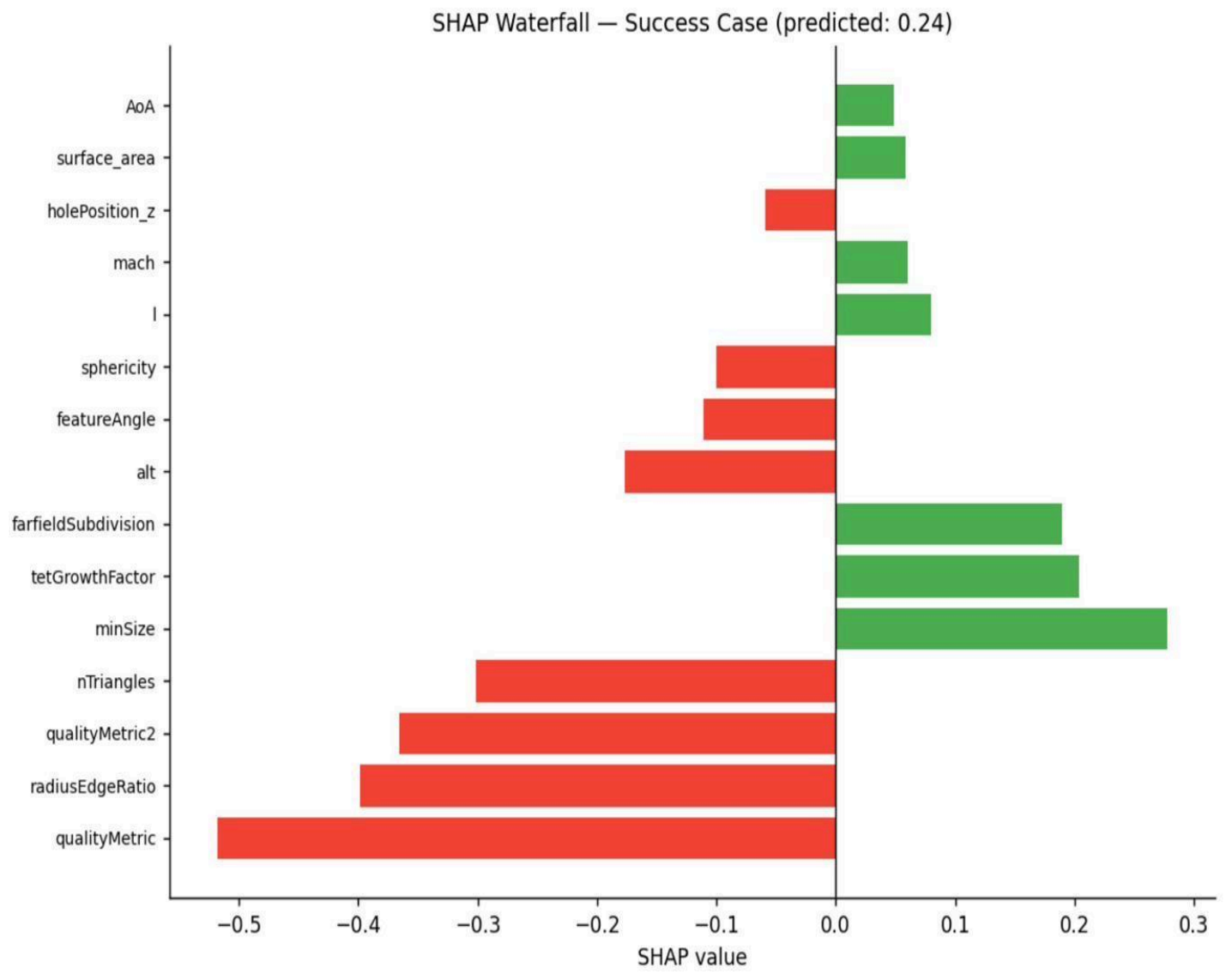

***Figure 12d:*** *SHAP waterfall for a single successful case, predicted success probability 0.24. Surface-mesh and Pentagrow levers carry the case; geometry and flight-condition features add smaller positive amounts; and the negative qualityMetric and radiusEdgeRatio terms are surface-mesh weaknesses the case overcame.*

**What drives SU2 flow-solver convergence.** We applied the Data Factory and Learning Engine pipeline to the original Latin-hypercube SU2 campaign of about 15,000 cases, of which about 13 percent converged. The records were cleaned, normalized, split into reproducible training, validation, and test sets, and used to train multiple AutoML predictors in parallel on Aurora and Frontier. Held-out model performance was then combined with ensemble feature importance and SHAP analysis to identify the drivers of convergence. The result reframes solver setup: user-specified flight conditions, especially Mach number, geometry scale, and altitude, dominate the convergence envelope, while solver settings such as multigrid and CFL-related parameters matter secondarily within that envelope. The learned models therefore set better candidate ranges for the refined run, but SU2 still executes and decides convergence.

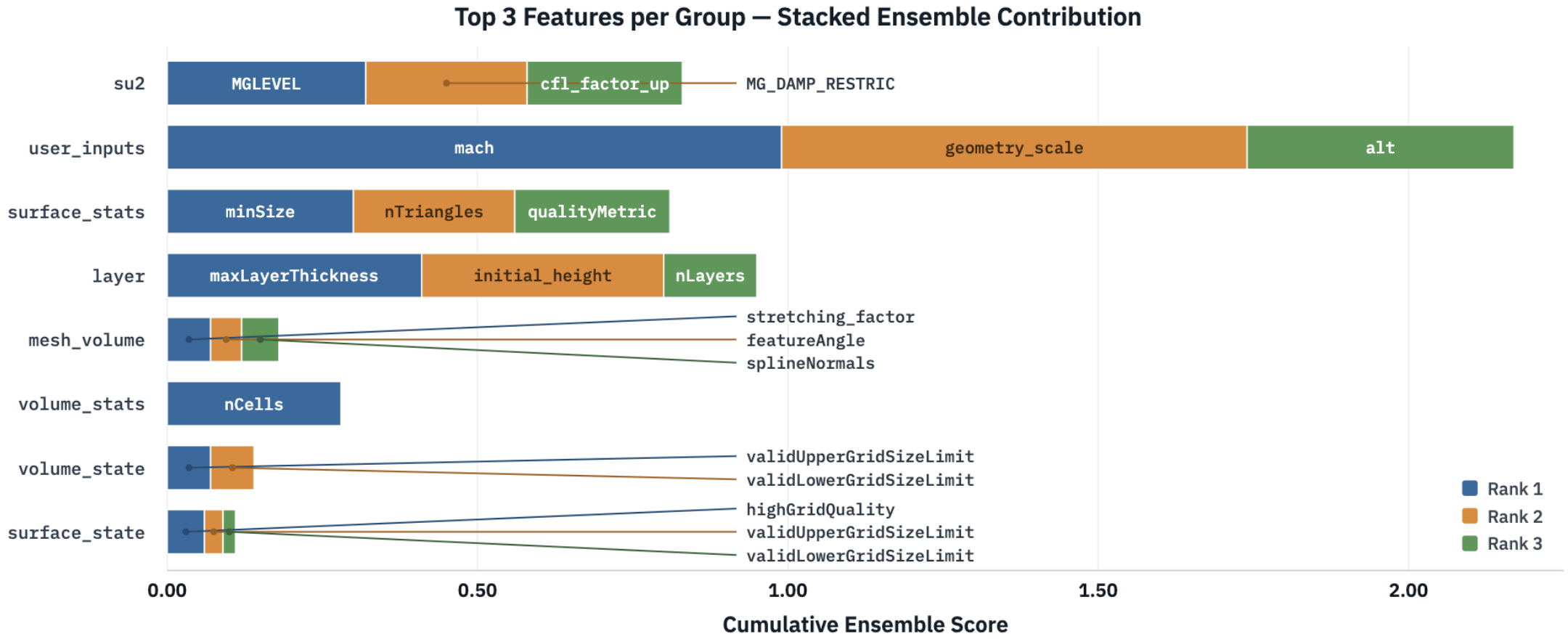

***Figure 12e:*** *Top 3 features per feature group as a stacked ensemble contribution. User inputs (Mach number, geometry scale, and altitude) dominate, because Mach and geometry scale together imply the Reynolds number, the strongest aerodynamic driver. SU2 multigrid and CFL parameters matter second, inside that flight-condition envelope.*

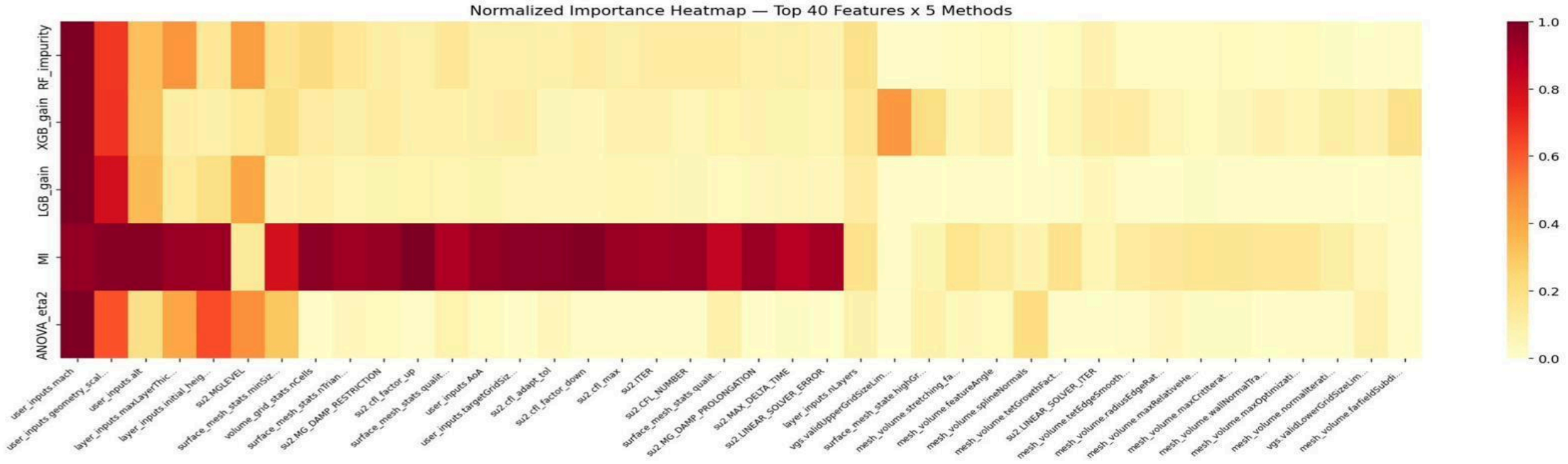

***Figure 12f:*** *Normalized importance for the top 40 features across the same 5 methods. user_inputs.mach and user_inputs.geometry_scale are top-ranked under every method, with altitude and maxLayerThickness completing the top tier and the SU2 multigrid level (MGLEVEL) uniformly important for convergence.*

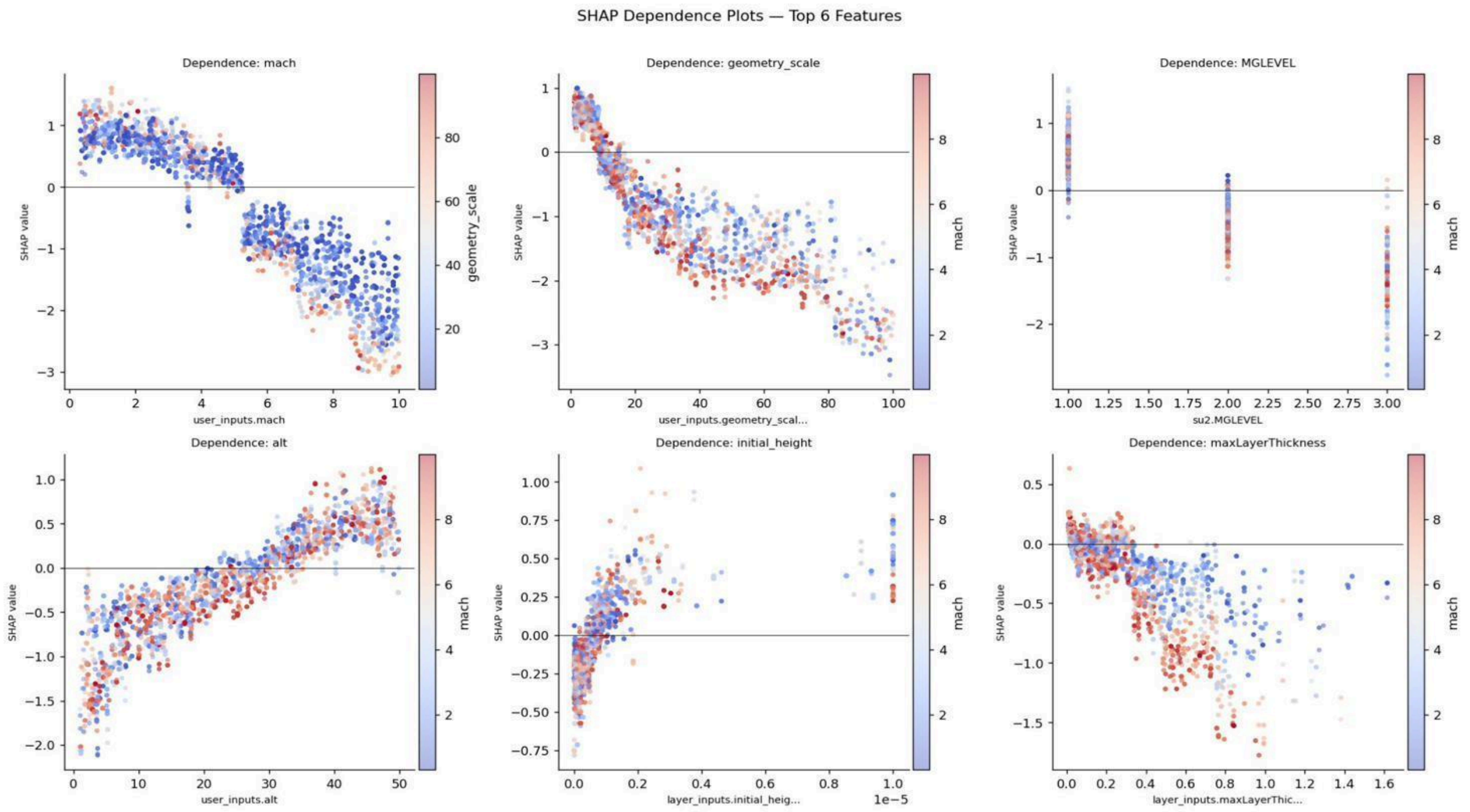

***Figure 12g:*** *SHAP dependence plots for the six most influential convergence features. The model associates lower Mach numbers with improved convergence and shows a sharp degradation above approximately Mach 5, larger geometry scales generally reduce convergence likelihood, MGLEVEL 1 is most favorable while MGLEVEL 3 has the strongest negative contribution, and small boundary-layer initial heights near the 10^-6 range are associated with improved convergence.*

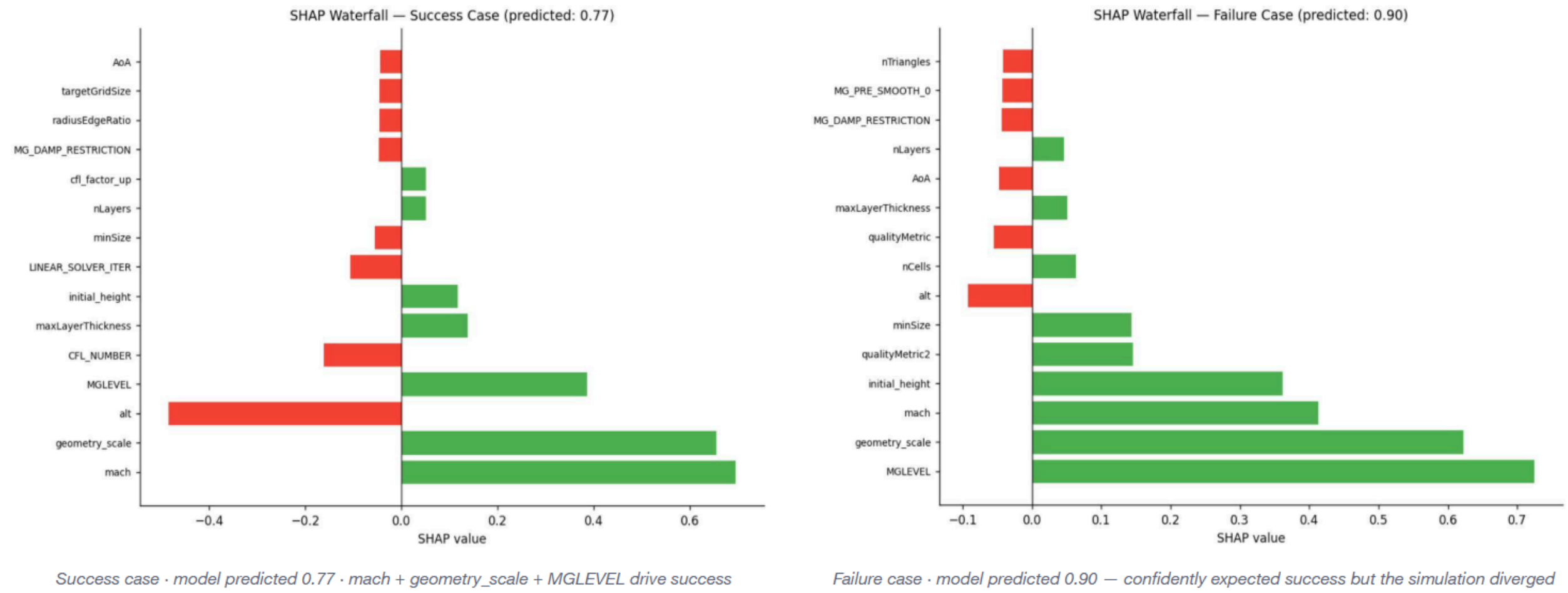

***Figure 12h:*** *SHAP waterfalls for a matched pair of cases. On the left, a correctly predicted success, probability 0.77, driven by favorable Mach, geometry scale, and MGLEVEL. On the right, a failure the model misjudged: it predicted 0.90 success from similarly favorable inputs, yet the simulation diverged. A slice of the parameter envelope is not yet captured by the present features, which is exactly why the validated solver remains the arbiter.*

## 5.4 Optimization, adaptation, and outcomes

The learned predictors are placed inside a multi-objective optimization loop. For a new geometry, a multi-objective genetic optimizer proposes candidate parameter sets and the learned predictor evaluates their expected outcomes, searching for settings that satisfy competing goals: producing a valid mesh, avoiding solver failure, improving convergence, preserving boundary-layer quality, and reducing unnecessary cost (Deb et al., 2002). In this formulation, a successful simulation without errors becomes an explicit optimization target rather than an after-the-fact filter.

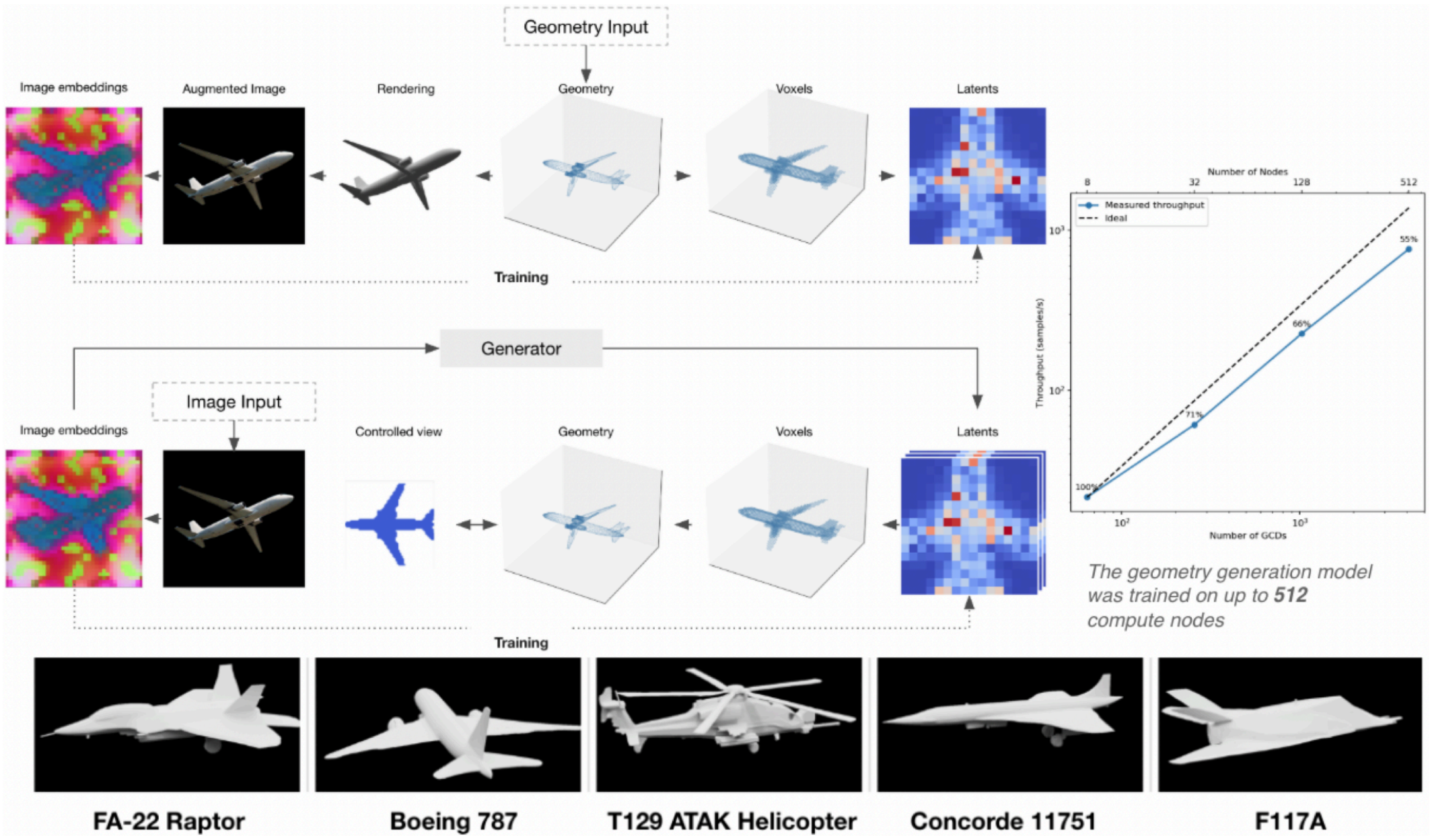

***Figure 13:*** *The generation model maps image embeddings and augmented renderings into meshable vehicle geometries. Representative outputs span 11 aerial vehicle classes, from fixed-wing subsonic and transonic to supersonic and hypersonic, eVTOL, flying-wing and blended-wing-body, delta-wing and canard, high-lift, rotorcraft, lighter-than-air, and unconventional configurations, with 673 geometries reaching successful volume-grid generation.*

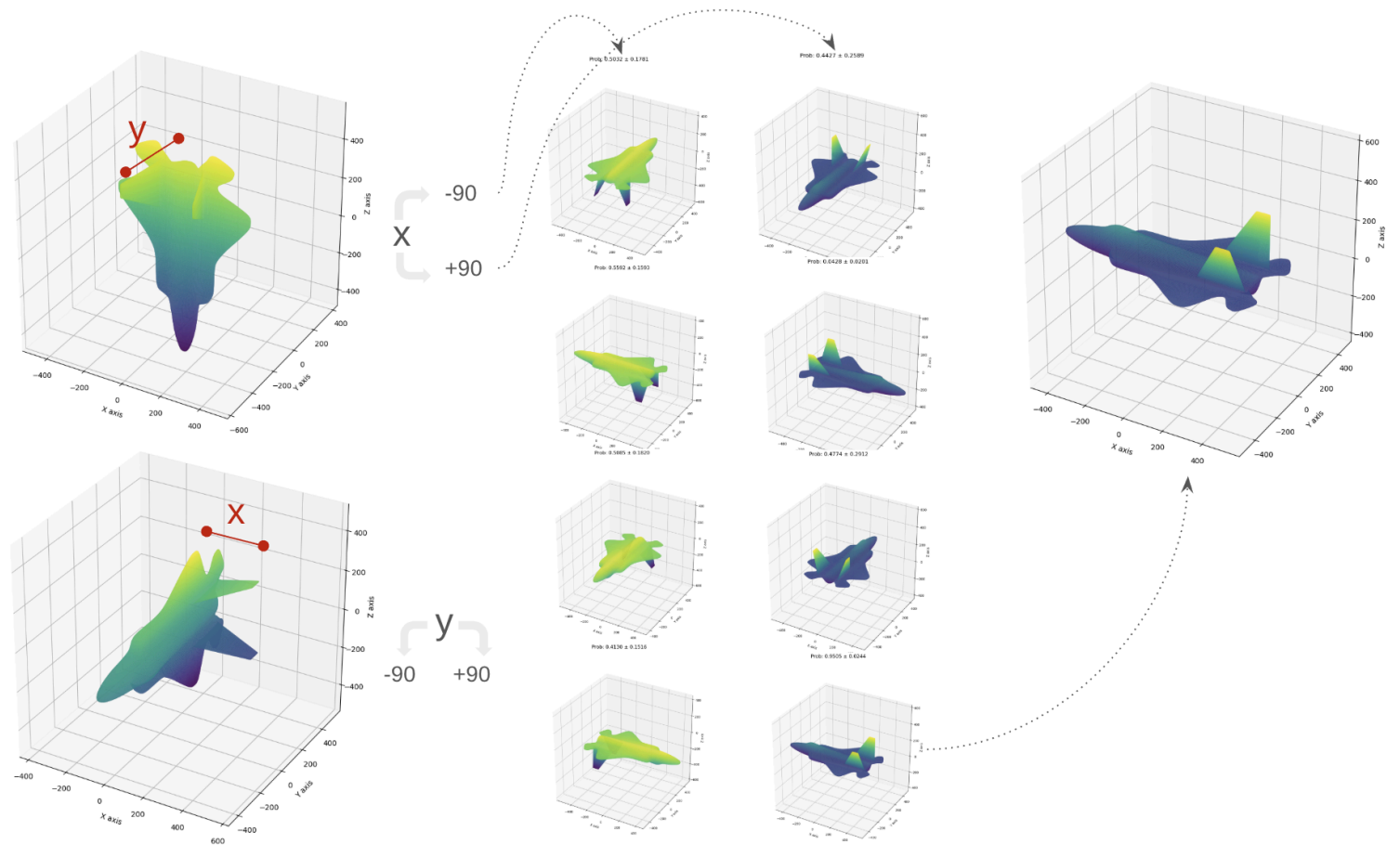

***Figure 14:*** *Generated geometries are rotated into a standardized aerodynamic frame so that inlet direction, lift and drag axes, and solver boundary conditions stay consistent across cases.*

At leadership scale, the search is wide: a containerized grid-quality pipeline runs at scale, in one example configuration across as many as 1,000 nodes at 2 to 3 nodes per case, producing 56 candidate surface meshes and 56 candidate volume-mesh configurations per case, from which the best are selected and re-solved. The adapted grid is then re-solved and reassessed until the grid converges (Figures 13 to 16).

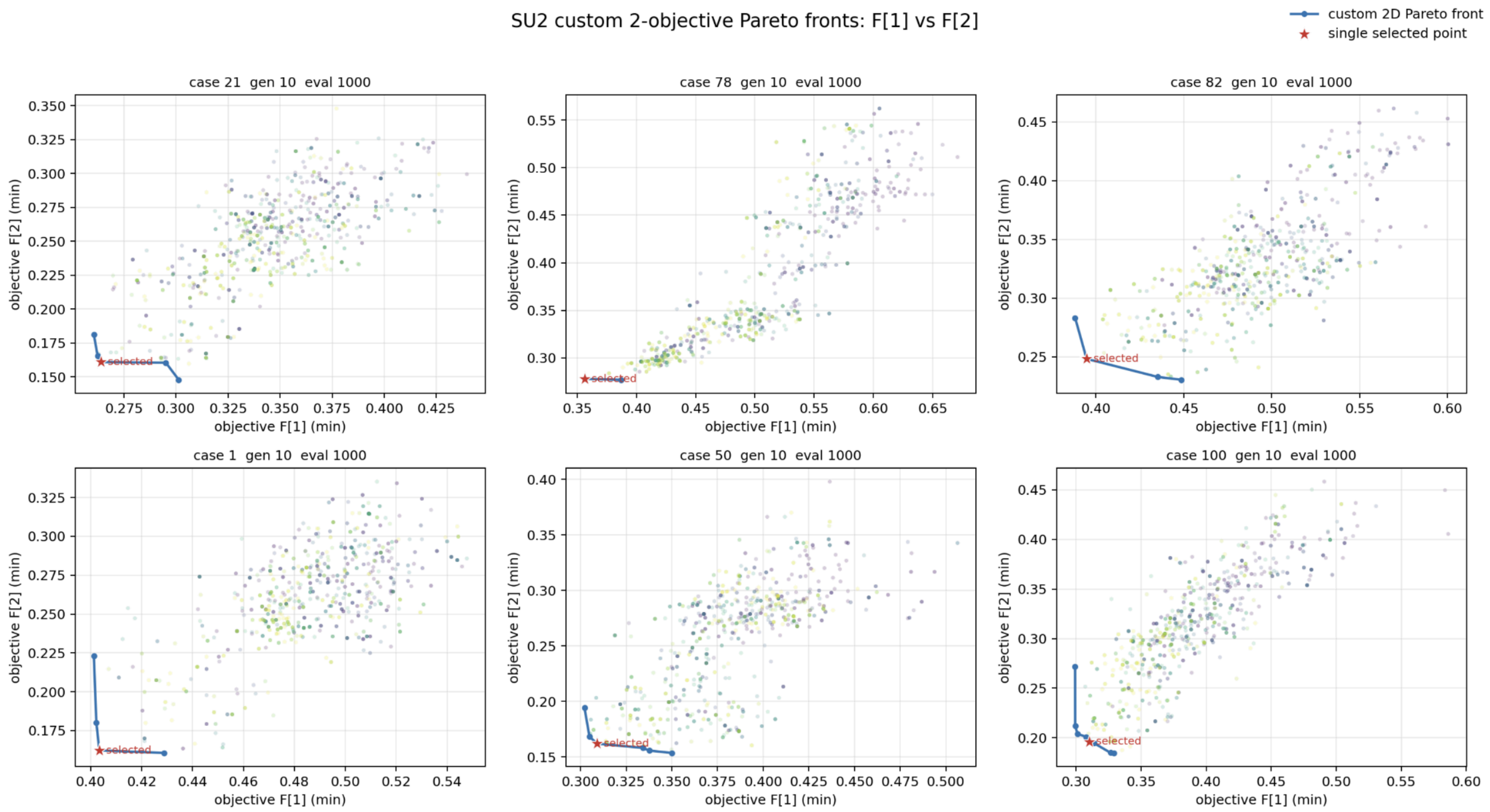

***Figure 15:*** *Candidate solver-parameter configurations evaluated across representative cases under 2 competing objectives, both minimized, with the estimated trade-off front for each case. Successful simulation without errors is treated as an explicit optimization target.*

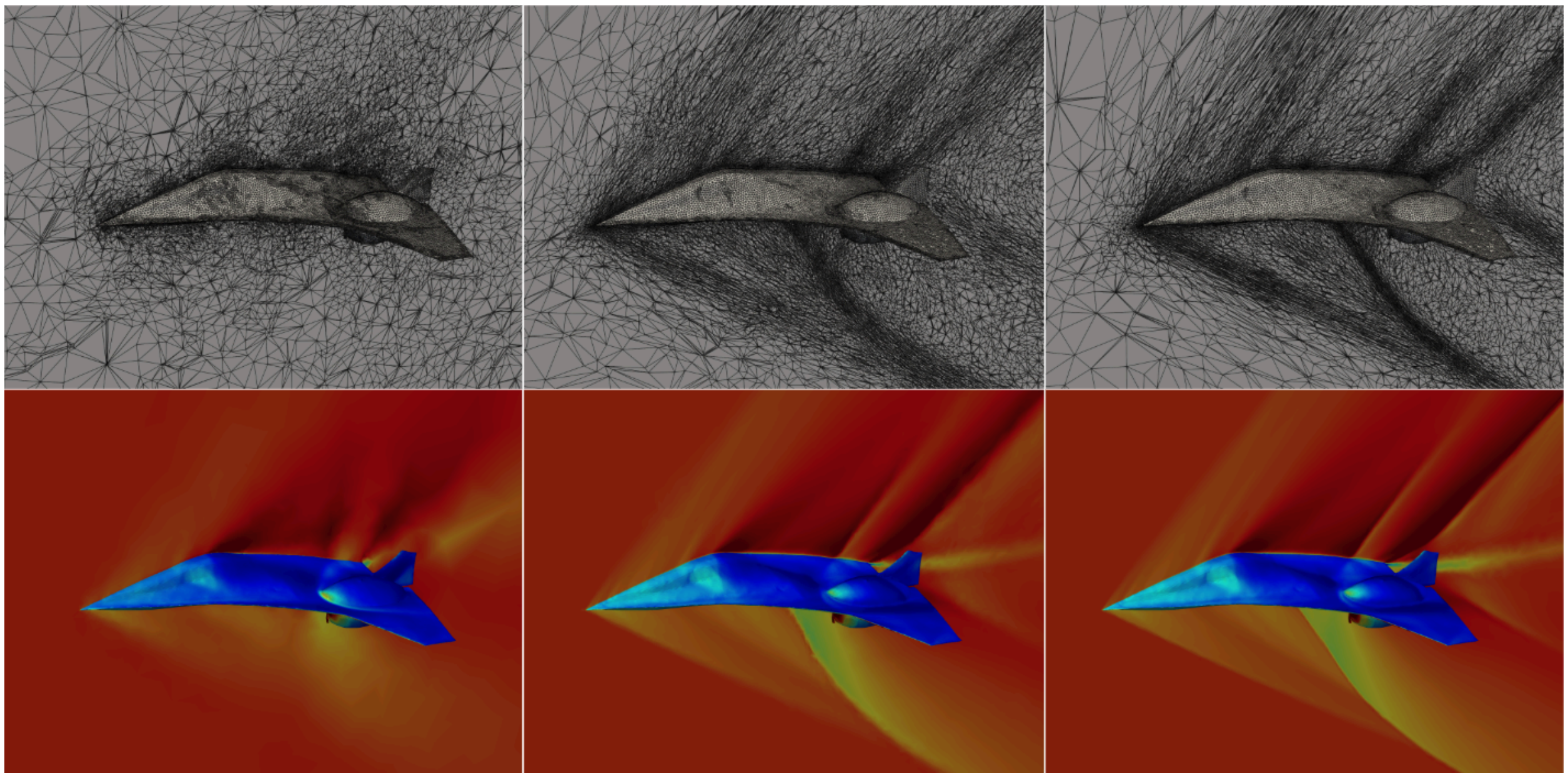

***Figure 16a:*** *Supersonic Spy Plane. Adaptive mesh progression. Evolution of the unstructured volume mesh through two solution-adaptive refinement cycles around the aircraft geometry, from baseline mesh to first and second adaptation cycles.*

**Outcome.** The combination of a more robust grid-generation method and learned parameter prediction substantially raised the yield of usable simulations on previously unseen aircraft geometries. As reported in our peer-reviewed work (Grosvenor, Zemlyansky, et al., 2025), boundary-layer capture, the percentage of wetted surface with y+ at or below 1 and at least eight growth layers, rose from 8% to 98%, and geometry-preparation failures fell from 88% to 2%. These figures are prior peer-reviewed results, and they hold on geometries the system had never seen, which is the relevant test for an automated setup tool. The updated workflow result, reported in Section 10.1, supersedes these as the current headline.

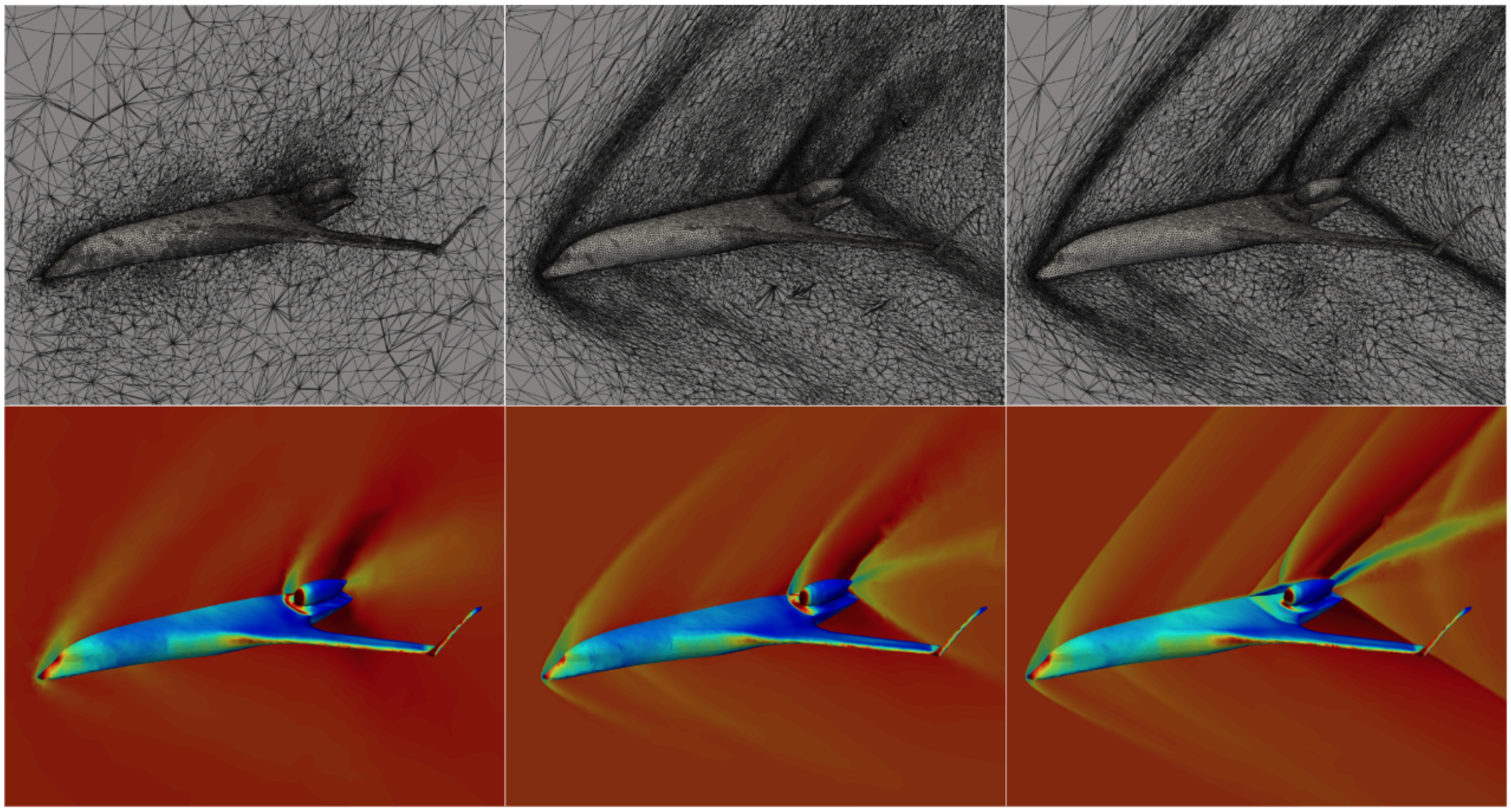

***Figure 16b:*** *Blended Wing Body. Adaptive mesh progression. Evolution of the unstructured volume mesh through two solution-adaptive refinement cycles around the aircraft geometry, from baseline mesh to first and second adaptation cycles.*

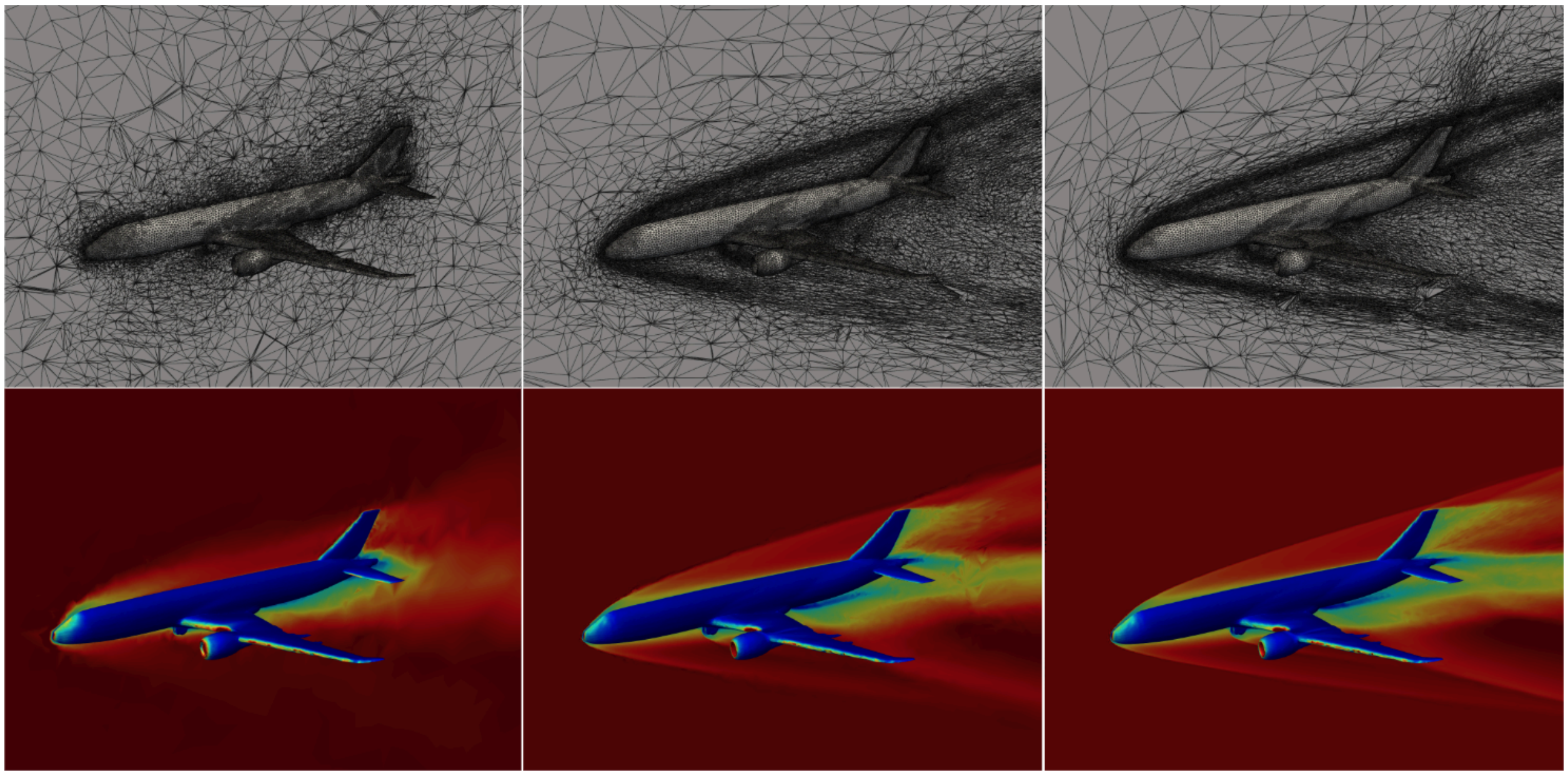

***Figure 16c:*** *Airliner. Adaptive mesh progression. Evolution of the unstructured volume mesh through two solution-adaptive refinement cycles around the aircraft geometry, from baseline mesh to first and second adaptation cycles.*

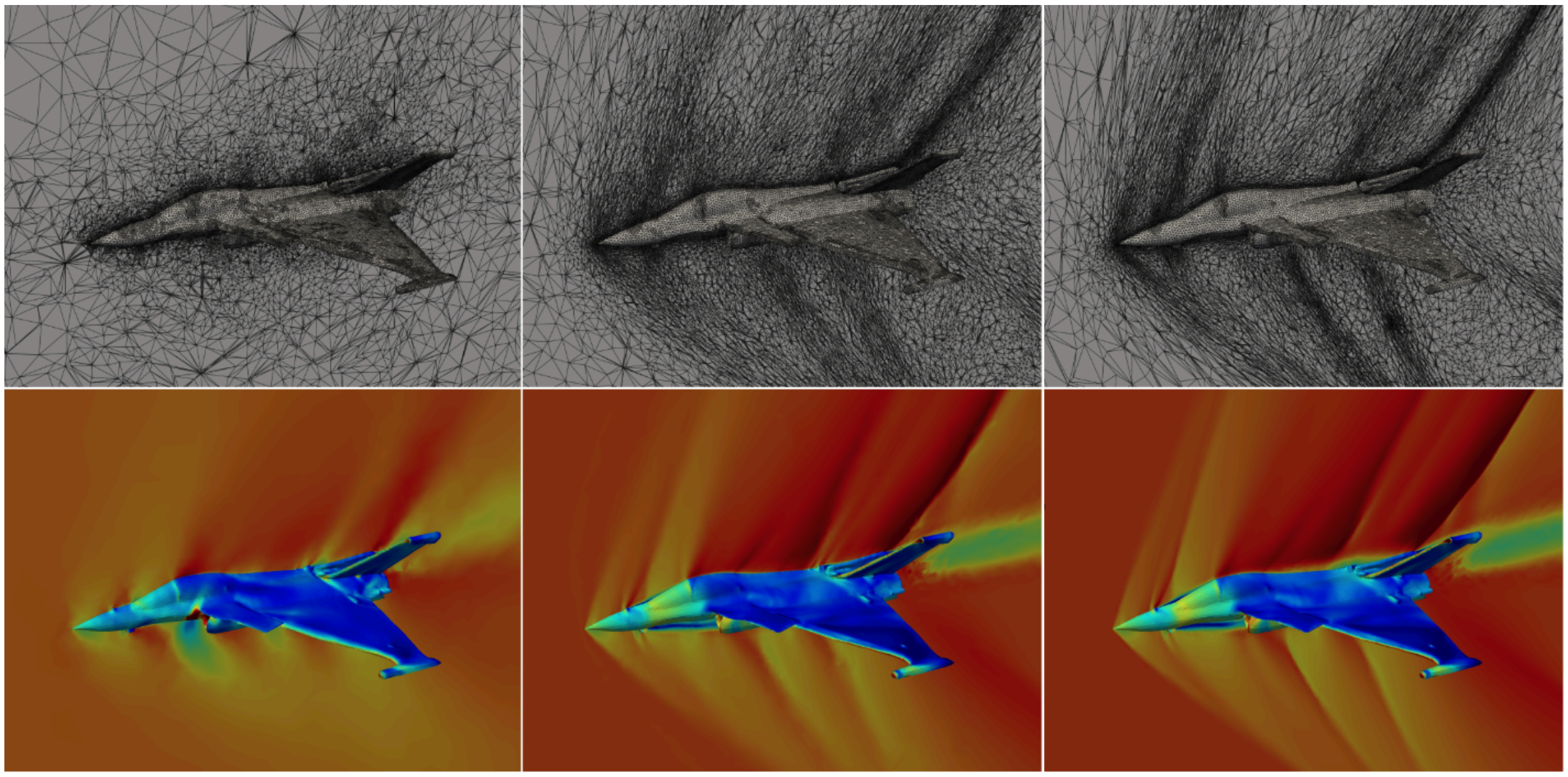

***Figure 16d:*** *Fighter. Adaptive mesh progression. Evolution of the unstructured volume mesh through two solution-adaptive refinement cycles around the aircraft geometry, from baseline mesh to first and second adaptation cycles.*

## 5.5 Software control and visualization in ParaView

Within the CFD workflow described in Section 5.1, ParaView provides an application for inspecting computed fields, constructing derived quantities, and preparing visual output. We evaluate GURU's ability to control these post-processing operations from goal instructions. These experiments assess application control; they do not establish the accuracy of the upstream CFD solution.

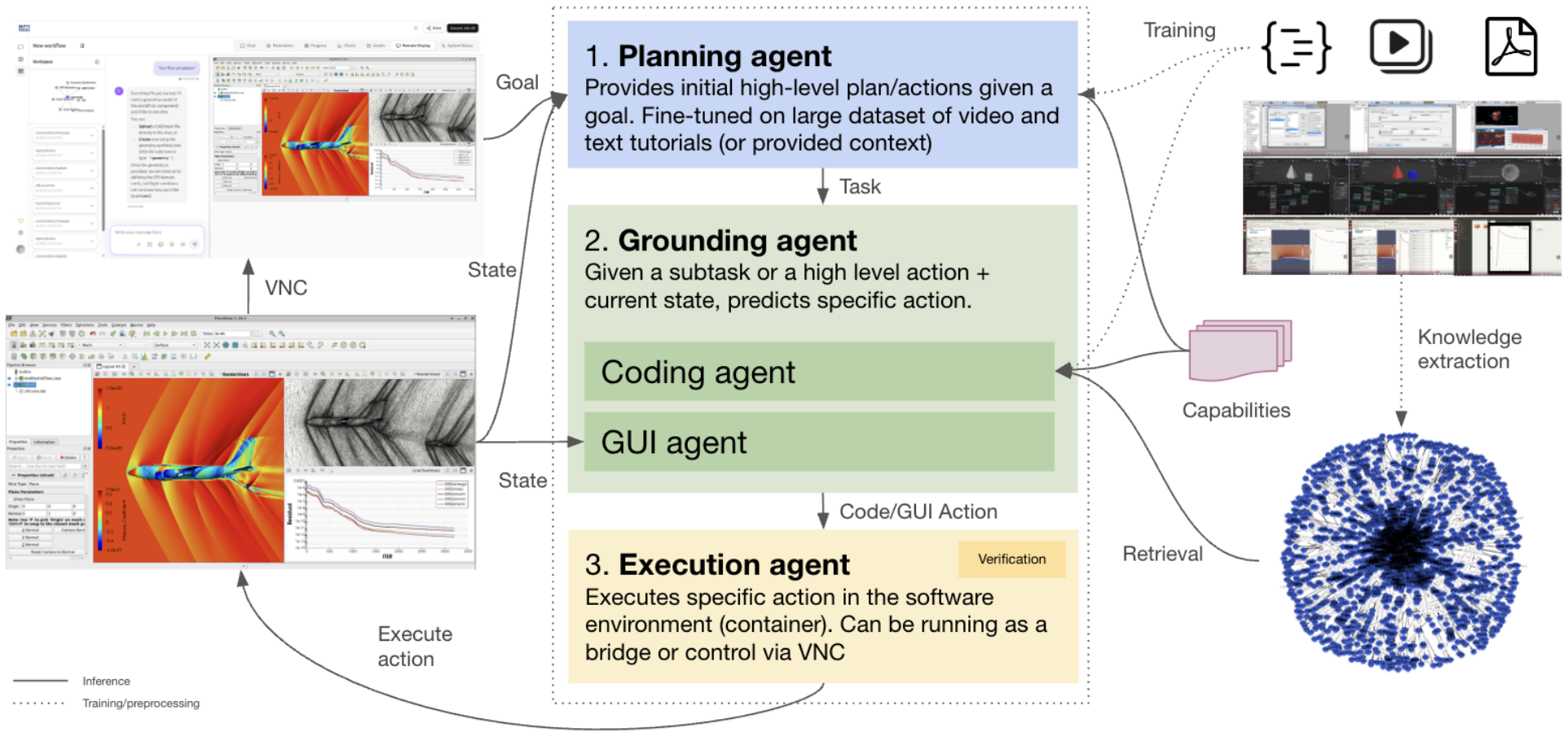

***Figure 17:*** *Software-control architecture and knowledge-learning loop for ParaView. The Planning agent decomposes a user goal into structured actions; the Grounding agent maps each action and the current application state to executable code or GUI operations; and the Execution agent applies the selected operation and returns the resulting state. Tutorials, documents, and interaction trajectories provide knowledge for retrieval and role-specific training.*

We ran three frozen experiments on Aurora. Experiments A and B used the same four goal families (capability-covered, generalization, original knowledge-graph, and curated knowledge-graph tasks), with ten goals per family; experiment C used ten separately checked API-focused goals. Every treatment repeats each goal three times. A varies the information supplied to the base weights; B crosses four adapter configurations with the knowledge graph off or on; C repeats that crossing with capability and helper catalogs hidden. A contains 1,080 attempts, B 960, and C 240. The frozen protocols use one control cycle without generation repair, temperature zero, thinking disabled, a 20,000-token estimated input budget, and a 2,048-token output limit. Results from different jobs are not combined into a synthetic cumulative ladder; Section 5.5.3 reports the latest ablation, a single run over 50 case definitions with three attempts per case and configuration.

### 5.5.1 Adapter fine-tuning

We fine-tune the Planning and Grounding roles independently using LoRA (Hu et al., 2022) on a locally hosted Qwen3.8-27B model, a 27-billion-parameter dense open-weight model (Qwen Team, 2026). The Planner is trained to produce structured action sequences from task context, while the Grounder is trained to convert those actions into executable application code. The base model remains frozen and shared between the two roles; at inference time, each request is routed to its corresponding adapter rather than combining the adapters into a single model. We also explored Decision Transformer training for GUI grounding as an alternative policy-learning approach; for the experiments reported here, we retained role-specific LoRA fine-tuning as the primary adaptation strategy.

Both adapters use rank r = 16, scaling alpha = 32, and dropout 0.05 on attention projections. The supervised datasets contain 6,667 Planning examples and 19,707 Grounding examples, with 5,281 and 15,583 examples in their respective training splits. Training uses completion-only loss, a learning rate of 2e-4, and a maximum sequence length of 9,216 tokens. The evaluated Planner is the step-12 checkpoint from a three-epoch training run, and the Grounder is the step-120 intermediate checkpoint from a one-epoch run. Both adapters were trained on Aurora using distributed data parallelism, with scaling runs extending to 512 nodes (Figure 50d). Training retained consistent convergence at large scale and substantially reduced time to solution as the node count increased.

### 5.5.2 Task coverage and remaining limitations

The ParaView task-coverage benchmark comprises 40 cases operating on a prepared CFD scene containing two aircraft, flow-volume and surface datasets, two spatial views, and existing charts. Requests span object-relative visualization, data processing, presentation, and export; five cases include multiple conversational turns. Table 4 summarizes the task distribution and representative goals.

**Table 4:** ParaView evaluation task coverage. Cases are grouped by their primary requested operation; examples are abbreviated.

| Task category | Task example | Total |
|---|---|---|
| Camera positioning and framing | Zoom to the second aircraft in the mesh view. | 5 |
| Spatial extraction and reference geometry | Create a flow slice 20% of the aircraft's length downstream of its rear bound. | 11 |
| Data transformation, analysis, and inspection | Create and display a 20-bin histogram of Mach. | 8 |
| Visual styling, annotations, and view controls | Set the aircraft surface opacity to 0.5 in the flow view. | 12 |
| Pipeline organization | Rename the surface source while preserving its downstream connections. | 1 |
| Image and data export | Save the flow view as a 1200 × 600 PNG. | 3 |
| **Total** | | **40** |

Performance remains uneven across this task space: several goals fail even when relevant graph context is retrieved, showing that knowledge availability does not by itself guarantee correct execution. Each attempt is scored pass or fail under the evaluation protocol of its run; because scoring conventions changed between earlier experiments, comparisons are made only within a single run. The evaluation is also limited by the small development benchmark, repeated use of the same goals, and the use of structured application state rather than the intended deployment interface, which operates only through the remote display (the Remote Framebuffer protocol used by VNC). The results should therefore be read as evidence of improved software control under the evaluated protocol rather than as production-level robustness.

### 5.5.3 Effect of state, API, and procedural knowledge

The latest ablation evaluates 50 ParaView case definitions, with three attempts per case. With base weights fixed, goal-only and state-only prompts yield no passing attempts without retrieval. Adding live API inventory to state gives 56 of 150 passes (37.3%); adding capability and helper catalogs instead gives 42 of 150 (28.0%). Supplying state, catalogs, and API together raises success to 82 of 150 (54.7%). These treatments change the available information, while execution permissions remain unchanged.

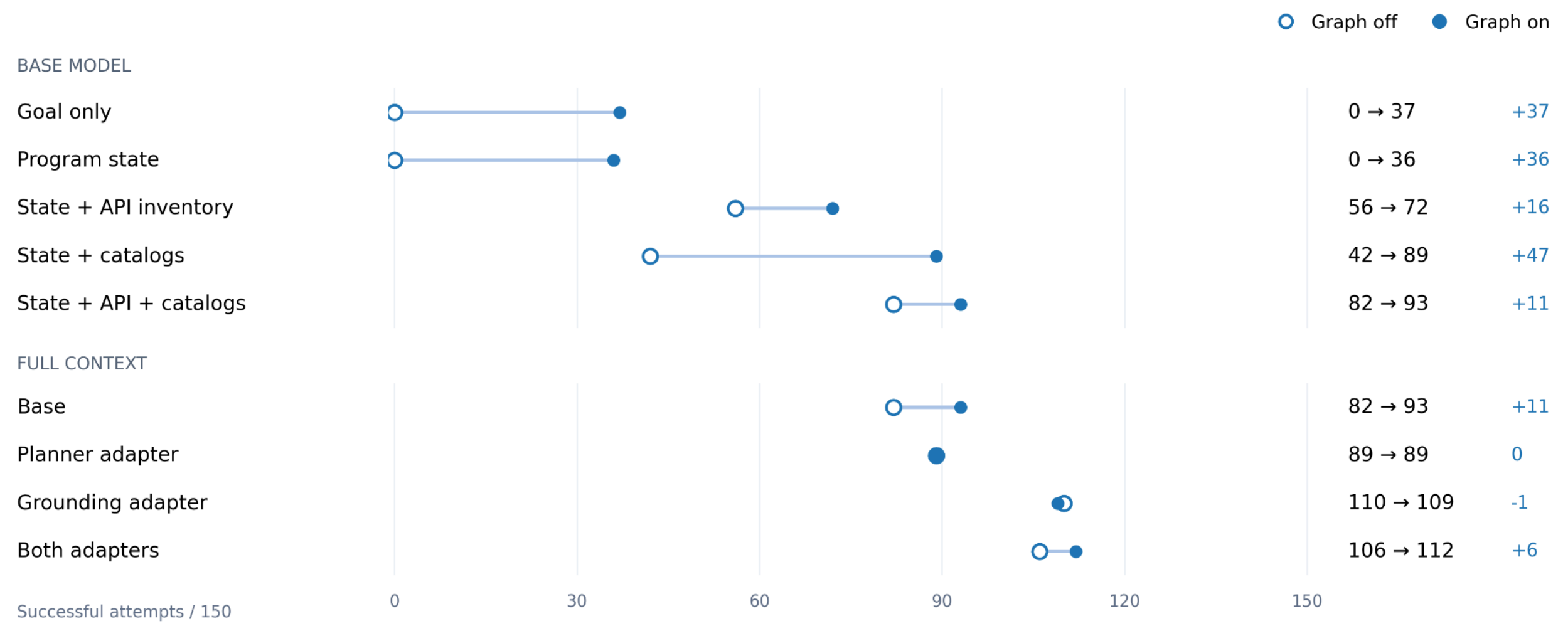

***Figure 18:*** *Ablation of context, retrieval, and learned adaptation for ParaView software control (successful attempts out of 150 per configuration). Explicit program state, API information, and capability catalogs improve base-model performance, while graph retrieval is most valuable when supplied context is limited. Under full context (state, API inventory, and catalogs), the role-specific Planning and Grounding adapters provide additional gains, with retrieval and adaptation showing complementary effects in some configurations.*

The knowledge graph helps most when other context is sparse. It raises goal-only success to 37 of 150 (24.7%), state-only success to 36 of 150 (24.0%), and state-plus-catalog success from 42 to 89 of 150 (59.3%). With full context, the gain is smaller: 82 to 93 of 150 (62.0%). API facts can also arrive through the graph when live inventory is disabled, allowing explicit knowledge to compensate for some missing local context.

With full context and no retrieval, the base model, Planner adapter, Grounding adapter, and both adapters pass 82, 89, 110, and 106 of 150 attempts. With retrieval, they pass 93, 89, 109, and 112. Both adapters plus the graph therefore improve over the full-context base without retrieval by 20 percentage points, from 54.7% to 74.7%. Most of that gain comes from the Grounding adapter, which alone reaches 110 of 150 (73.3%) without retrieval. Retrieval adds six passes beyond the dual-adapter result, provides no aggregate gain to Planner adaptation, and slightly reduces Grounding-only success; at 50 case definitions, differences of a few attempts are not statistically distinguishable. The benefits are complementary in some configurations rather than uniformly additive.

The ablations use the knowledge graph described in Section 4.4.1. Saved observations were reused; automatic semantic matching and application-scope filtering were disabled. These are attempt-level scores over overlapping tasks under native structured-state access, not measured remote-display-only performance.

At the task level, this progression expands the range of completed visualization and post-processing operations. Graph retrieval with goal-only prompts enables source renaming, visibility and opacity changes, fixed-range Mach thresholding, and tetrahedralization. Full local context without retrieval enables component-focused views, aircraft-relative flow slices, fixed-isovalue contours, and Mach histograms. Grounding adaptation adds successful text annotations, screenshot export, and CSV export of sampled line data. The combined Planning, Grounding, and retrieval configuration also completes cross-view slice reproduction and center-plane clipping. Each example passed all three attempts in its stated configuration; the aggregate gains do not mean that every previously successful task remains successful as components are added.

Together, these results provide a software-control example of GURU's hierarchical neuro-symbolic platform, connecting the training factory in Section 4 to the agent organization and Common Internal Language in Section 3. Knowledge extracted from documents, tutorials, and exploration supplies explicit procedures and API contracts, while role-specific adapters teach agents to turn goals into plans and executable operations. Planning and Grounding coordinate this translation, and execution checks distinguish generated commands from achieved results. Within the CFD application, this approach addresses post-processing: an operator can request a flow slice, derived field, or visualization while the system selects and executes the application commands. The results support GURU's central design principle that autonomy emerges from the composition of learned specialists, explicit knowledge, and external checks, while the physical validity of the underlying simulation remains with the validated solver and engineering acceptance criteria.

### 5.6 Open and production solvers

The results reported in this paper use an open-source CFD solver so that every setup decision, readiness check, and validation criterion can be published and reproduced. The workflow is not tied to that solver. Geometry conditioning, mesh-quality assessment, boundary-condition assignment, solver-parameter selection, and post-run validation are represented as structured workflow steps rather than as solver-specific scripts, and the same agents already drive a production air-vehicle solver from the HPCMP CREATE suite in sponsored engineering work. Those production runs are outside the scope of this paper. They establish that adapting the workflow to a production solver consists of training the setup and post-processing agents on that solver's documentation, tutorials, and example cases, while the readiness checks and the solver-confirmed validation remain unchanged. The role is the same in both settings: GURU reduces manual setup effort and carries provenance, and the solver produces the validated physics that decides each case. This matches the trajectory the CFD Vision 2030 Roadmap projects, in which the first durable impact of agentic workflows appears in case setup, error recovery, post-processing, and provenance capture rather than in replacing the solver (Wissink et al., 2026).

## 6. Application II: Space Domain Awareness

Space domain awareness is a second demanding application, because it requires continuous reasoning over large, irregular, and safety-critical time series. The orbital environment contains tens of thousands of tracked objects, and each must be watched for maneuvers, stability changes, departures from its orbit family, and possible close approaches. Traditional screening propagates and compares objects pairwise across the catalog, which becomes expensive as the catalog grows. GURU treats the problem as representation learning and workflow automation: historical orbital records become compact state embeddings, a learned world model forecasts future behavior, and specialized agents coordinate ingestion, screening, anomaly detection, and operator reporting. This application is delivered through OrbitGuard, the space-operations service built on GURU, and parts of this work were presented publicly at the 2026 Department of the Air Force Modeling, Simulation and Analytics Summit (Grosvenor, 2026).

## 6.1 Orbital data and the JEPA world model

Two-line element records from the public catalog (U.S. Space Command, 2026b) are converted to Earth-centered inertial (TEME) state-vector sequences by standard SGP4 propagation (Vallado et al., 2006), then normalized by orbital regime. Each object history becomes a multivariate time series of position, velocity, and the time gap between observations. The time gap matters, because orbital observations are irregularly sampled and the interval itself carries information about uncertainty and trajectory evolution. The Data Factory turns these records into context and target windows from which the Learning Engine trains a JEPA world model. The encoder maps a window of past state into a latent embedding, and the predictor matches the embedding of a future window, conditioned on the prediction horizon. This is well suited to orbital data, because raw state vectors are dominated by deterministic propagation, while the operationally meaningful signals, maneuver signatures, anomalous drift, and the evolution of a close approach, appear as smaller structured deviations. By predicting in embedding space rather than reconstructing raw trajectories, the world model learns a representation organized around behavior rather than coordinate recovery.

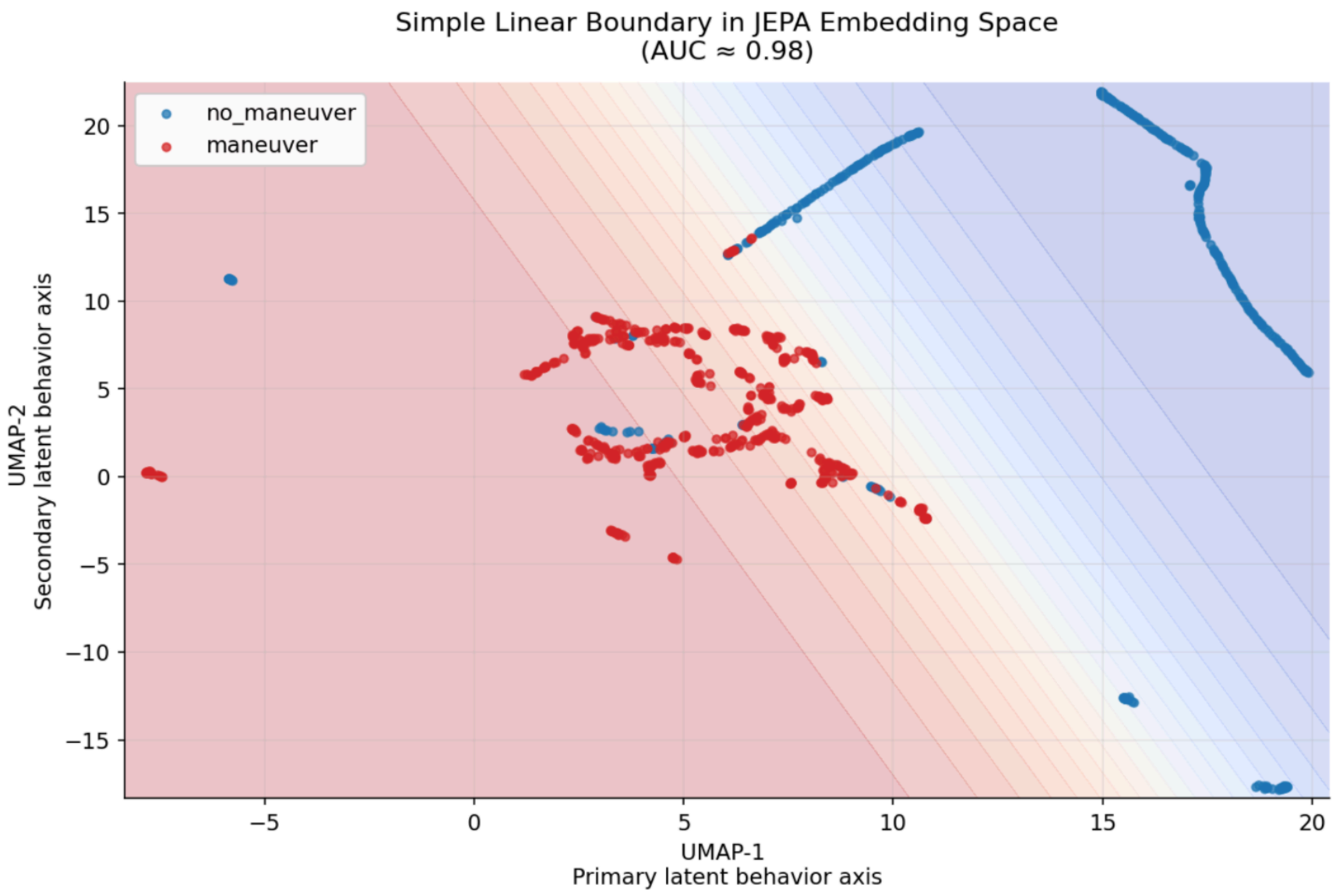

***Figure 19:*** *2D projection of JEPA world-model embeddings for orbital time-series windows. Non-maneuver windows and maneuver windows separate cleanly, and a simple linear boundary achieves an area under the curve near 0.98 with no labeled supervision during pre-training and no deep downstream classifier fit to the embedding.*

**A representation that organizes behavior on its own.** Even when trained only to forecast short horizons, the learned representation is strikingly well organized. A two-dimensional UMAP projection of the embeddings (McInnes et al., 2018) separates maneuver windows from non-maneuver windows so cleanly that a single linear boundary in the projection separates the two classes with an area under the curve near 0.98 across 14,710 objects, consistent with the maneuver-detection result reported in earlier peer-reviewed work (Grosvenor, Wahab, et al., 2025). This separation appears without any labeled supervision during pre-training and without a deep downstream classifier fit to the embedding, which is evidence that the self-supervised future-prediction objective naturally organizes the embedding space around the orbital changes that matter for space operations. The best encoder, with about 55 million parameters trained on tens of millions of state windows, produced this result, and the same embedding supports maneuver detection, out-of-family classification, and the conjunction-risk scoring described below (Figure 19).

## 6.2 Horizon-conditioned forecasting and conjunction screening

Because the predictor is conditioned on the prediction horizon, the same world model can forecast where an object is headed over the window operators care about. The earlier model described only the next observation. The extended model takes a look-ahead interval, up to 72 hours for operational screening and at most 7 days, from as much as 28 days of history, and predicts the embedding of the future trajectory rather than a single point. This is what conjunction screening needs, because the operational question is where two objects will be over the next 72 hours, not at their next update. It also enables retrospective discovery of patterns that were deliberately staggered in time to avoid notice, by sweeping the look-ahead forward and backward through the historical record (Figure 21).

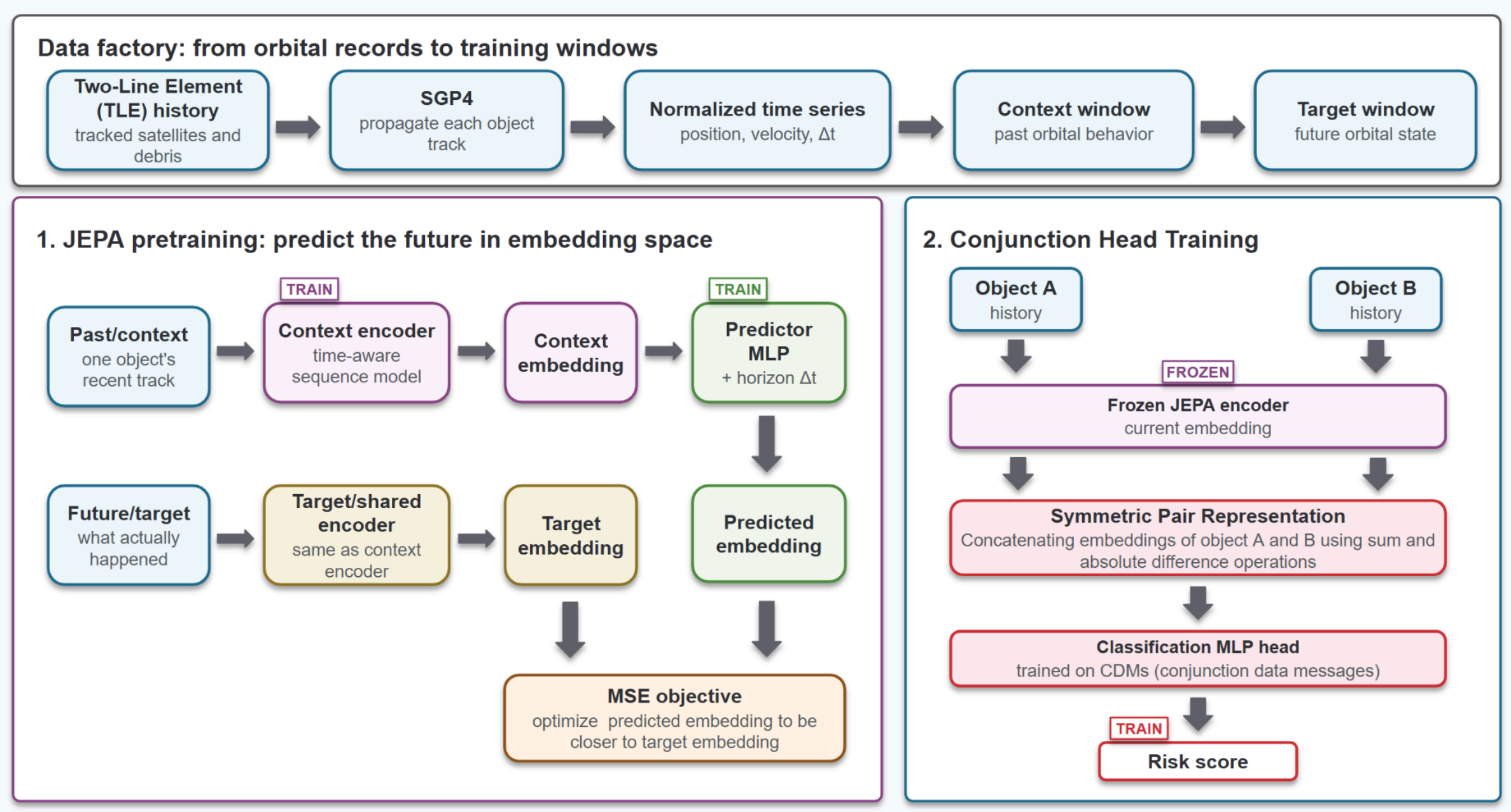

***Figure 20:*** *Two-phase design. In pre-training, the world model learns from context and target windows split from up to 28 days of state-vector history, with a slow exponential-moving-average target encoder and a Gaussian anti-collapse regularizer optimized under an MSE objective in embedding space. In the conjunction stage, the frozen encoder produces embeddings for the history of two objects, which are combined into a symmetric pair representation and read by a conjunction head trained on public conjunction-message labels to output a collision risk score.*

Conjunction screening at catalog scale is organized as a five-stage hierarchy that narrows the whole catalog to a short, ranked list, with each stage deliberately cheaper than the next so that easy non-threats are discarded early and expensive physics is spent only on the few pairs that survive. The catalog holds about 34,420 objects, and screening each active satellite against everything else implies on the order of 500 million candidate pairs, which must be redone as orbits change. Stage 0 flags objects of interest from their anomaly signals. Stage 1 applies geometric pre-filters, an altitude-band test and an orbit-crossing test, that remove pairs which can never meet before anything is propagated. Stage 2 propagates each survivor once across a 7-day window and caches the track, so a pair never triggers redundant computation, using a parallel propagation method that follows established conjunction-screening practice (Healy, 1995; Kelso and Alfano, 2005; Stevenson et al., 2023). Stage 3 scores risk by solving for the moment of closest approach, the miss distance, and the probability of collision (National Aeronautics and Space Administration, 2023). Stage 4, a fusion and ranking agent, is designed to fuse anomaly evidence, risk scores, and high-resolution propagation of the most threatening pairs into a single ranked, decision-ready alert list, with a closed feedback loop that re-orders the queue when a newly suspicious object appears; at the time of writing, Stages 0 to 2 run, Stage 3 is in refinement, and Stage 4 is next (Figure 22a). The risk model is trained on labeled examples built from the public conjunction record, 3,898 actual reportable events retrieved from Space-Track (U.S. Space Command, 2026a), expanded with generated examples that apply the same official reporting criteria. Because the pipeline never builds or stores the full pairwise graph, its cost tracks the small number of pairs that survive, which is what allows it to scale toward a catalog of 100,000 objects without a step change in hardware (Figures 20 and 22).

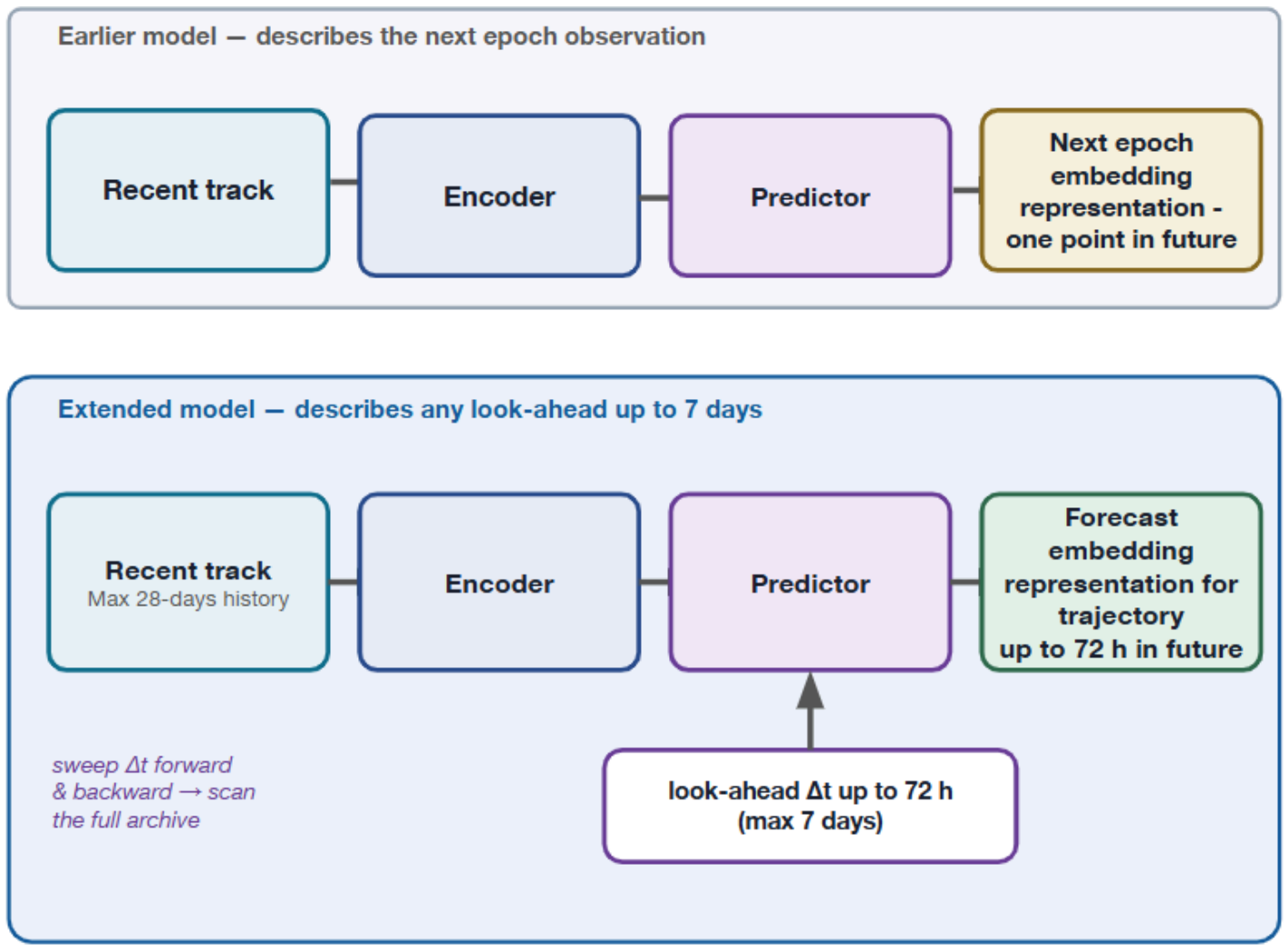

***Figure 21:*** *The horizon-aware forecaster that produces the forward tracks. The extended model takes a look-ahead interval, up to 72 hours (at most 7 days), from as much as 28 days of history, and predicts the embedding of the whole future trajectory rather than a single point. Sweeping the look-ahead forward and backward turns the same model into a scanner across the historical archive for behavior staggered in time to avoid notice. The forecaster is a joint-embedding predictive architecture (LeCun, 2022; Assran et al., 2023).*

The world model contributes to conjunction screening in two distinct ways. Its encoder, trained only to forecast, supplies the object embeddings that the conjunction risk head reads (Figure 20); the forecasting objective is what organizes those embeddings around behavior. Its horizon-conditioned predictor supplies forward embeddings and prediction-error signals that flag objects whose behavior departs from forecast (Stage 0). Miss distance, time of closest approach, and collision probability come from physics-based propagation, not from the embedding.

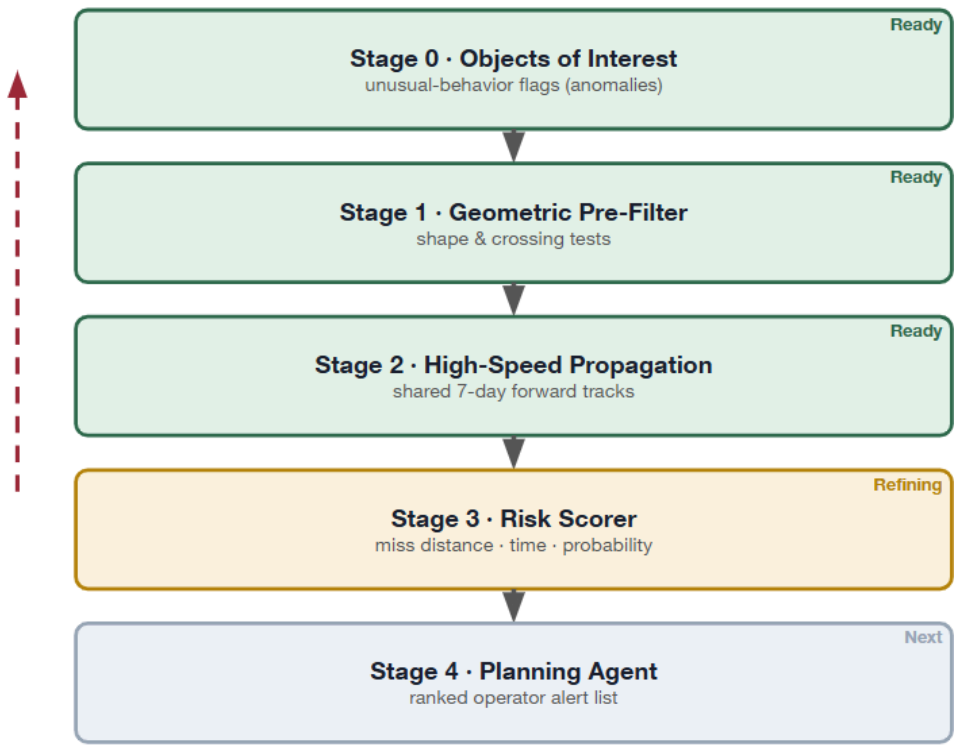

***Figure 22a:*** *The 5-stage hierarchy. Stage 0 flags objects of interest from anomaly signals, Stage 1 applies geometric pre-filters, Stage 2 propagates survivors once and shares the 7-day forward tracks, Stage 3 scores risk from miss distance, time, and collision probability, and Stage 4 fuses the evidence into a ranked operator alert list. The status labels reflect maturity at the June 2026 review: Stages 0 to 2 built and running, the risk scorer in refinement, the fusion and ranking agent next.*

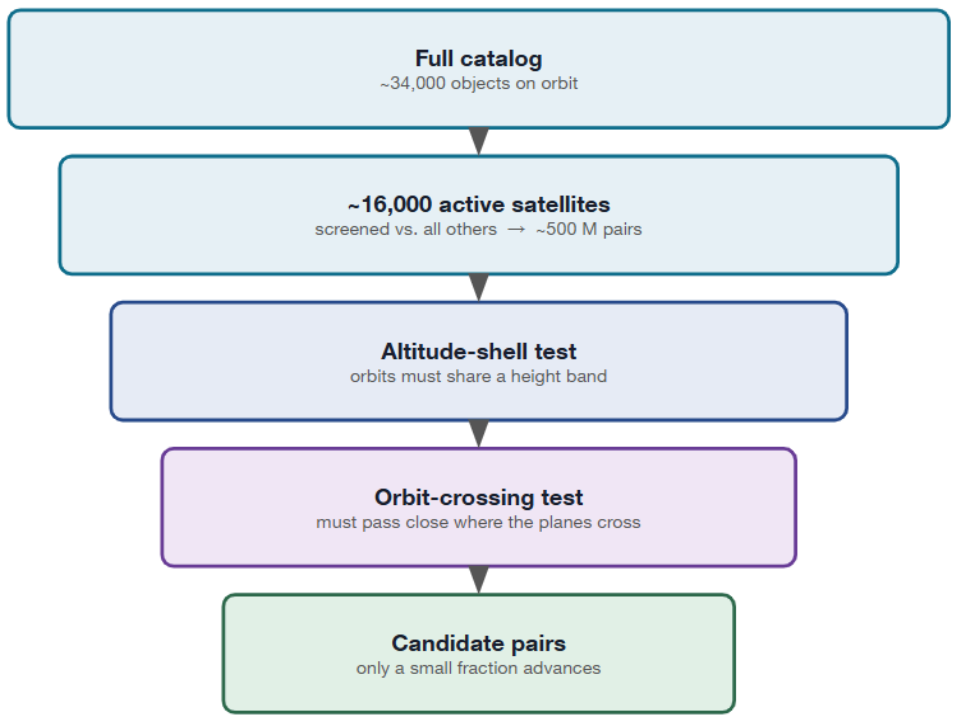

***Figure 22b:*** *Narrowing about 500 million pairs before propagating a single orbit. The catalog holds roughly 34,420 objects, of which about 15,879 are active satellites screened against everything else. An altitude-shell test discards pairs whose orbits do not share a height band, and an orbit-crossing test drops pairs that stay far apart where their planes cross, so only a small fraction survives to propagation. The method follows the SOCRATES screening approach (Kelso and Alfano, 2005) and the staged-filter baselines of Stevenson et al. (2023).*

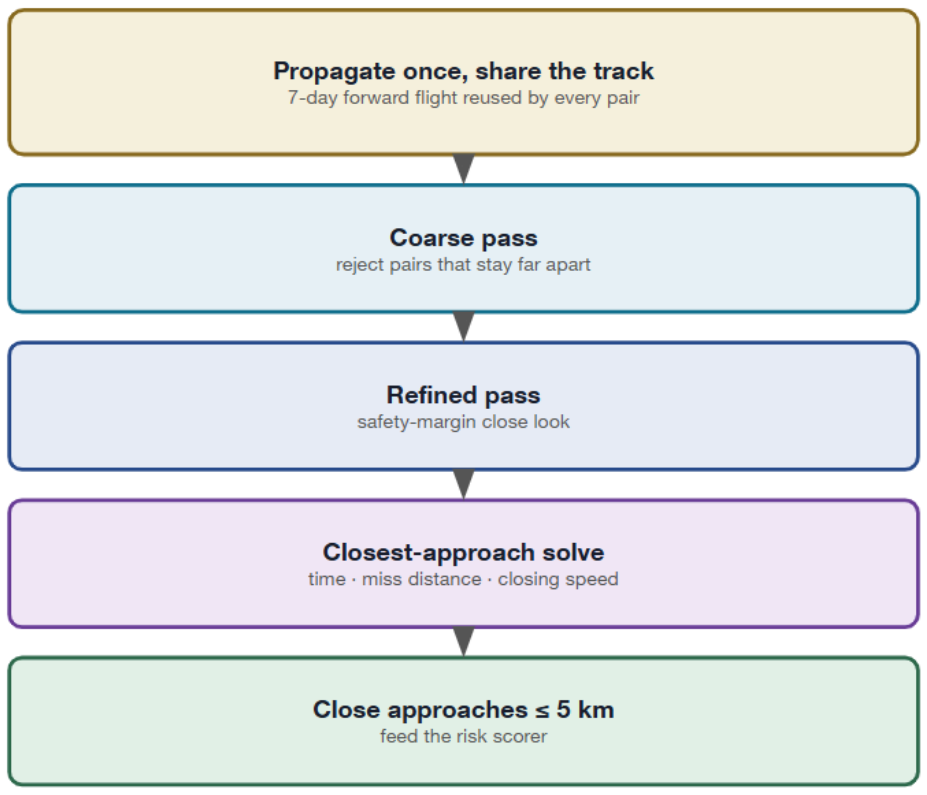

***Figure 22c:*** *From surviving pairs to real close approaches. Each surviving object is propagated once across a 7-day window and its track is held in a shared cache, so no pair triggers redundant computation. A coarse pass rejects pairs that stay far apart and a refined pass examines the rest, both carrying a deliberate physical safety margin. For each remaining pair the system solves for the moment*

*of closest approach, the miss distance, and the closing speed, keeping approaches within a 5-kilometer screening distance. The coarse-to-fine sieve follows Healy (1995) and the ESA smart-sieve (Alarcon-Rodriguez et al., 2002).*

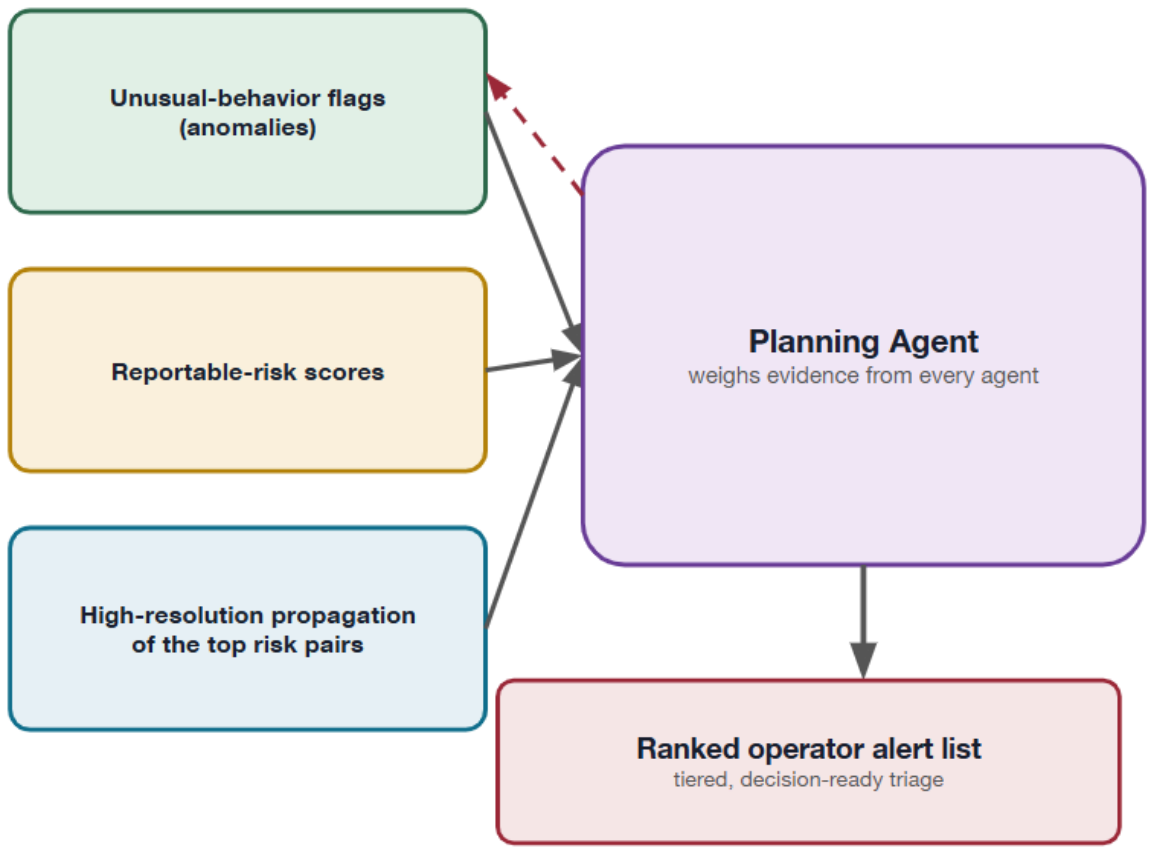

***Figure 22d:*** *Stage 4, the fusion and ranking agent (designed), turns evidence into one prioritized decision. It brings together the anomaly flags, the reportable-risk scores, and high-resolution propagation of the most threatening pairs, and weighs them into a single ranked, decision-ready alert list. A closed feedback loop re-orders the queue, so a newly suspicious object is pushed to the front and screened first.*

## 6.3 Maneuver classification and the virtual range

Detection and classification of maneuvers are trained with help from a simulation-based virtual range, a high-fidelity environment that generates synthetic encounters to overcome the scarcity of labeled real events. Mission-analysis simulation produces about 200,396 records spanning three maneuver classes, no maneuver, orbit raising or lowering, and high change in velocity, with broad orbital-element coverage.

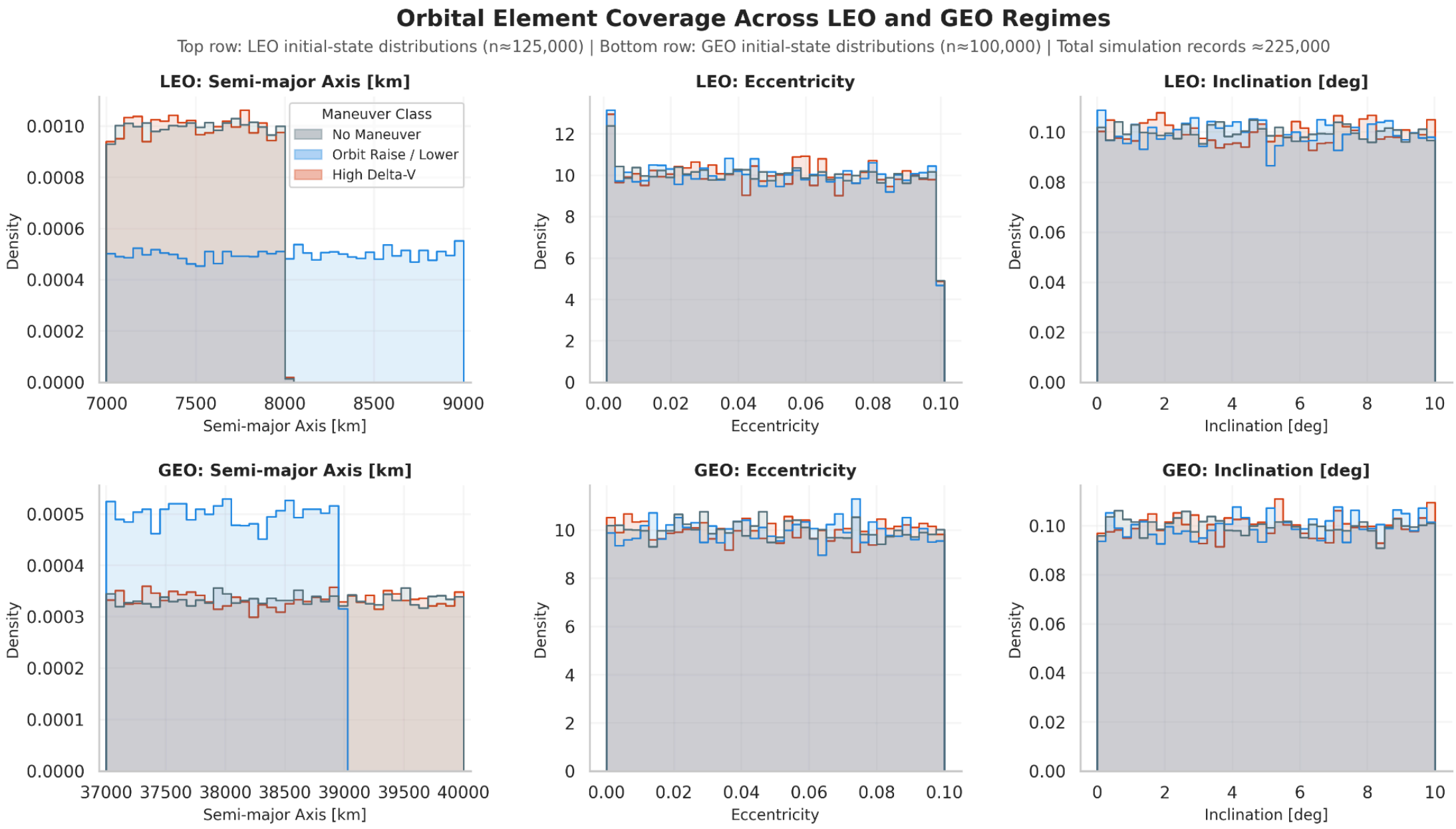

***Figure 23a:*** *Simulation-based virtual range training data. 200,396 GMAT simulation records across three maneuver classes (no maneuver, orbit raise or lower, high delta-v) with broad orbital-element coverage.*

A t-SNE projection of this dataset (van der Maaten and Hinton, 2008) shows clean separation across the three classes with low-Earth and geostationary markers appearing across all regions, which confirms that the representation captures both maneuver type and orbital regime rather than synthetic artifacts. A classifier that combines a patch-based time-series transformer (Nie et al., 2023) with a transformer encoder and the JEPA world model, trained entirely on this virtual-range data, reaches per-class precision, recall, and F1 between 0.95 and 1.00 across all maneuver types (Figure 23c). The result is not driven by the dominant non-maneuver class, which is what operational credibility requires, and the classifier can be extended to additional maneuver types such as station-keeping and in-track and cross-track maneuvers (Figure 23).

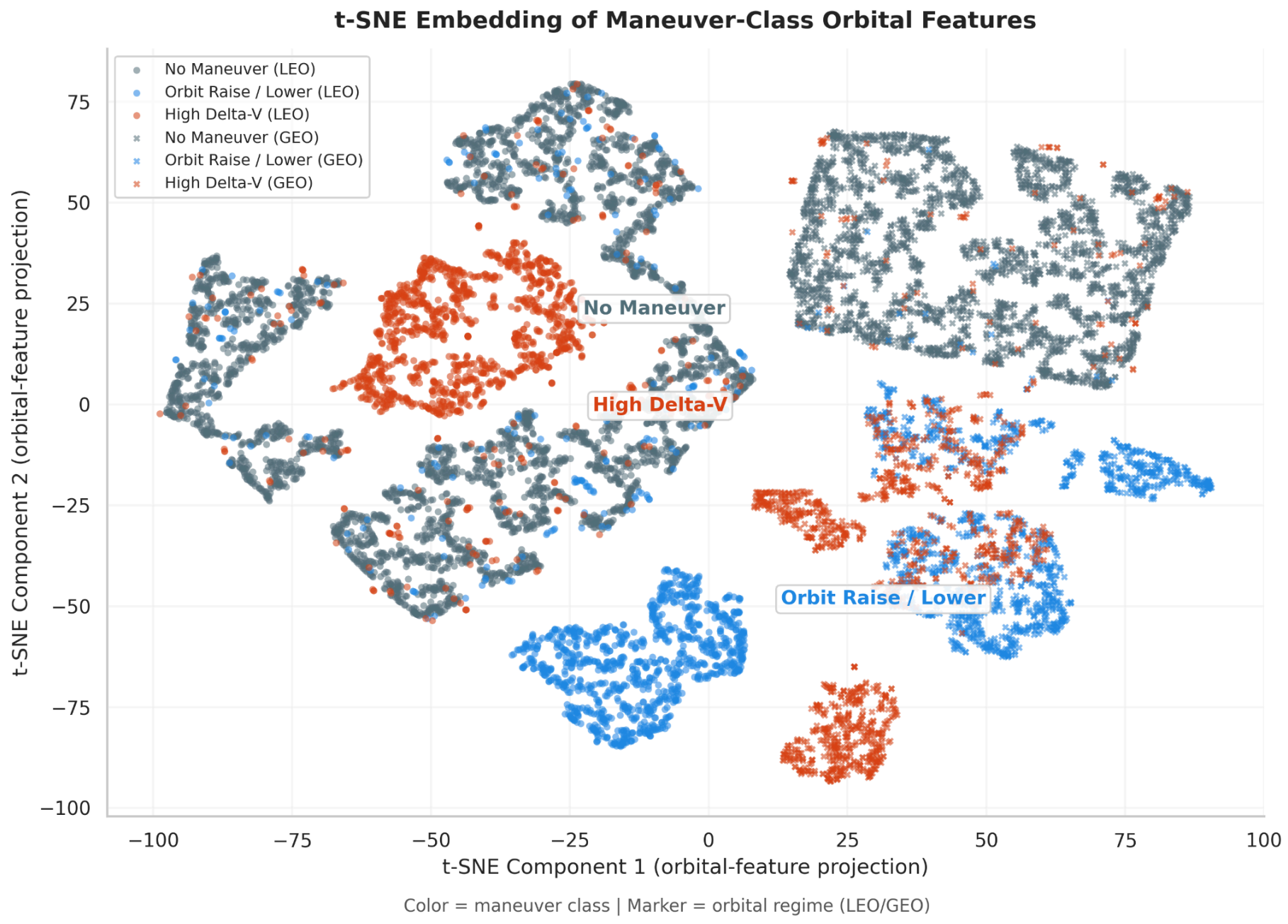

***Figure 23b:*** *A t-SNE projection of the simulation dataset, showing clear separable structure across all three maneuver classes, with low-Earth and geostationary markers appearing across all class regions, confirming the embedding captures both maneuver type and orbital-regime context rather than random synthetic variation.*

## 6.4 Validating the simulation layer and scoring confidence

Because synthetic data underpins the training, the simulation layer is itself validated against reality. Across 1,626 historical maneuvers from 19 satellites, 13 in low-Earth orbit and 6 in geostationary orbit, drawn from public agency records, each maneuver was re-simulated and compared against the satellite's observed post-maneuver orbit, with 607 maneuvers having tracking data available for validation. For routine station-keeping and small trim burns in low Earth orbit, the median reconstruction error is 1.7 to 1.9 kilometers (Figure 24b), and in a representative case it matched the real maneuver to within 0.73 kilometers across about 4.5 revolutions. The bracketing two-line element sets used as truth themselves carry errors of order a kilometer, so agreement at this level is close to the limit the truth data can resolve. A blind-spot analysis then mapped where the simulation is reliable and where it is not. Multi-burn maneuvers fail validation far more often than single burns, about a third of the time against roughly 1 in 20, and a substantial fraction of recorded maneuvers fall outside the available tracking window and cannot be validated, which is a data limitation rather than a model deficiency. This kind of honest characterization of limits is part of the assurance argument, not a footnote to it (Figure 24).

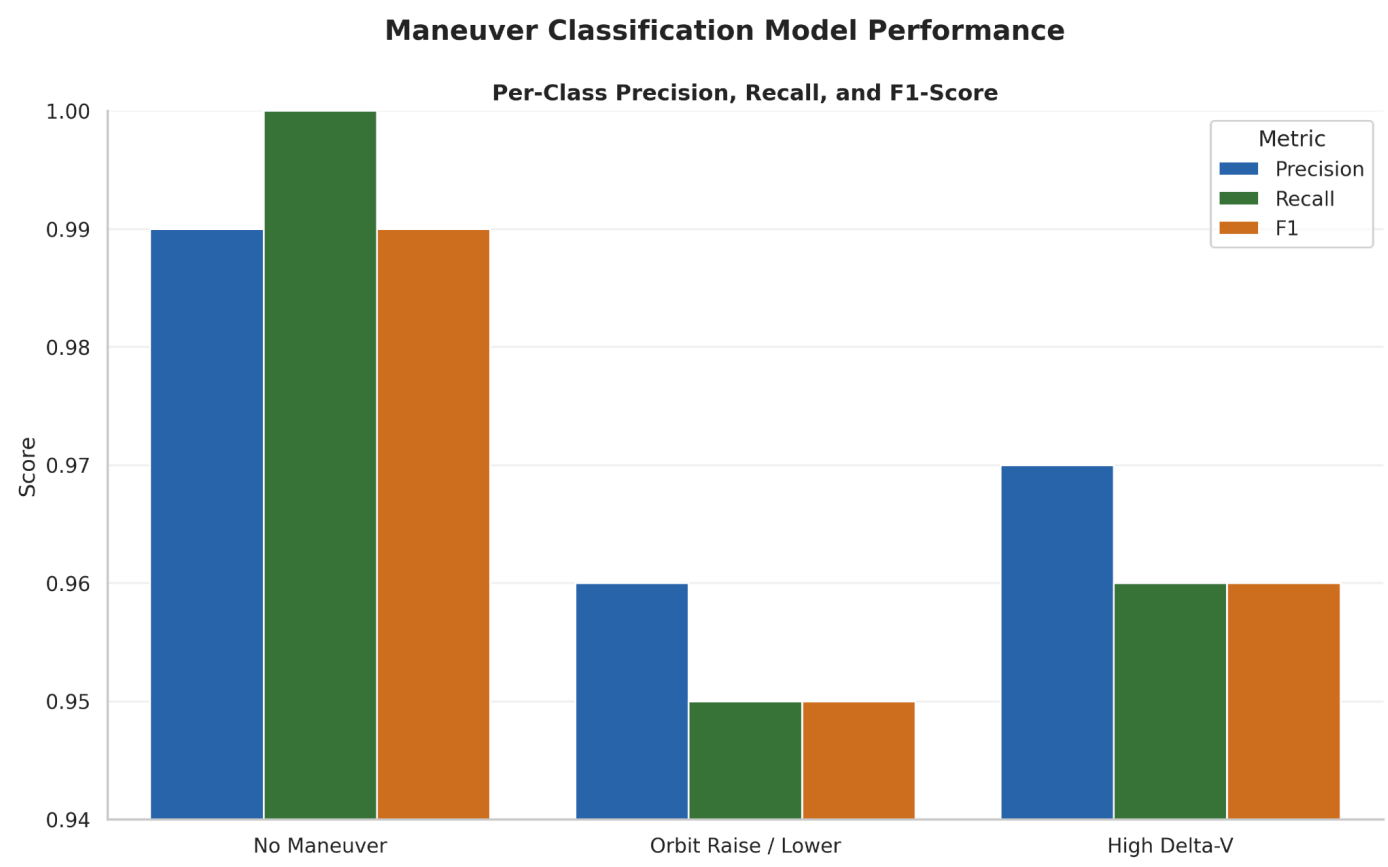

***Figure 23c:*** *PatchTST plus Transformer plus JEPA classifier trained entirely on simulation-based virtual-range data, with per-class precision, recall, and F1 between 0.95 and 1.00 across all maneuver types. The result is not driven by the dominant baseline class.*

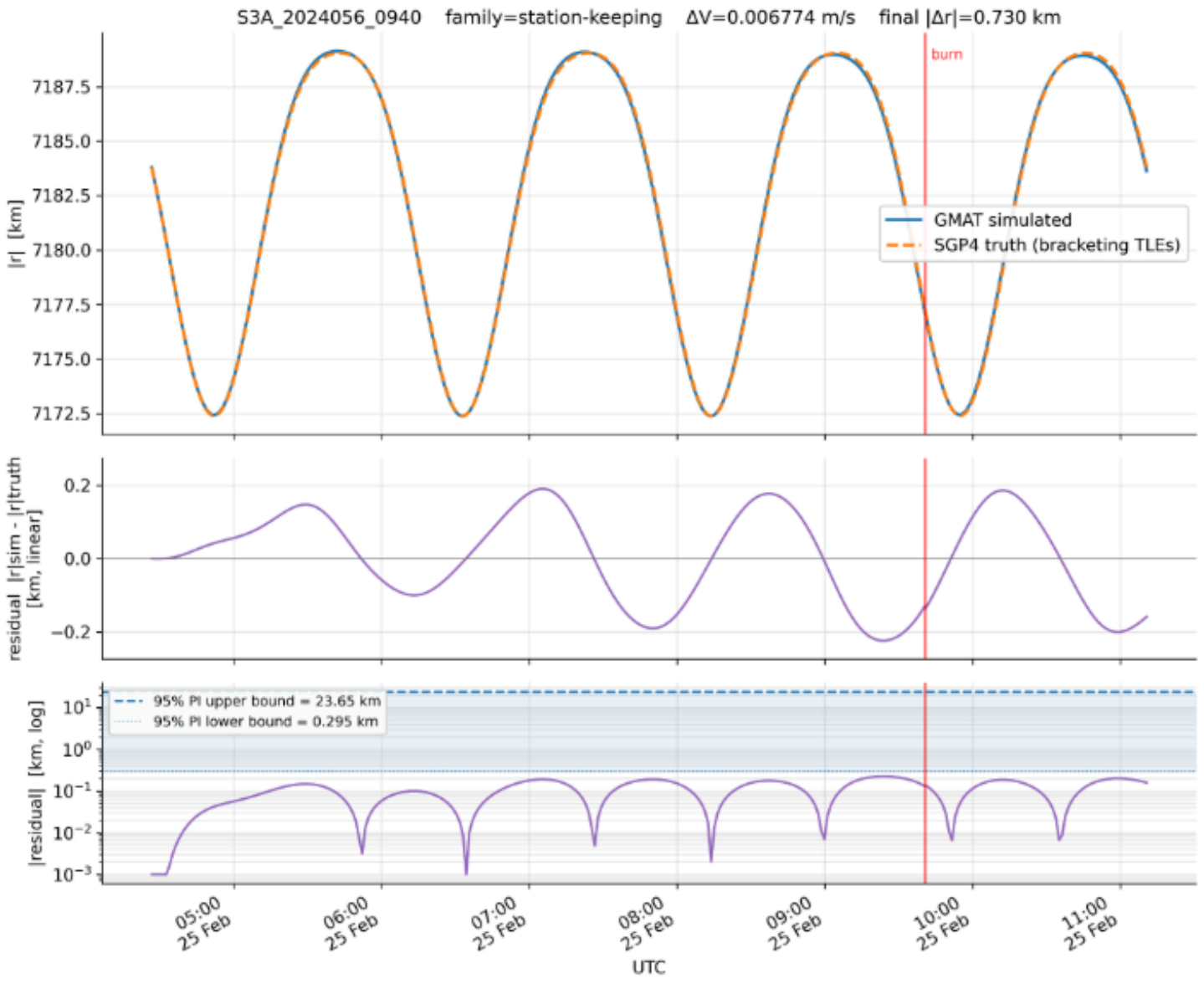

***Figure 24a:*** *A single validated case, a small station-keeping burn on Sentinel-3A on 25 February 2024. The GMAT-simulated track sits directly on the observed orbit, reconstructed by SGP4 propagation of the bracketing two-line element sets, across about 4.5 revolutions. After re-flying the real maneuver, the simulation matched the observed orbit to within 0.73 km.*

Operationally, every detection carries a measure of how much to trust it. Each alert is stress-tested by applying about 30 small, repeatable perturbations that mimic measurement uncertainty and rerunning the detection to see whether the call holds. The operator sees a confidence score, an agreement rate, and a flip rate, which distinguish a robust detection from a borderline one and let the operator prioritize which alerts to examine first. The same orbital simulation also supports the generation of difficult adversarial scenarios for training and evaluation. A purpose-built parallel propagator runs about 1,000 times faster than the mission-analysis tool while preserving sub-meter accuracy over long trajectories, which makes it practical to search for hard encounter geometries and to train and stress-test the detection, prediction, and response systems.

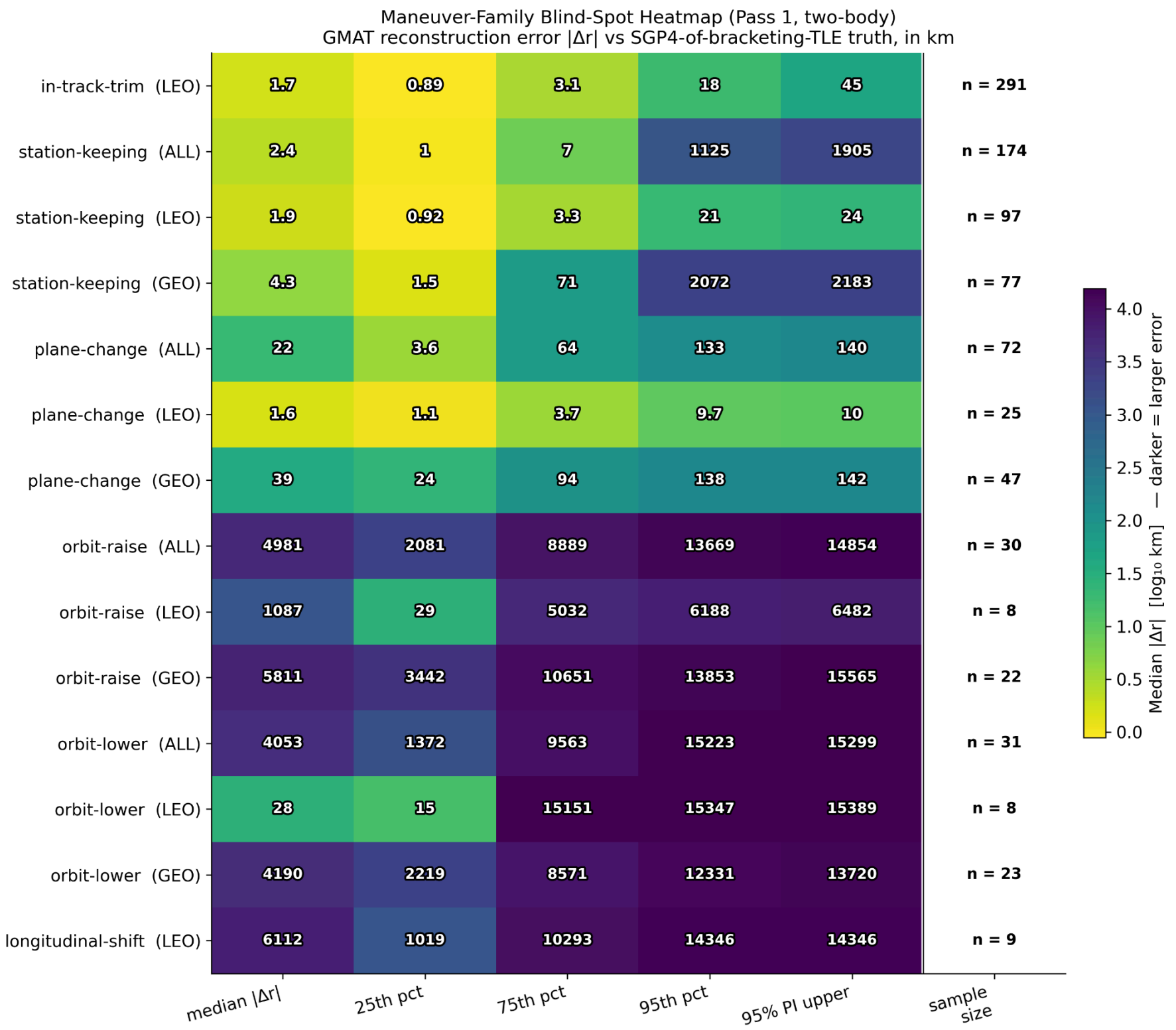

***Figure 24b:*** *The maneuver-family blind-spot analysis. All 607 validated maneuvers are grouped by maneuver family and orbital regime, and color encodes the median reconstruction error against the bracketing-TLE truth on a log scale. Single-burn families in low-Earth orbit reproduce well, while multi-burn and large transfers land thousands of kilometers off: multi-burn maneuvers fail validation about 33% of the time against about 5% for single burns. The analysis also exposes a coverage gap, since roughly 60% of recorded maneuvers fall outside the available tracking window and cannot be validated, a data limitation rather than a model deficiency.*

## 6.5 Synthetic orbital-interaction data and maneuver-intent classification

Historical proximity events are rare, incomplete, and weakly labeled, which makes them insufficient for training models that must tell benign maneuvering apart from suspicious or threat-like behavior. To close that gap, GURU generates synthetic interaction episodes in a controllable orbital simulation. The environment samples diverse target and chaser states, propagates both objects forward, and lets an artificial actor apply impulsive maneuvers in a local orbital frame. This is a war-gaming environment (Figure 25) in which a policy can explore counterfactual behaviors such as delayed approach, fast intercept, fuel-limited pursuit, evasive phasing, and deceptive trajectory changes. The output is a large library of labeled trajectories with known intent, full state history, maneuver timing, fuel use, and closest-approach geometry.

To make generation scalable, the propagation runs on a custom PyTorch orbital propagator that stands in for GMAT during high-throughput scenario generation. GMAT stays useful for high-fidelity validation and spot checks, but script-driven high-fidelity propagation is too slow to generate rare adversarial cases at scale. The propagator runs fully vectorized on GPU or XPU hardware and evaluates hundreds of thousands of candidate trajectories at once. It models chaser, target, and no-burn baseline trajectories, applies impulsive burns in the local VNB frame, tracks fuel use, and computes miss distance, closest approach, and relative speed.

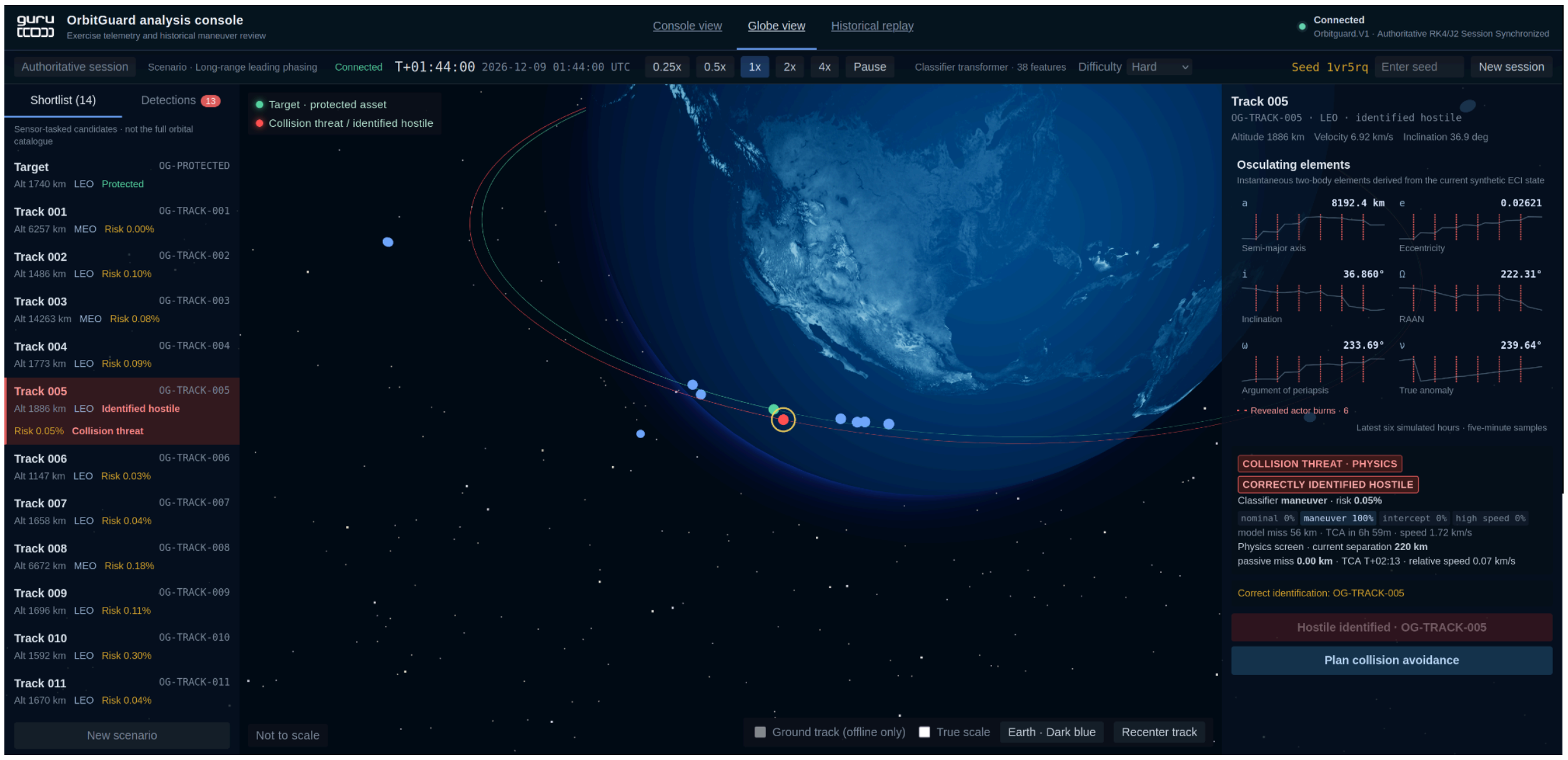

**Figure 25:** *OrbitGuard Globe View from a synthetic exercise. The view displays the protected satellite, candidate tracks, projected orbits, threat classifications, and proposed avoidance trajectories in an interactive three-dimensional view, and supports operator training and visual analysis of how synthetic adversarial behavior evolves into observable orbital behavior.*

| run | <=10 km | <=1 km | <=100 m |
|---|---|---|---|
| fixed 100 | 93.09% | 10.64% | 0.206% |
| fixed 300 | 95.70% | 21.60% | 0.445% |
| fixed 100 refine v2 | 96.27% | 61.17% | 4.384% |

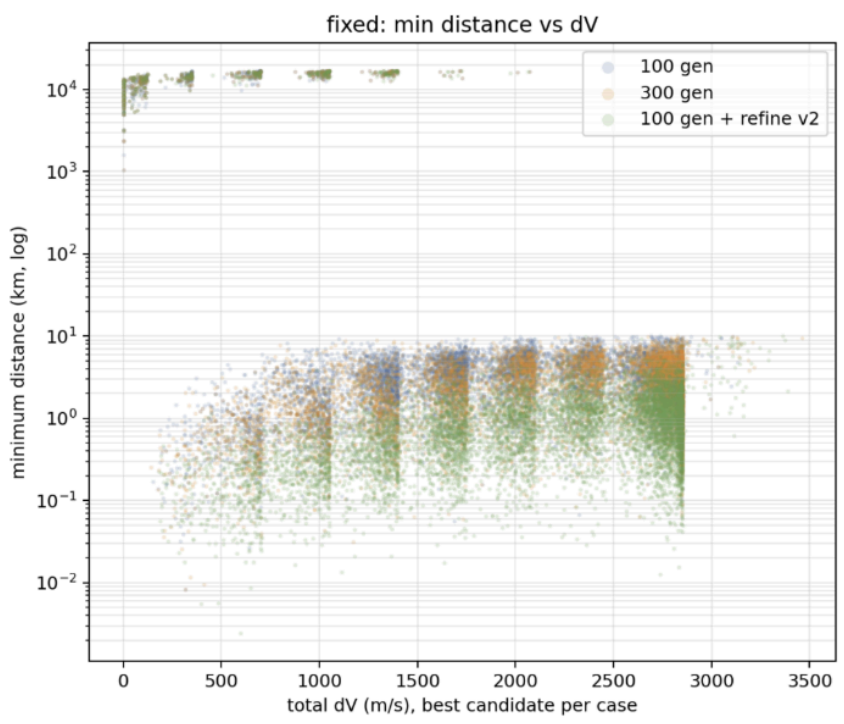

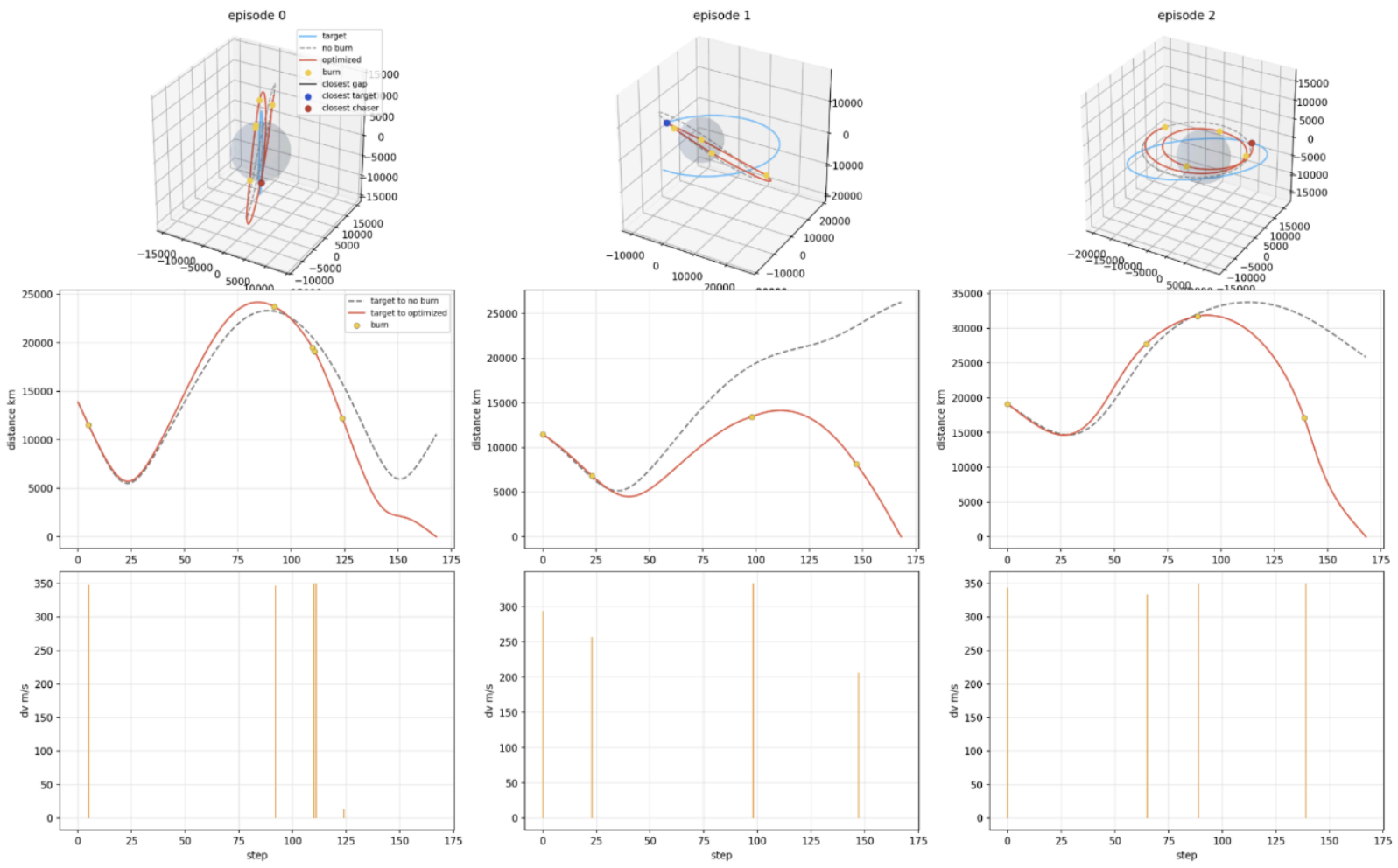

**Figure 26:** *Synthetic intercept generation and refinement performance. The table compares a 100-generation evolutionary search, a 300-generation search, and a 100-generation search with local refinement, and shows that refinement produces the strongest sub-kilometer and sub-100-meter results. The delta-v versus minimum-distance scatter plot shows the tradeoff between maneuver effort and intercept quality, with refined candidates reaching lower miss distances without a proportional rise in total delta-v. The trajectory panels show target, no-burn baseline, and optimized chaser paths with burn locations, illustrating how timed maneuvers turn natural misses into close intercepts.*

Three complementary optimizers generate the maneuvers. Reinforcement learning trains policies directly in the environment to optimize objectives such as intercept quality, fuel efficiency, delayed detectability, or high relative velocity; its reward combines terms for progress, proximity, miss distance, fuel use, maneuvering, early detection, intercept, speed of approach, and failure. Backpropagation through time treats the burn sequence as differentiable control variables and optimizes burn timing and magnitude through the propagator. Evolutionary search maintains populations of variable-length burn schedules, recombines and mutates them, and then locally refines the best candidates. Together they produce diverse trajectory families rather than a single narrow class of encounters.

The generator's purpose is not optimal interception, which classical targeting solves directly for simple cases, but a diverse library of difficult encounters, fuel-limited, multi-burn, and deliberately deceptive, for training and stress-testing detectors. Across a large generated set, a population search followed by local refinement produced the strongest close-intercept results. We compared three configurations of the evolutionary search. All three placed more than 93 percent of cases within 10 kilometers, but they differed sharply at closer ranges. A 100-generation search reached 18.64 percent of cases within 1 kilometer and 0.206 percent within 100 meters. A 300-generation search reached 21.60 percent and 0.445 percent. A 100-generation search followed by local refinement reached 61.17 percent within 1 kilometer and 4.384 percent within 100 meters. Because every generated case carries ground-truth intent and complete state and action provenance, a classifier can learn from the maneuver history, the timing, the relative geometry, the fuel expenditure, and the deviation from a ballistic baseline, rather than from proximity alone (Figure 26).

## 6.6 The space-operations agent workflow

Within GURU, space domain awareness is implemented through its OrbitGuard service as a hierarchical agent workflow that mirrors the pattern used for computational fluid dynamics. A data-retrieval agent ingests updated orbital and conjunction data from public and operator sources. A navigation agent assembles per-object time series and maintains catalog state. Predictor agents compute and cache future-state embeddings.

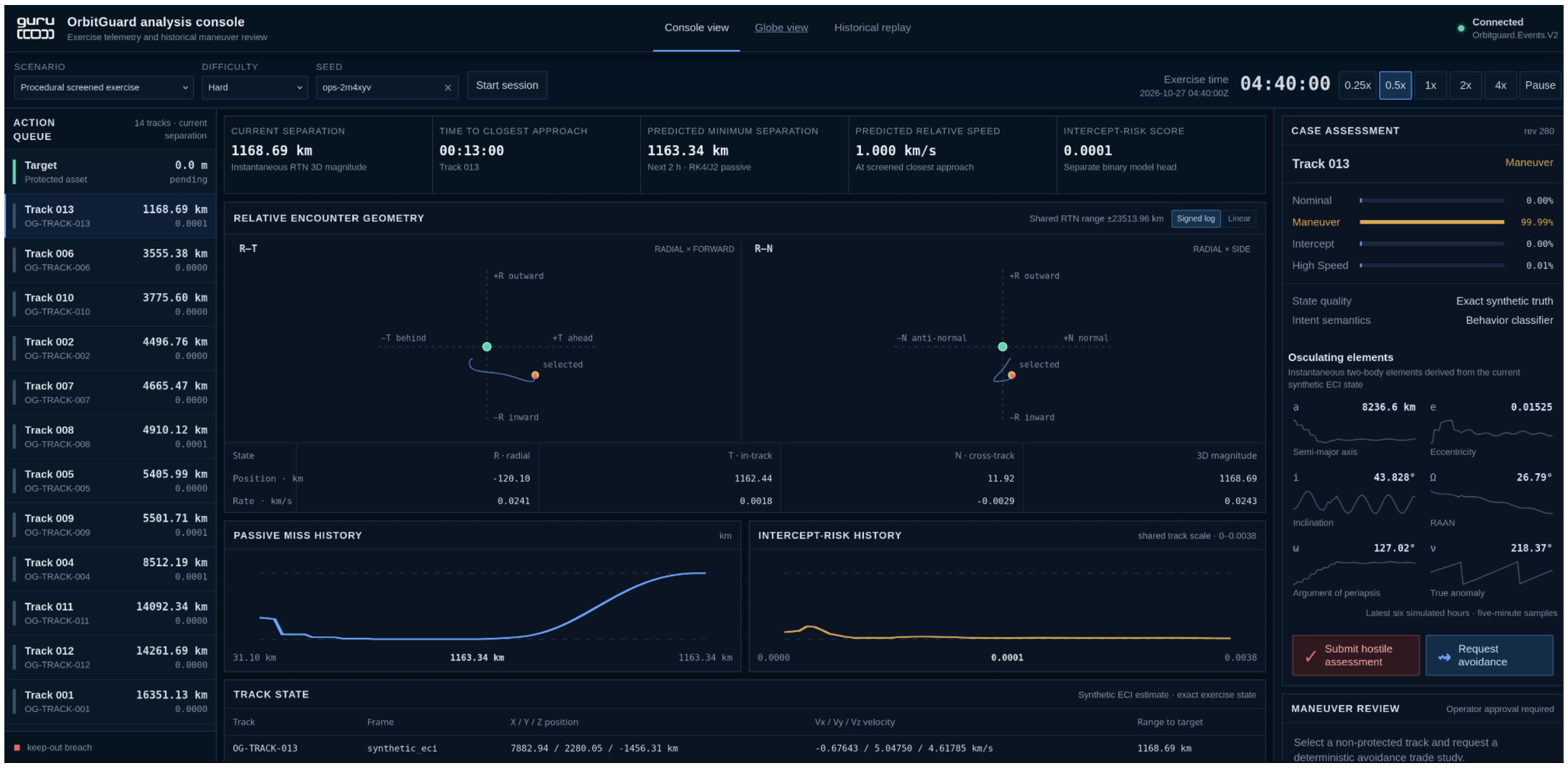

***Figure 27:*** *OrbitGuard Analysis Console for a synthetic exercise, showing track tables, relative encounter geometry, orbital-element histories, conjunction metrics, classifier outputs, maneuver plans, and an event timeline. Avoidance maneuvers are deterministic trade studies that require operator approval. The console supports detailed policy evaluation and trains operators to interpret evidence, identify threat-like behavior, and select a response. Video:* [https://youtu.be/aAnbW2aGpN8](https://youtu.be/aAnbW2aGpN8)

Anomaly agents watch for maneuvers, stability changes, and out-of-family behavior. A conjunction agent applies the pre-filters and evaluates candidate pairs with the learned risk head. A reporting agent consolidates the results into operator-facing alerts and drives the operator's visualization software through the same streamed-display control used elsewhere in the system. When an operator selects a track, the console can request a deterministic avoidance trade study for the protected asset, and any resulting maneuver requires operator approval before it is issued (Figure 27). The live ingestion path processes more than 20,000 objects at about a 2-minute end-to-end latency and feeds the screening graph.

# 7. Application III: Airspace and Reentry Decision Support

Reentering space objects pose hazards to aircraft and can disrupt airspace operations (Wright et al., 2025). The Federal Aviation Administration's first safety alert on space launch activity, issued in January 2026, tells operators that when a mishap occurs debris has fallen within or near the debris response area, that pilots should exercise extreme caution near these areas, and that the FAA activates a debris response area by notice during an unplanned catastrophic event involving falling debris (Federal Aviation Administration, 2026a). Knowing where debris will be, and when, before and during such an event is the capability this section builds toward. We apply GURU to this problem through three connected stages: debris detection and state estimation, propagation and prediction, and aircraft conflict assessment and avoidance. The application is designed to combine radar observations, atmospheric data, and physical simulations to estimate debris motion and its uncertainty, and to update predictions as new observations arrive. The intended outcome is timely information for airspace decisions: which aircraft or routes may be affected, what areas and altitude ranges should be avoided, and when those restrictions can be lifted. The reference event throughout is the Long March 4C rocket-body reentry over the upper Midwest of the United States on January 24, 2026, supplemented by constructed synthetic encounters. The decision-support application uses the same Common Internal Language and agent platform described in the preceding applications, which makes it evidence that the platform transfers across domains rather than being rebuilt for each. The validated references in this application are the physics-based propagators and NASA's DAIDALUS detect-and-avoid logic; GURU orchestrates them and presents their outputs as decision support, and operational authority remains with the Federal Aviation Administration and air traffic control.

## 7.1 Hierarchical approach

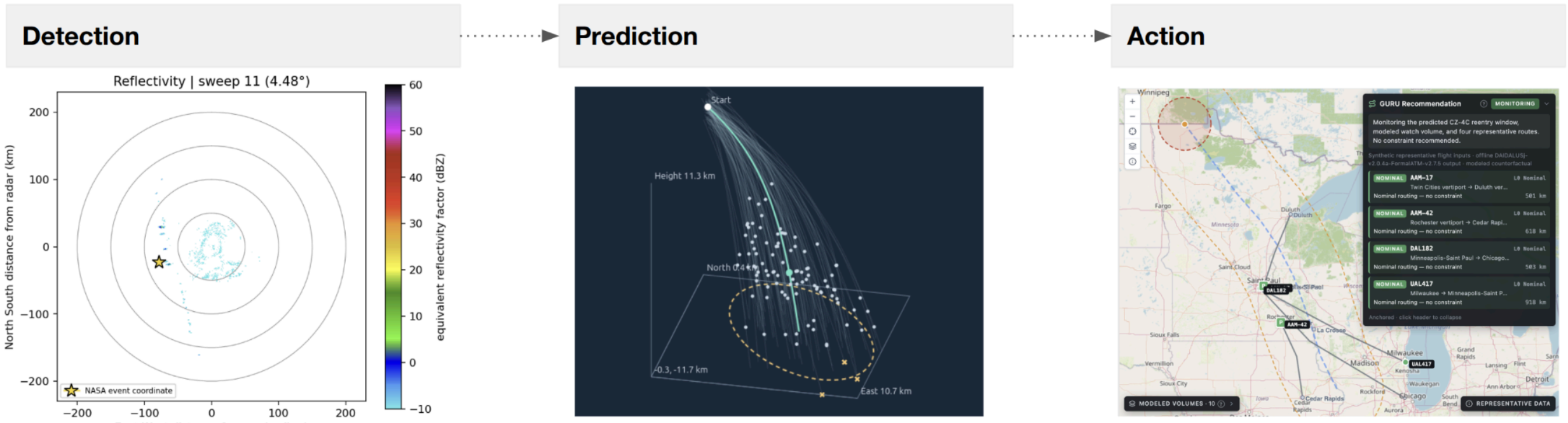

***Figure 28:*** *From radar observations to aircraft avoidance. The detection stage identifies candidate echoes and estimates their location and motion. Propagation explores possible debris trajectories and affected regions under uncertain initial conditions. The decision stage compares potential hazards with*

*aircraft routes and presents avoidance options. These panels illustrate separate project demonstrations; they do not establish a validated end-to-end reconstruction of the historical event.*

**Detection and state estimation.** Weather-radar measurements are screened for unusual returns, grouped into candidate echoes, and compared across scans (Fries & Fries, 2010). Echo locations, observation times, beam-height intervals, and Doppler measurements provide complementary constraints on position and motion. The initial demonstration derives a starting position from a seven-cell cluster; subsequent analysis estimates motion for four regional echo sequences.

**Propagation and prediction.** Starting from specified or sampled entry conditions, Orekit propagates the parent body to the breakup stage (Maisonobe et al., 2010), TITAN simulates a prescribed component-breakup scenario (Peddakotla et al., 2022), and a fragment-descent model built on SciPy (Virtanen et al., 2020) propagates the resulting fragments under gravity, aerodynamic drag, and wind. Multiple runs vary entry conditions, breakup configurations, fragment properties, and atmospheric conditions. Their trajectories produce landing and altitude-crossing regions that summarize the simulated spread and support comparison with timestamped radar observations.

**Decision and avoidance.** This stage brings debris predictions together with aircraft positions and routes to assess potential encounters by location, altitude, and time. The dashboard displays event evidence, modeled hazard regions, and candidate reroutes. Separate synthetic encounters use DAIDALUS to evaluate conflicts and heading guidance (Muñoz et al., 2015), with an additional forecast-based planner screening maneuvers against predicted fragment motion.

## 7.2 Detection and state estimation

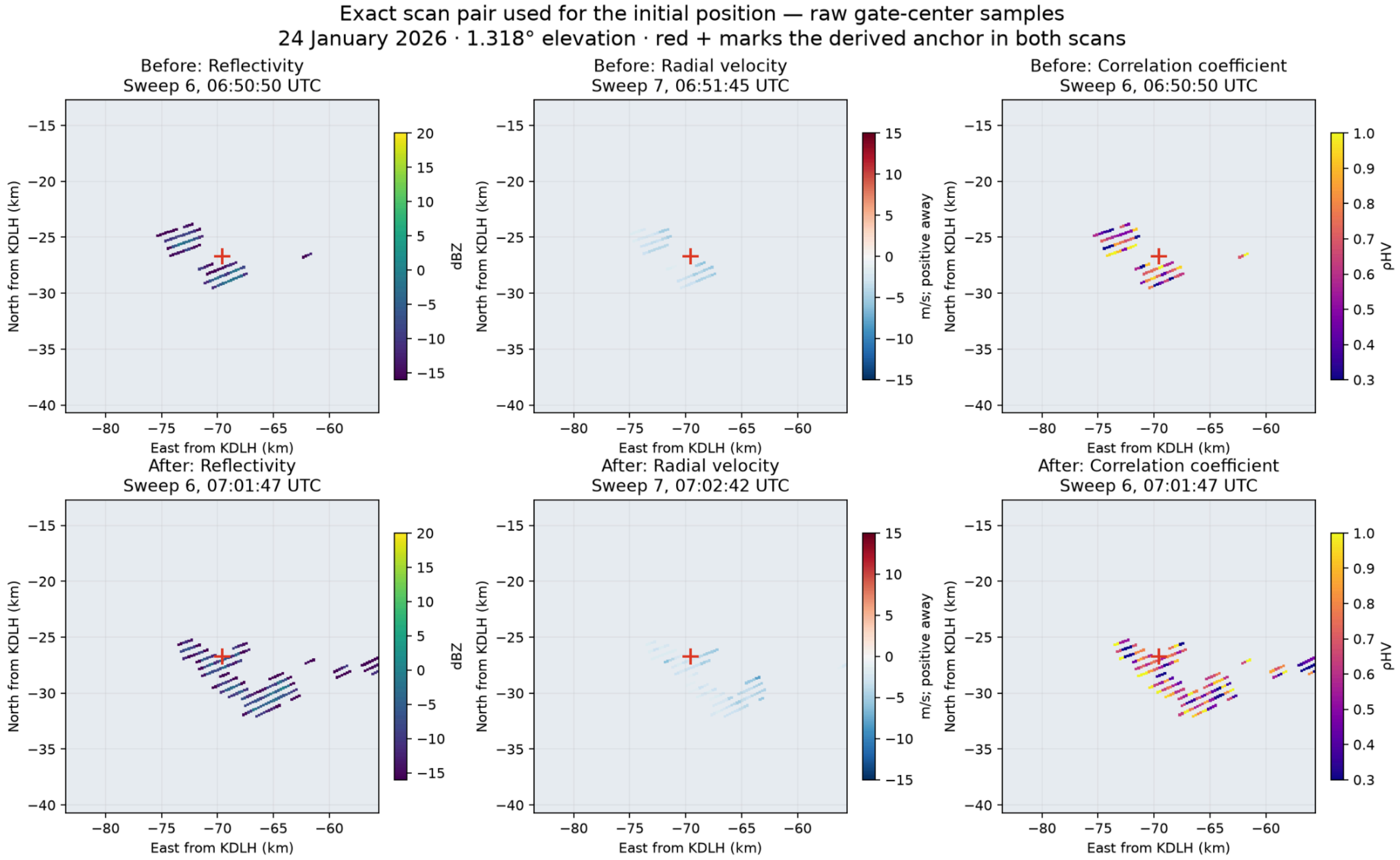

***Figure 29:*** *Radar measurements surrounding a candidate detection. Columns show reflectivity, radial Doppler velocity, and dual-polarization correlation coefficient before and after the selected detection.*

*Points represent native radar gate centers; diagonal patterns follow the radar's sampling geometry and do not indicate debris travel direction. Red crosses mark the position later derived from the seven-cell cluster. Coordinates are relative to the fixed KDLH radar, and panel timestamps differ because radar moments were collected in separate sweeps.*

The detection stage processes historical NEXRAD (Next Generation Weather Radar) observations, integrating reflectivity, radial Doppler velocity, and dual-polarization correlation metrics (Fries & Fries, 2025). Every observation samples a finite volume along a given radar beam at a specific timestamp and tilt angle. As a result, each gate has defined spatial dimensions, while different moment fields may not be strictly concurrent. Reflectivity reflects echo intensity, Doppler measurements constrain motion along the line of sight, and polarimetric variables help characterize scatterer properties. No single metric independently establishes the presence of reentering fragments.

Automated detection of meteorite signatures in weather-radar data has previously been explored using unsupervised learning (Smeresky et al., 2021). In our implementation, an initial detection pass mapped observations onto a 1 km grid, selecting returns with reflectivity of at least −10 dBZ and an increase of at least 6 dB over the preceding scan. Adjacent qualifying cells were grouped into clusters of at least three cells and subsequently linked across multiple elevation sweeps and radar volumes. Their mean spatial center and beam height defined the starting coordinates for an initial propagation test, paired with an assumed velocity vector.

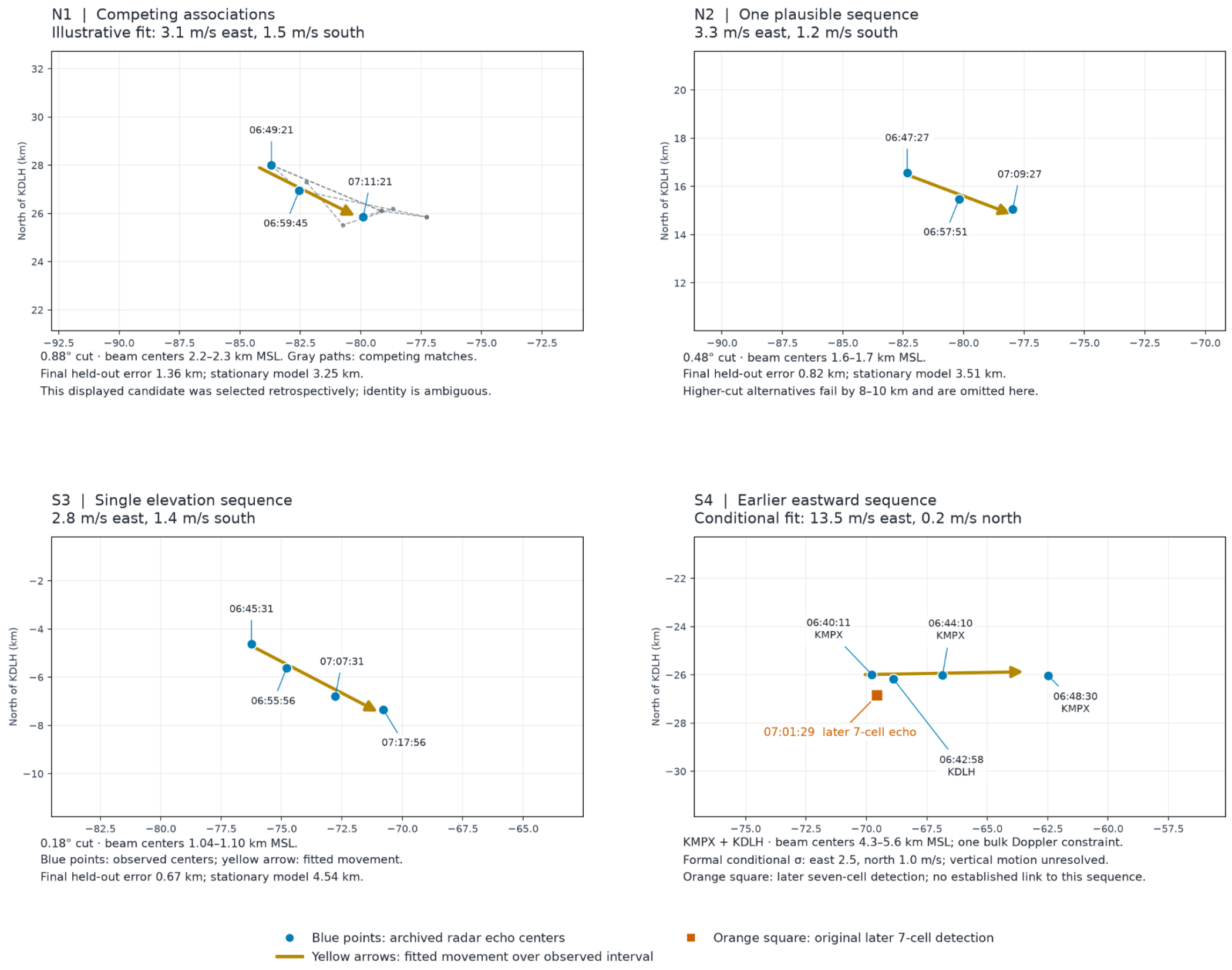

***Figure 30:*** *Conditional motion estimates for four regional echo sequences. Blue markers show selected radar echo centers, and yellow arrows indicate fitted movement over their observation intervals. N1, N2, and S3 exhibit slow southeastward motion; the earlier S4 sequence exhibits faster eastward motion. Gray paths in N1 show competing associations. The orange square in S4 marks the later seven-cell detection, which has not been linked to that sequence.*

Subsequent analyses preserved native gate geometry, precise timestamps, and beam-interval bounds to evaluate competing track associations over successive scans. The four highlighted regional sequences yield conditional motion fits: approximately 3 m/s toward the east-southeast for N1 to S3, and a faster 13.5 m/s eastward movement in the earlier S4 sequence. Projecting the estimated motion onto the radar line of sight enables a consistency check against measured Doppler velocity (Fries & Fries, 2010). These motion estimates characterize observed echo patterns; descent rates, physical fragment identities, and linkage to the primary seven-cell detection have not been demonstrated.

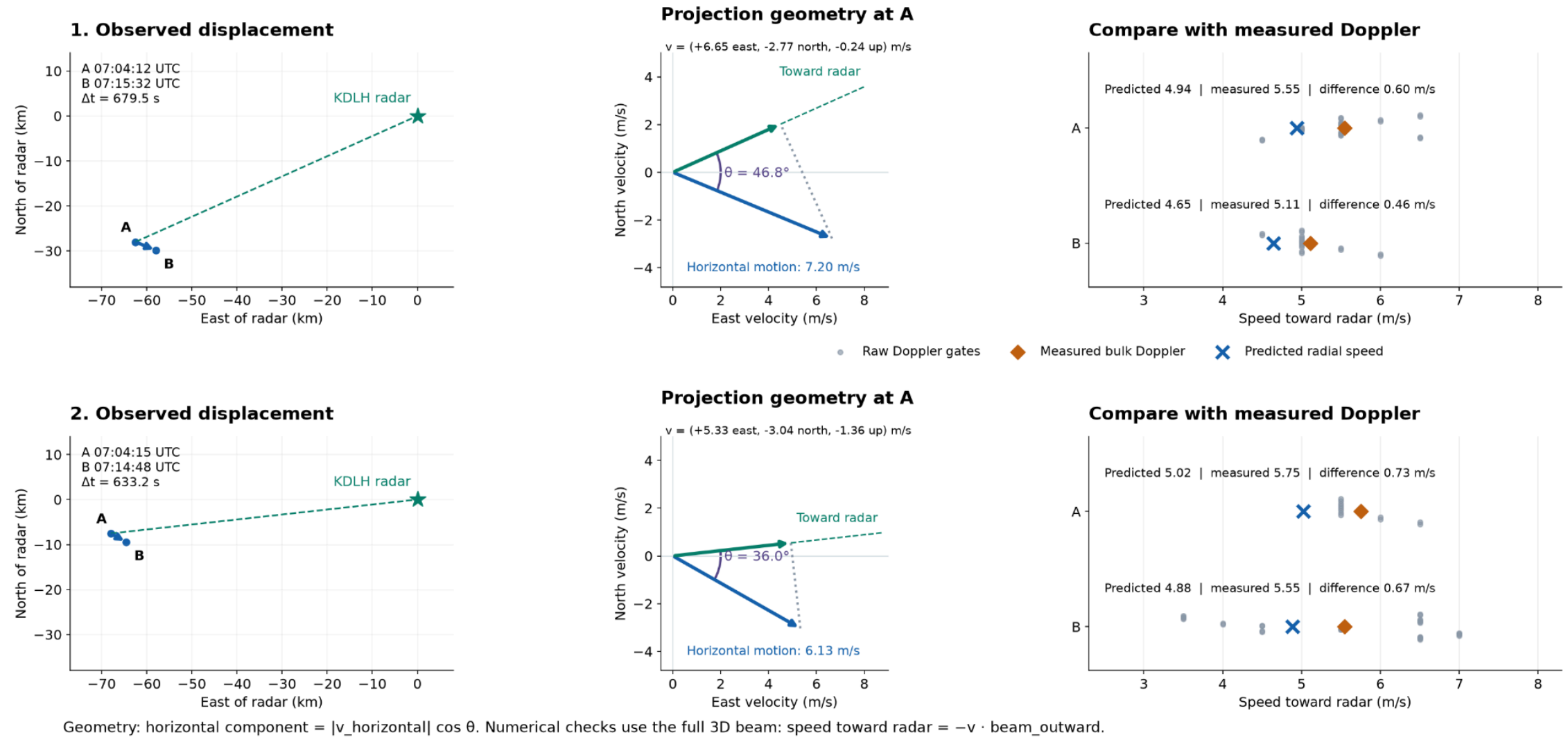

***Figure 31:*** *Checking displacement-based motion against radial Doppler measurements. Each row shows a selected observation pair: measured displacement on the left, projection of the inferred velocity toward the radar in the center, and comparison with Doppler measurements on the right. Gray points are individual Doppler gates, orange diamonds are bulk estimates, and blue crosses are predicted radial speeds. The displayed differences are approximately 0.46 to 0.73 m/s. This agreement supports local kinematic consistency under the selected association.*

## 7.3 Propagation and prediction

Propagation converts a specified parent state into alternative fragment trajectories. The full-entry experiments use Orekit for the parent trajectory, TITAN for prescribed component-breakup scenarios, and the SciPy-based descent model for the subsequent descent of surviving fragments. Simulations of the real event begin from assumed entry conditions informed by event context. Synthetic cases use explicitly defined initial conditions. The stages exchange positions, velocities, times, and fragment properties so that the resulting trajectories can be displayed in a common geographic and temporal reference.

Atmospheric descent predictions depend on wind and fragment properties, including mass, shape, and aerodynamic drag (Towner et al., 2022). Fragment descent includes gravity and aerodynamic drag based on velocity relative to the surrounding air. Atmospheric wind, temperature, and pressure come from NOAA High-Resolution Rapid Refresh (HRRR) pressure-level analyses downloaded from the public Amazon Web Services archive (Dowell et al., 2022). The sampler interpolates the fields horizontally, vertically, and between analysis times along each trajectory, calculating dry-air density as pressure divided by the gas constant and temperature. Above the available weather levels, the model uses a declared atmospheric extension; the broad-entry experiment blends into an upper-atmosphere density table. Lightweight, high-area fragments decelerate strongly and can become wind-driven, while compact components retain more downrange motion.

Repeated simulations explore uncertainty in entry speed, direction, descent angle, breakup conditions, and component properties. Each breakup result also supports additional, cheaper realizations that perturb fragment aerodynamics, separation velocity, and atmospheric biases. The broad-entry experiment contains ten parent and breakup scenarios, producing 100 nominal fragments across the alternative scenarios and 400 stored descent paths.

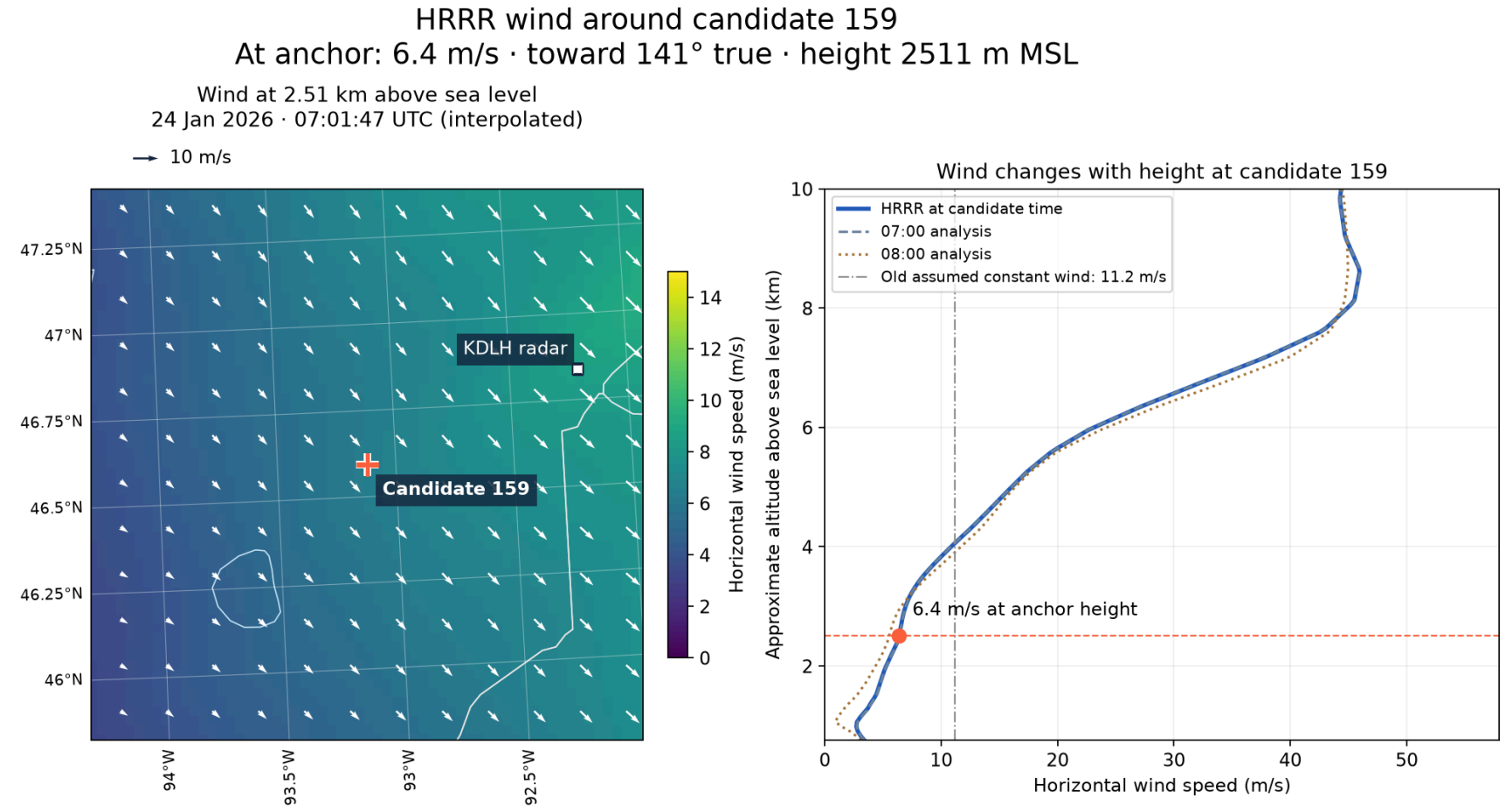

***Figure 32:*** *HRRR wind field near a candidate. The map shows interpolated horizontal wind at approximately 2.51 km above sea level at 07:01:47 UTC on January 24, 2026. Colors encode wind speed and arrows indicate the direction of air motion. The vertical profile compares the interpolated field with the adjacent 07:00 and 08:00 analyses and the earlier constant-wind assumption. At the candidate location and height, the interpolated wind is approximately 6.4 m/s toward 141 degrees true. These are atmospheric-model fields, not debris velocities.*

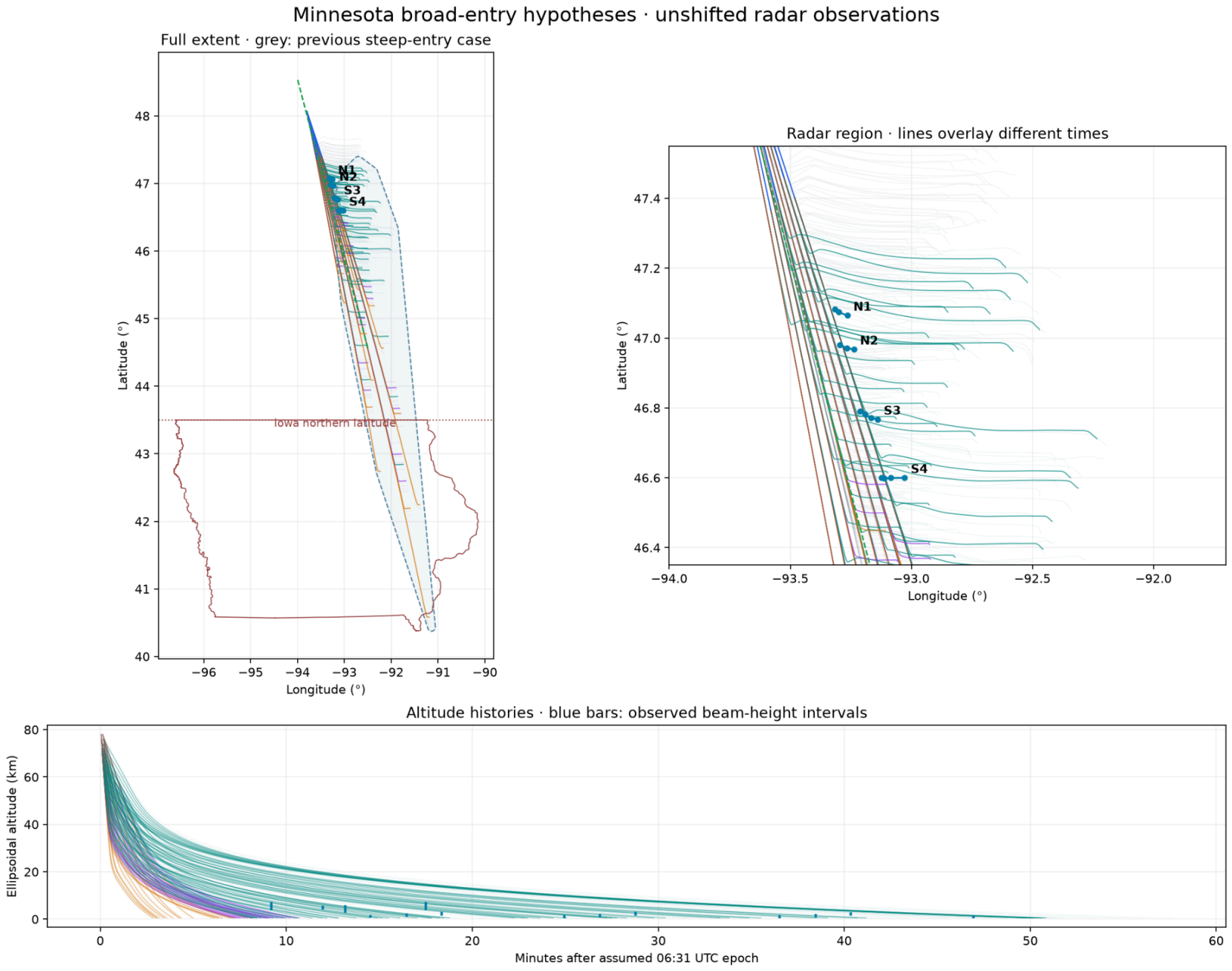

***Figure 33:*** Alternative Minnesota reentry scenarios compared with radar observations. The full-region map shows simulated trajectories extending from northern Minnesota toward Iowa; the inset resolves their relationship to the N1 to S4 radar regions. Orange, purple, and teal distinguish dense, intermediate, and lightweight component families, while gray paths show the earlier steep-entry case. The lower panel shows altitude histories against time after the assumed 06:31 UTC epoch, with blue bars indicating

observed beam-height intervals. Map paths overlay different times: spatial proximity alone does not demonstrate agreement with a radar observation.

## 7.4 Uncertainty regions

Reentry uncertainty can be propagated through Monte Carlo sampling or distribution-based approaches, including Gaussian-mixture methods coupled to fragmentation simulations (Williamson et al., 2025). Here, uncertainty is explored through alternative complete scenarios and additional fragment-descent realizations. The complete scenarios vary the parent entry and breakup configuration, including the number and properties of fragments. The additional realizations vary descent parameters within each scenario. Fragments belonging to one realization share atmospheric biases, preserving the fact that they encounter a common weather environment. Overlaying all trajectories therefore shows mutually alternative possibilities, rather than one event containing every displayed fragment.

Landing regions summarize simulated ground endpoints, while altitude-crossing regions summarize where fragments pass through selected heights. Smoothing and an explicit spatial margin produce more readable boundaries around these samples. Such regions help compare scenarios and identify potentially affected areas, but an all-time outline discards encounter timing. Aircraft assessment must retain when fragments occupy each altitude and location: two trajectories crossing on a map may never approach one another in space and time.

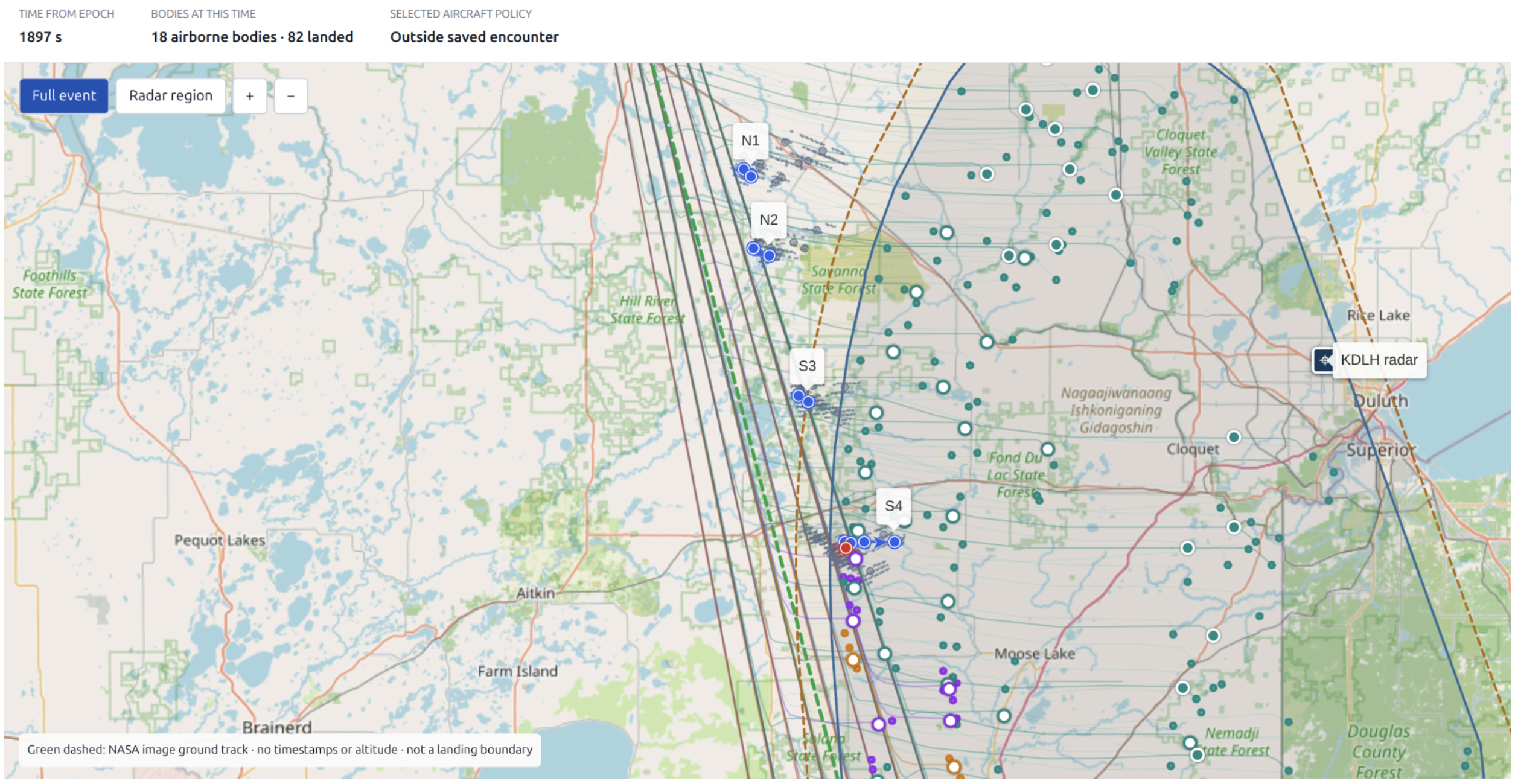

***Figure 34:*** Simulated debris trajectories, uncertainty regions, and radar observations. The map overlays alternative fragment paths and smoothed coverage bands with timestamped radar echoes and candidate motion estimates. The bands summarize the simulated spread with added spatial margins. Radar overlays support comparison with observations.

## 7.5 Avoidance and airspace decisions

Avoidance combines predicted fragment positions and velocities with aircraft states and intended routes. The relevant hazard is debris occupying the aircraft's altitude range at the encounter time, rather than the final landing area alone. DAIDALUS evaluates conflicts and maneuver guidance from the states supplied to it; the surrounding application is responsible for representing fragment hazards, supplying forecasts, and applying aircraft responses. The historical dashboard adds an evidence and decision view, showing traffic intersections with modeled bands and candidate reroutes, but those bands are not themselves a validated reconstruction of the debris field.

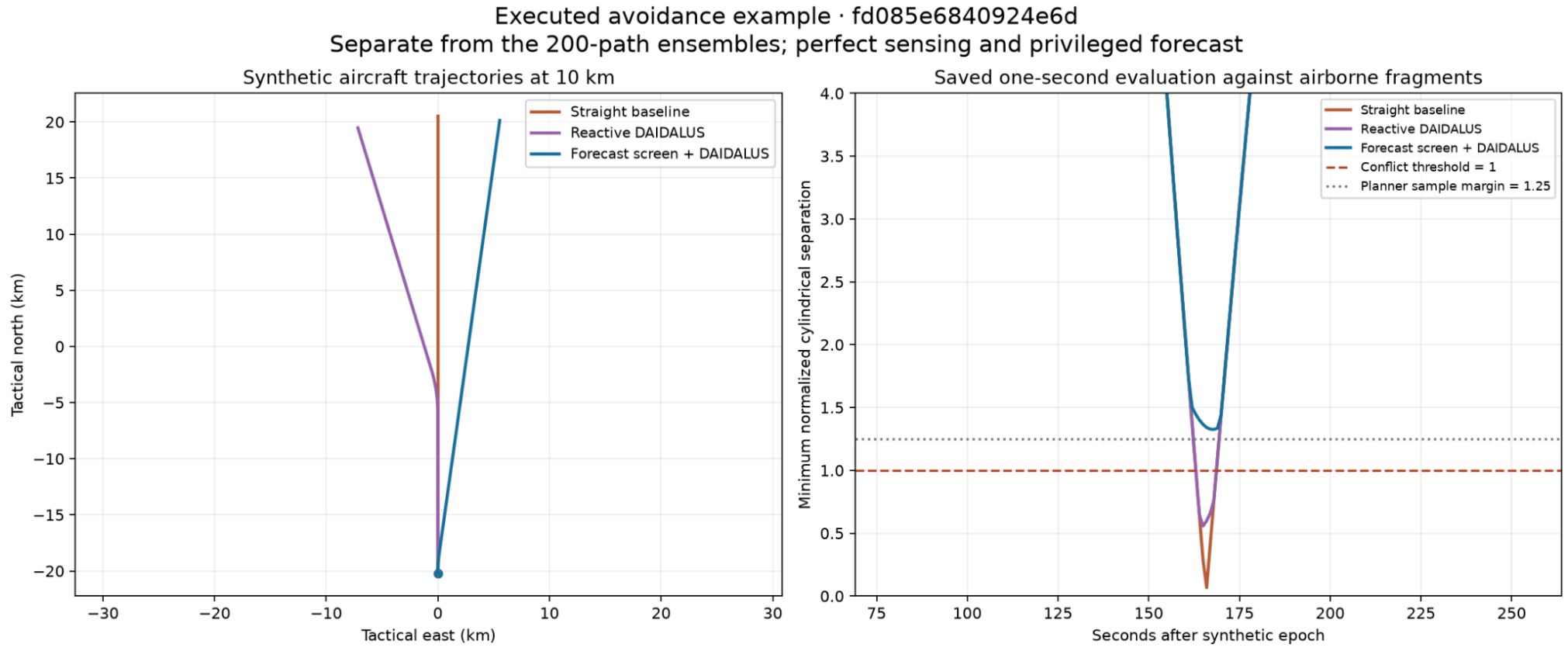

***Figure 35:*** Aircraft avoidance in a constructed synthetic encounter. The left panel compares a straight flight path, reactive DAIDALUS guidance, and an additional forecast-based maneuver screen followed by DAIDALUS monitoring. The right panel shows minimum normalized separation from airborne fragments at one-second evaluation intervals; values below one violate the configured protection thresholds. Reactive guidance turns the aircraft but still produces sampled violations, whereas the forecast-assisted maneuver avoids them in this example. The experiment uses perfect simulated states and access to future fragment trajectories and is separate from the uncertainty ensembles.

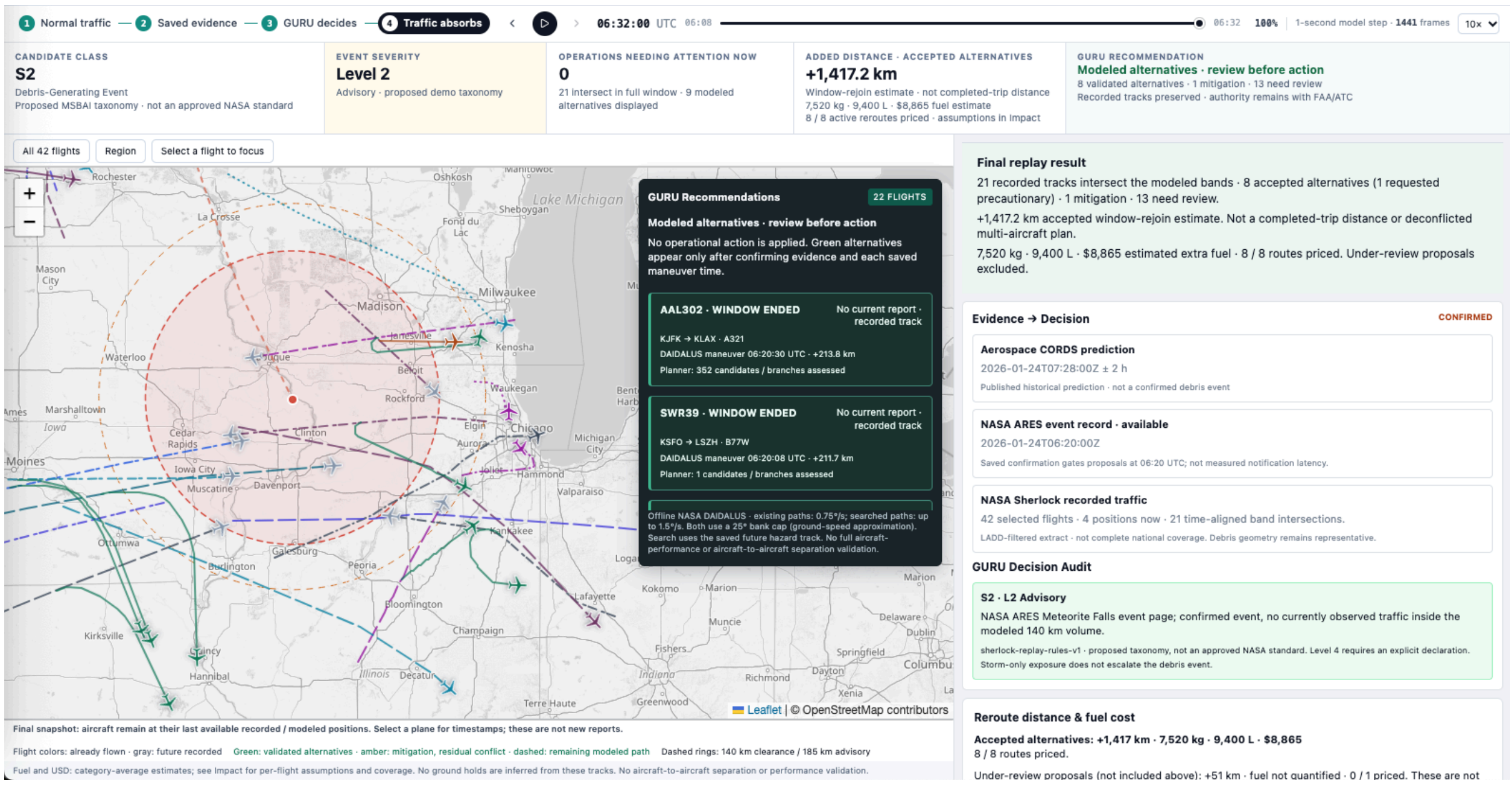

***Figure 36:*** Historical traffic replay with modeled hazard bands and candidate reroutes. The dashboard combines recorded aircraft tracks, event evidence, illustrative debris bands, and proposed route alternatives. The evidence panel distinguishes published predictions, event confirmation, and traffic records; the decision panel summarizes modeled intersections and rerouting costs. The displayed alternatives are simulation outputs, not aircraft actions that occurred during the event. Also identified in the dashboard - limits on route completion, cost estimation, and multi-aircraft conflict assessment. *Video: [https://youtu.be/Jtyf78QsXlE](https://youtu.be/Jtyf78QsXlE)*

The executed synthetic example constructs an aircraft encounter at 10 km altitude using the simulated fragment trajectories. At one-second updates, the DAIDALUS interface supplies aircraft and airborne-fragment states to DAIDALUS, using experimental protection thresholds of 2 km horizontally and 500 m vertically. The comparison evaluates straight flight, reactive heading guidance, and a separate planner that screens possible turns against nonlinear fragment forecasts. In the displayed case, reactive guidance changes heading but still violates the thresholds at some evaluated times; the forecast-assisted maneuver avoids those sampled violations. That result depends on perfect simulated state information and privileged access to the future trajectories. Connecting this demonstration to the full uncertainty ensemble requires a rule for combining alternative hazards and evaluating an entire aircraft maneuver against them. One option is to require clearance across every considered scenario; a probability-based alternative would first require calibrated scenario weights and an explicit risk criterion.

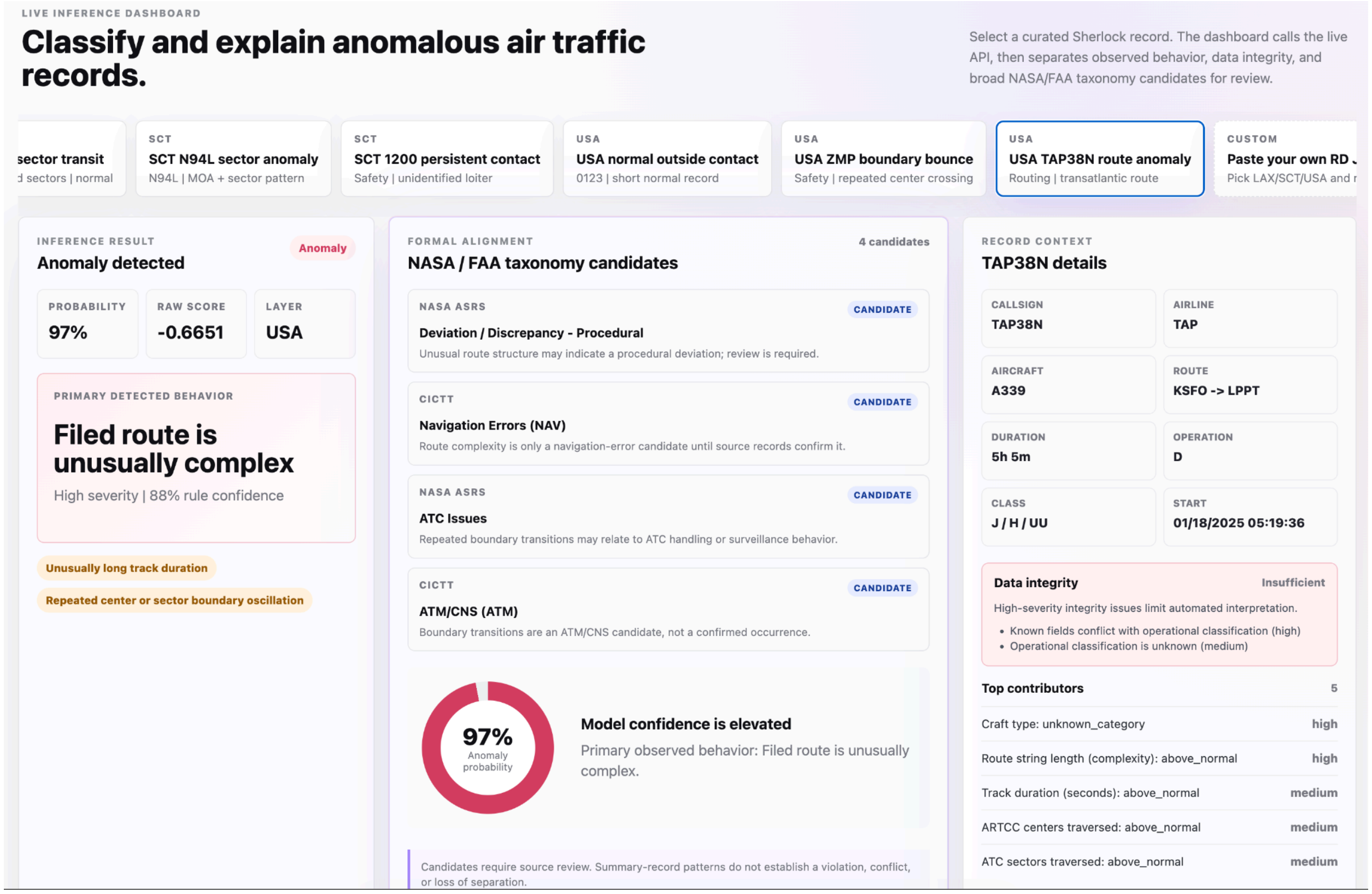

***Figure 37:*** *Flagged anomalies are organized within a four-class taxonomy of data quality, operational, safety-relevant, and routing, with associated confidence values and evidence trails supporting interpretation and response selection. The example record is shown with its data-integrity flag; taxonomy alignments are candidates for source review, not findings of a violation, conflict, or loss of separation.*

DAIDALUS natively evaluates an ownship against one or more intruders and returns aircraft-specific alert levels and maneuver guidance. GURU's contribution around it is event-level orchestration, hazard representation, and aggregation across traffic, with event classification and flight exposure maintained independently: mitigation can reduce an individual aircraft's conflict level without reducing the recorded severity of the originating debris event. This distinction preserves the causal relationship between the hazard, the response, and the resulting traffic state.

The replay in Figure 36 uses recorded traffic from NASA's Sherlock archive, with 42 selected flights evaluated at a one-second model step. Over the full event window, 21 recorded tracks intersected the modeled hazard bands; the system produced 8 accepted route alternatives, 1 mitigation, and 13 cases flagged for review, and it reports the added distance and fuel of the accepted alternatives as a window-rejoin estimate rather than a completed-trip distance or a deconflicted multi-aircraft plan. The event was classed as a debris-generating event at an advisory severity level under a proposed demonstration taxonomy. The alternatives are presented for review before action: no operational action is applied in the replay, recorded tracks are preserved, and authority remains with the Federal Aviation Administration and air traffic control.

### 7.6 Anomaly categorization

A second demonstrated capability is anomaly categorization. Each flagged anomaly is assigned a primary class and supporting secondary classes within a four-class taxonomy comprising data quality, operational, safety-relevant, and routing anomalies. The dashboard associates each classification with a confidence value and an evidence trail (Figure 37). This implements the human-understandability requirement by connecting an anomaly indication to its interpretation, supporting evidence, and candidate rule-aligned response. Further development will align these categories and recommended actions with NASA and Federal Aviation Administration terminology and procedures; the recorded-traffic replay in Section 7.5 is a first assessment, and evaluation under operationally representative conditions remains open. Collectively, these results demonstrate transfer of the platform's representation and agent organization to airspace decision support, with operational validation remaining an explicit research objective.

### 7.7 Toward an end-to-end workflow

Combining radar observations, atmospheric data, and debris simulations can improve situational awareness by showing where fragments may travel, when they could cross flight altitudes, and which routes may be affected. Explicit uncertainty regions help decision makers consider a range of plausible outcomes, while new observations can help narrow those possibilities. Connecting these predictions to aircraft conflict assessment could support earlier rerouting, targeted temporary restrictions, and clearer criteria for reopening airspace, reducing exposure while limiting unnecessary disruption. The current demonstrations establish useful components of this workflow; reliable operational risk reduction still requires validated predictions and avoidance assessment across the uncertainty range. The partition of authority is the one used in every other application: GURU proposes candidate responses, validated propagation and DAIDALUS confirm, and the controller decides.

## 8. Application IV: Hierarchical Supervision of Multi-Aircraft Teams

### 8.1 The control problem

Direct vehicle-level control does not scale to large uncrewed-aircraft teams. Under current U.S. civil small-UAS rules, one person may not manipulate the controls, serve as remote pilot in command, or act as visual observer for more than one aircraft at a time without a waiver (Operation of Multiple Small Unmanned Aircraft, 2026). This is a regulatory constraint rather than a universal technical limit, but it reflects the attention demanded by direct control. DARPA reported the same operational baseline for military uncrewed aircraft, at least one operator per vehicle, while its Collaborative Operations in Denied Environment (CODE) program demonstrated the alternative: one mission commander supervising multiple cooperating aircraft, including trials with communications offline and GPS unavailable (Defense Advanced Research Projects Agency [DARPA], 2019). At team scale, sequential human input cannot continuously choose a route, resolve a collision, assign an objective, and recover a broken radio link for every aircraft before the state changes again (DARPA, 2019; Adams et al., 2023). DARPA's OFFensive Swarm-Enabled Tactics (OFFSET) program considered teams of up to 250 air and ground robots and therefore treated swarm autonomy, rapid tactics generation, and human-swarm interaction as one problem (DARPA, n.d.). Its field studies later showed that one person could supervise 100 heterogeneous robots, but the measured workload repeatedly crossed an overload threshold (Adams et al., 2023). The design objective is therefore not to remove the human. It is to move the human from continuous piloting to mission intent, constraints, exception handling, and authorization, while the aircraft perform time-critical estimation, coordination, and motion locally. The scenario evaluated here is a simulated contested setting; the civil rule illustrates the attention constraint and is not a claim about any civil operation.

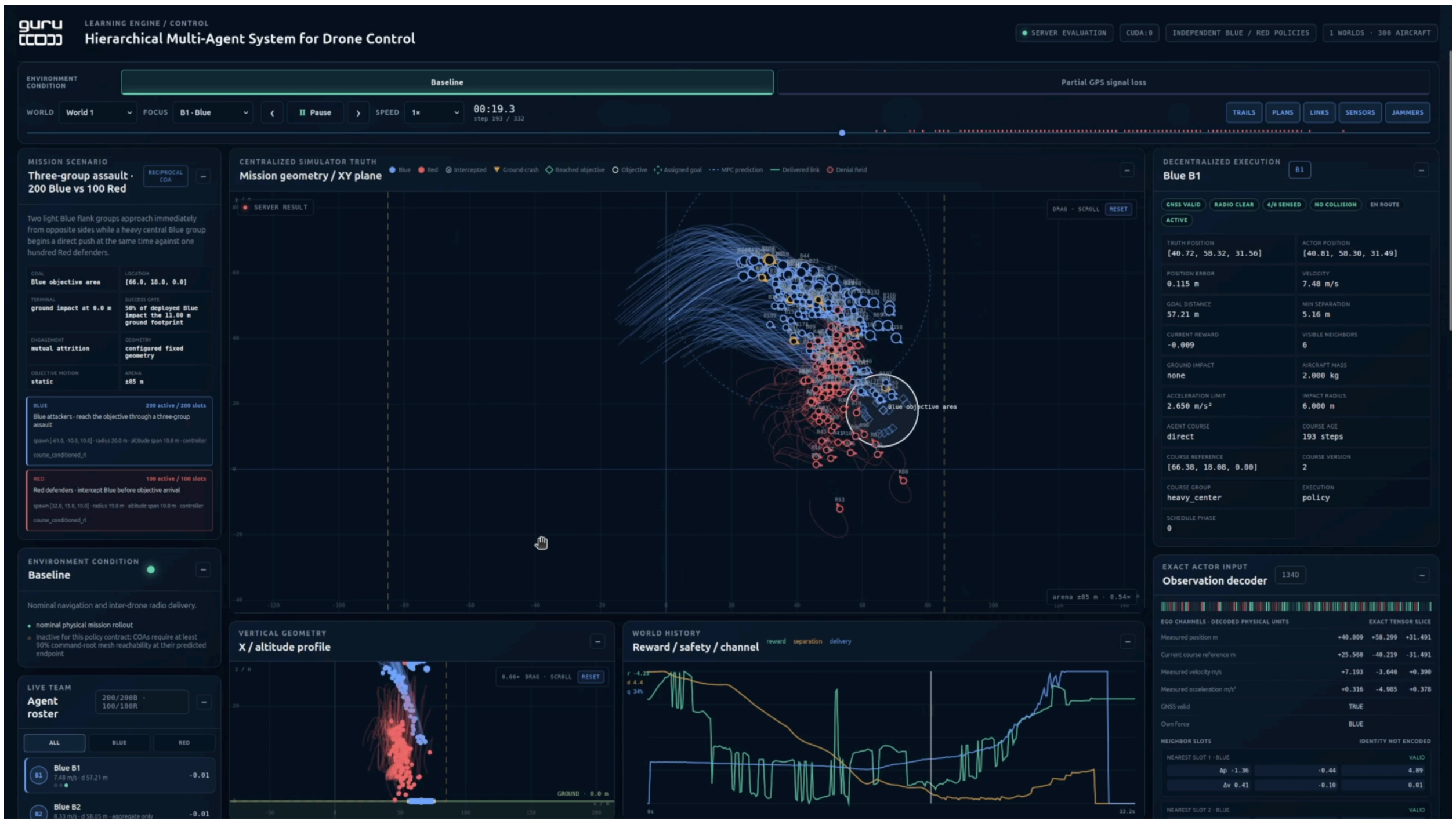

***Figure 38:*** *Interactive multi-aircraft evaluation interface. The dashboard combines synchronized map and altitude views with scenario controls, timeline playback, mission outcomes, communication and navigation-denial status, and per-aircraft diagnostics. Operators can inspect each agent's local observation, assigned course and goal, received messages, physical characteristics, policy or MPC output, and predicted trajectory, while selecting separate policies and checkpoints for each force and rerunning configurable Torch-based evaluation scenarios.* *https://youtu.be/9kXCKju5cyA*

## 8.2 Hierarchical control system

No single controller should solve mission planning, adversarial coordination, collision avoidance, and vehicle stabilization at the same rate or with the same information. We divide the problem by horizon and responsibility. A slow planning layer selects a course of action, target region, role, or subteam assignment. A faster action layer converts that intent into a short-horizon trajectory or bounded motion command. A navigation layer estimates state, executes the command, and reports deviations. Symbolic rules operate across these layers: they reject an option that lacks enough active aircraft or communication connectivity, preserve a minimum option duration, and force a defined fallback when no nominal option remains legal. Recent multi-agent work supports this decomposition: a high-level learned policy selected abstract tactical targets while decentralized model-predictive control produced dynamically feasible motion, improving sample efficiency and safety relative to end-to-end and shielding-only baselines in its benchmark (Studt and Schildbach, 2025).

This structure instantiates the GURU platform rather than adding a separate autonomy stack. In GURU terms, mission and course selection belong to the Planning layer; policy or MPC outputs belong to the Action layer; estimation and vehicle execution belong to the Navigation layer; and typed preconditions, limits, and fallbacks belong to the symbolic verification layer. The Learning Engine supplies the common training and comparison machinery for these agents, while model-based checks and runtime rules remain outside the learned policy's authority. This follows the paper's core division of labor: learned models propose and adapt; explicit rules and model-based checks constrain what may execute (Sections 3 and 4).

The intended deployed hierarchy is decentralized at inference. Each aircraft runs the same local policy function using its own observation, received messages, and persistent state. Centralized information is permitted during training, where it can reduce multi-agent non-stationarity without becoming a deployment dependency. This centralized-training, decentralized-execution pattern is established in multi-agent actor-critic research (Lowe et al., 2017). Distributed PPO supplies a complementary scaling mechanism: many simulated worlds collect experience in parallel and synchronize policy updates, an approach previously shown to scale reinforcement learning to billions of simulated navigation frames (Wijmans et al., 2020).

## 8.3 Prototype

The prototype implements a GPU-resident, three-dimensional, behavior-level multi-aircraft simulator in PyTorch, with the interactive evaluation interface shown in Figure 38. Every world stores position, velocity, acceleration state, force membership, activity, mission state, and communication state in batched tensors. An aircraft observation can contain measured ego motion, a relative goal, a fixed number of ally and opponent slots, mission geometry, its mass and impact radius, GNSS validity, and delayed message slots. The channel model supports finite radio range, fixed delay, packet loss, and spatial radio and GNSS denial. These effects are synthetic stressors, not a validated radio-frequency propagation model: jammers are static spherical fields, dropout is memoryless, and local MPPI presently treats sensed neighbors as constant-velocity obstacles. They can measure policy degradation under controlled faults, but they do not establish communications realism.

The learned controller is a parameter-shared PPO actor (Schulman et al., 2017). The actor uses only each aircraft's local observation at execution, while a centralized, permutation-aware critic may read the global training state. The current policy receives both its final goal and a fixed course label (such as direct, left flank, right flank, disperse, mesh hold, or encircle). During training, a small auxiliary term rewards progress toward the current course reference and penalizes unnecessary turning; it supplements, and does not replace, the mission reward. The same network can occupy either force role. A mixed opponent pool can freeze the Blue or Red force independently in each simulated world and train the remaining aircraft against scripted direct motion, adaptive pursuit, local MPPI, historical checkpoints, or self-play. The course is supplied rather than selected: a meta-policy can call the existing course-and-goal interface without changing the lower policy.

MPC provides both a training opponent and a model-based baseline. The local Model Predictive Path Integral (MPPI) controller samples action sequences on the accelerator, rolls them through the same point-mass transition used by the environment, scores goal error, terminal speed, control effort, smoothness, separation, collision, and boundary costs, and executes the first action before replanning. This follows the sampling-based, GPU-parallel MPPI formulation of Williams et al. (2017). A contract-aware mode additionally tracks reference knots, corridors, speed limits, arrival windows, and uncertainty-inflated separation. Above it, the hierarchical MPC baseline searches a finite adversarial game over maneuver profiles, maintains persistent subteams and roles, applies typed legality rules, and issues versioned per-aircraft contracts.

## 8.4 Results

The clearest learning result is a fixed-protocol checkpoint gate from the 16.78-million-world-step course-conditioned run. A world counts as a success when at least half of the deployed Blue aircraft reach the objective area. On 256 unseen worlds with fixed seed 88,001, checkpoints at 6.29, 10.49, and 16.78 million world steps increased the Blue target-arrival fraction from 0.148 to 0.202 to 0.607 and world success from 0 to 0.008 to 0.484. Repeating the final checkpoint on untouched seed 89,001 yielded 0.587 arrival and 0.434 success; the two-seed means were 0.597 and 0.459. A separate paired, four-seed multistage comparison improved world success from 0.561 for iteration I1 to 0.695 for the selected I2 response; objective fraction rose from 0.751 to 0.764 and Blue survival from 0.740 to 0.762. The same comparison also increased the fraction of worlds with a collision from 0.021 to 0.039, so it supports improved mission performance, not an unconditional safety improvement.

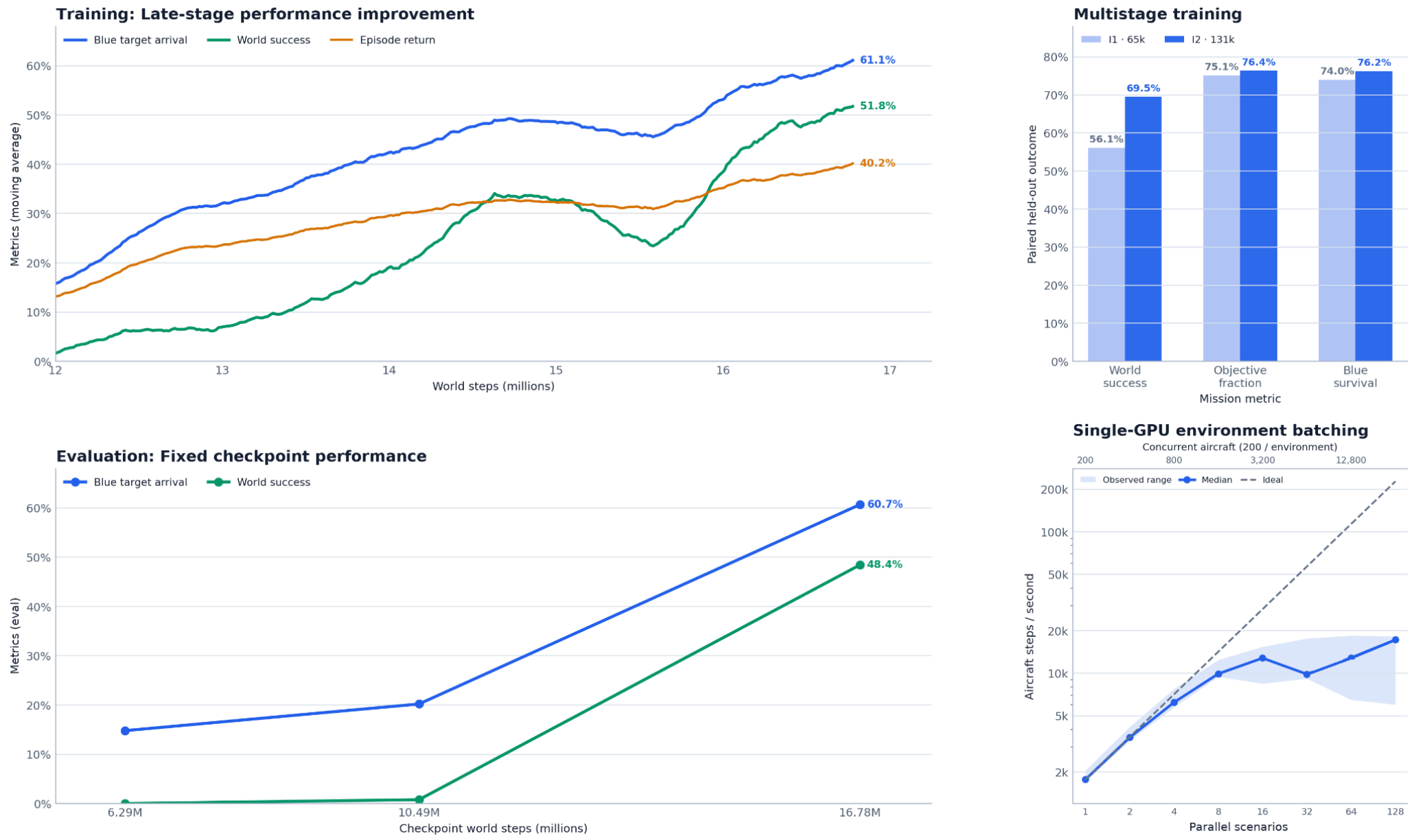

***Figure 39:*** *Multi-aircraft training and evaluation results. (Top left) Moving averages of Blue target arrival, world success, and episode return during late-stage training. (Bottom left) Fixed-checkpoint evaluation on 256 unseen worlds. (Top right) Paired held-out comparison of multistage training iterations I1 and I2. (Bottom right) Single-GPU throughput as the number of parallel 200-aircraft scenarios grows, with the observed range across repeats.*

This result also marks the current limit of verification in this application. In the present prototype, the symbolic rules govern which tactical options are legal and when a fallback must be invoked. Separation between aircraft is handled as a cost in the local MPPI controller rather than as a verified constraint, which is why the collision-world fraction can rise as mission performance improves. Promoting separation from a cost to a hard, independently checked constraint is the step that would bring this application under the same verification standard as the others.

The tensorized design also produced measurable within-GPU scaling. With 200 interacting aircraft per world, one GPU advanced the full hierarchical controller and environment at a median 1,774 aircraft-steps per second for one world and 17,248 aircraft-steps per second for 128 parallel worlds, or 25,600 concurrent aircraft and a 9.72-fold throughput increase. Throughput is close to ideal up to about eight parallel worlds and then saturates, with wide spread across repeats (Figure 39). The sweep used 40 timed steps after five warm-up steps and three repeats. These measurements show that goal- and course-conditioned behavior improved with training and that the simulator can evaluate many 200-aircraft worlds concurrently; they are single-GPU results, not leadership-system scaling.

Together, these results indicate that hierarchical autonomy can shift the operator from aircraft-level control to specifying mission goals, courses, and constraints for the team. Learned policies can adapt behavior, local MPC can ground intent into short-horizon motion, and symbolic verification can reject illegal options or invoke defined fallbacks before execution. Although the present evidence is simulation-based, this separation of responsibilities provides a credible path toward supervising large teams without requiring one human decision stream per aircraft.

# 9. Application V: Toward Microreactor Digital Twins and Supervised Autonomous Operation

Of the five application areas in this paper, the nuclear microreactor digital twin is the most demanding, and it is the sharpest test of the principle the architecture has followed throughout: that learned models accelerate and rank while validated physics confirms and decides. This workflow couples neutronics, thermal-fluid transport, thermomechanics, materials limits, shielding, and control-drum response into a single system, the kind of coupled, leadership-scale reactor multiphysics that exascale efforts have begun to make tractable (Merzari et al., 2023). An error in any one of those couplings carries a severe safety consequence.

The goal is a control-grade digital twin of a compact, transportable nuclear microreactor, together with a verified basis for operating it under human supervision. This section presents a research direction and the evidence assembled toward it: a population-scale characterization of the design space, validation against published reactor-physics and shielding benchmarks, and a working prototype of the anomaly-detection capability. The test case is an open reference microreactor in the helicopter-portable class that Tam Fortis Solutions is developing. Its configuration is assembled from microreactor designs that national laboratories have published and from the parameters Tam Fortis has itself made public, and the reactor detail used here does not exceed what Tam Fortis publishes; the company's detailed product design is not part of this work. The architecture, methods, and results are intended for the wider reactor-physics and computational-science communities, not for any single design.

## 9.1 Why this needs validated physics, not a proxy that replaces it

A nuclear reactor compact enough to be transported and operated where no crew is present, and where a command link may be degraded or absent, must carry out routine operation itself for stretches while a human supervises from a distance and retains the authority to intervene or shut it down. This is a human-on-the-loop regime, distinct from the human-in-the-loop approach in which a person approves each action. The regulatory expectations for such a system are already written. The Nuclear Regulatory Commission's October 2024 gap assessment for artificial intelligence expects physics-based models in safety-significant roles, its proposed microreactor licensing framework of May 2026 explicitly anticipates autonomous operation (U.S. Nuclear Regulatory Commission, 2024, 2026), and national-security guidance of June 2026 sets the assurance bar at reliable, robust, steerable, and controllable under rigorous test, evaluation, validation, and verification (National Security Presidential Memorandum 11, 2026). A recent operating-experience study prepared for the NRC Office of Nuclear Regulatory Research by Idaho National Laboratory reaches the same conclusion from the reliability-data side. In digital instrumentation and control systems the dominant common-cause concern is a systematic software fault, a deterministic error common to every identical copy that fails them together once it is triggered, and the established defenses are design diversity and fault-tolerant architectures that switch to an independent backup (Liu and Ma, 2026). That study also recommends validated digital twins, synthetic failure-scenario generation, and machine-learning analytics to strengthen reliability and risk modeling for these systems. GURU is built to that shape. A learned model that is the sole judge of its own output would be exactly such a single systematic-fault path, so the deciding path is validated physics, functionally and implementationally diverse from the learned proposer, and any out-of-envelope condition hands control to a conventional physics-based protective fallback. A proxy that replaces validated physics, serves as the sole judge of its own output, or becomes the safety basis does not meet these expectations.

Our approach rejects that design. The learned models we use rank, prune, compress, and accelerate which validated physics to run, and the validated solvers, the open codes OpenMC, NekRS, OpenFOAM, MOOSE, and Cardinal, confirm and decide (Romano et al., 2015; Fischer et al., 2022; Weller et al., 1998; Permann et al., 2020; Merzari et al., 2021).

**The partition.** This is where the paper's opening tension is resolved for a safety authority. It is the need for hybrid intelligence stated in Section 2.1, and the partition below is our implementation of it for reactor control. GURU does not blend learned and validated computation. It subordinates the first to the second. The designed control architecture is a strict separation of authority in three tiers. A learned layer proposes and ranks candidate subsystem actions, such as control-drum corrections. It is an approximation, it holds no safety authority, and it is the only tier where an approximation is allowed to live. A grounding tier compares each proposal and its predicted state against the live state, and when the prediction-error signal exceeds its validated threshold the proposal is demoted before it is eligible to act. The deciding tier is validated physics in reduced, control-grade form with a certified domain of validity and certified error bounds, established offline by the high-fidelity campaign. It holds veto authority and it is non-compensatory, so a high rank from the learned layer cannot buy down a constraint violation. This deciding tier is not a learned or statistical stand-in for the core and is never described as one. It is the validated physics itself, reduced to control grade, and nothing in the deciding path replaces it.

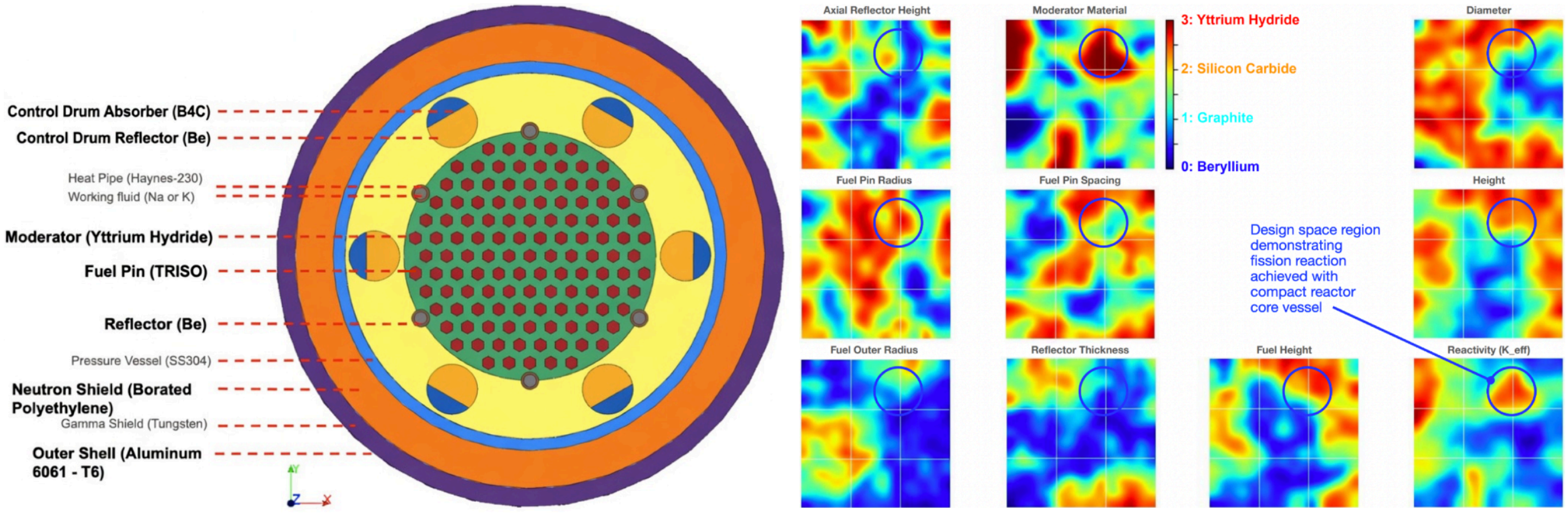

***Figure 40:*** *Reactor configuration and OpenMC neutronics exploration. Left: core cross-section showing TRISO fuel pins, moderator, heat pipes, control drums, reflector, pressure vessel, and shielding. Right: self-organizing maps derived from large-scale OpenMC simulations, illustrating interactions among key design variables and identifying regions that satisfy criticality and compactness objectives.*

**The affirmative mechanism.** Concretely, when the digital twin evaluates a control response to a transient or an anomaly, it does not let a learned model's self-assessment decide. It runs an ensemble of validated-physics variations of that exact control action, on the order of 100, and produces a statistical rating of the physical validity of the decision before any command reaches an actuator. The certified envelope it enforces is concrete, not abstract: reactivity bounds, fuel and TRISO-particle temperature limits, heat-pipe operability, and the conservation invariants, held together as one non-compensatory set. The controller is therefore anchored to live physics at every step, never to a model standing in for physics. Any action that violates a hard limit, or any condition that leaves the safe envelope, hands the plant to a conventional physics-based protective fallback and escalates to the human supervisor. This same mechanism is the reason the work requires leadership-class computing. Inference for a single reactor is cheap. Generating the coupled-physics database, training the world model and the control policies, and running these physics-anchored validation ensembles at the scale needed to establish and certify the controller's envelope are not.

## 9.2 Characterizing the design space

The first stage is a leadership-scale characterization of the design space, built on a precursor already executed on the Frontier supercomputer. More than 100,000 neutronics criticality evaluations were run there with the open Monte Carlo transport code, including a 5,000-case Latin hypercube exploration over 18 design variables that produced a clean inverse relationship between core diameter and external dose and identified critical configurations, with a best single configuration whose effective multiplication factor was about 1.023. These transport calculations use the ENDF/B-VIII.0 evaluated nuclear data library (Brown et al., 2018) and a performance-portable Monte Carlo implementation that runs across Intel, NVIDIA, and AMD GPUs (Tramm et al., 2024). A learned screening model trained on that data is used to accelerate the search through a much larger space under multi-objective optimization. It only ranks and prunes which configurations are worth a full evaluation. Feasibility is decided solely by a validated transport confirmation, never by the screening model alone, so the learned stage accelerates which physics is run and never replaces the physics that decides. The scientific target is to map whether the feasible region reaches the simultaneous corner of small diameter, low external dose, and reliable criticality that, to our knowledge, no published analysis has shown can be reached at once (Figure 40).

The same screening approach extends to the shielding stack. Learned models predict the external dose and the multiplication factor for each candidate material layout, and an optimizer searches for configurations that hold criticality near target while lowering dose (Figure 41).

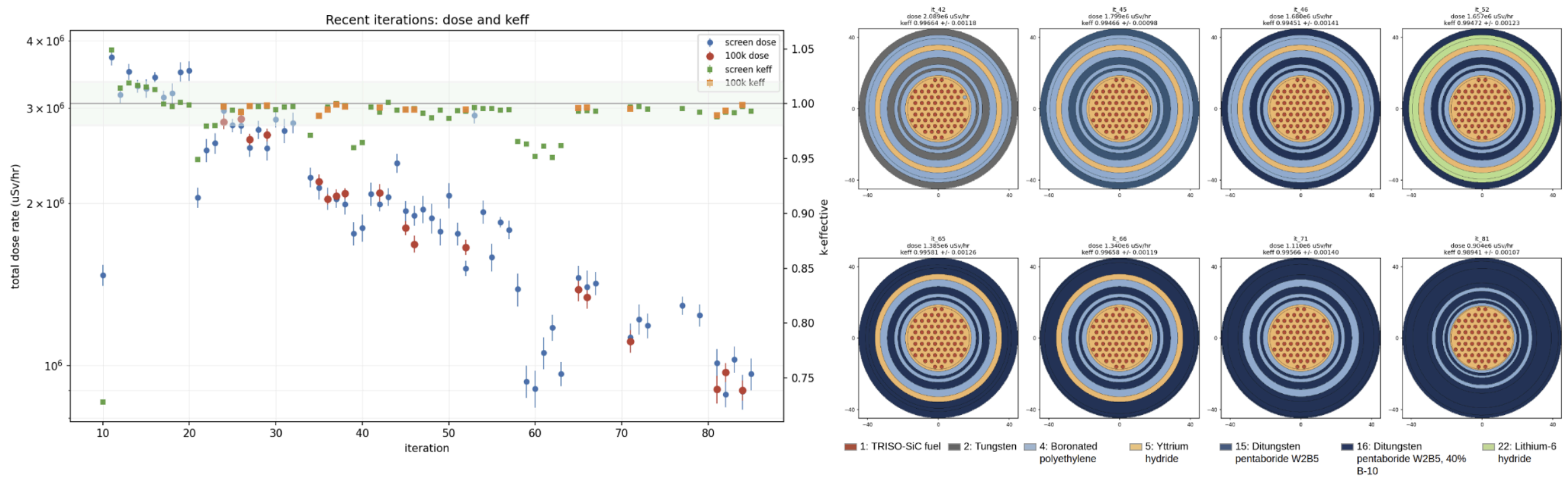

***Figure 41:*** *Learned models predict dose rate and the effective multiplication factor for each candidate shielding layout, while the optimizer searches material configurations that keep the multiplication factor near target and reduce external dose. Later iterations reach lower-dose designs while keeping the reactor viable.*

## 9.3 Validation against published benchmarks

Trust in these results rests on validation against established reactor-physics and shielding benchmarks. The open transport code was compiled specifically for the leadership systems, because prebuilt binaries scaled poorly, and was validated against a suite of benchmarks with public experimental data. On the initial-criticality benchmark of the HTR-10 pebble-bed test reactor, the model reproduced the measured effective multiplication factor of 1.01310, plus or minus 0.00030, to within about 59 pcm (59 units in the fifth decimal place), which is well inside the acceptance band, with the small positive bias consistent with the absence of modeled structural void channels. The model also tracked measured dose across more than four orders of magnitude on a reactor skyshine benchmark (Figure 42b), followed the published criticality trend of the KRUSTY reflector-height benchmark (Figure 42c), and reproduced a 14 MeV slab shielding benchmark within acceptance criteria (not shown). Optimized hexagonal-lattice configurations reached effective multiplication factors slightly above 1, demonstrating feasibility under realistic constraints, while simpler circular pin arrangements did not, because of reduced neutron economy.

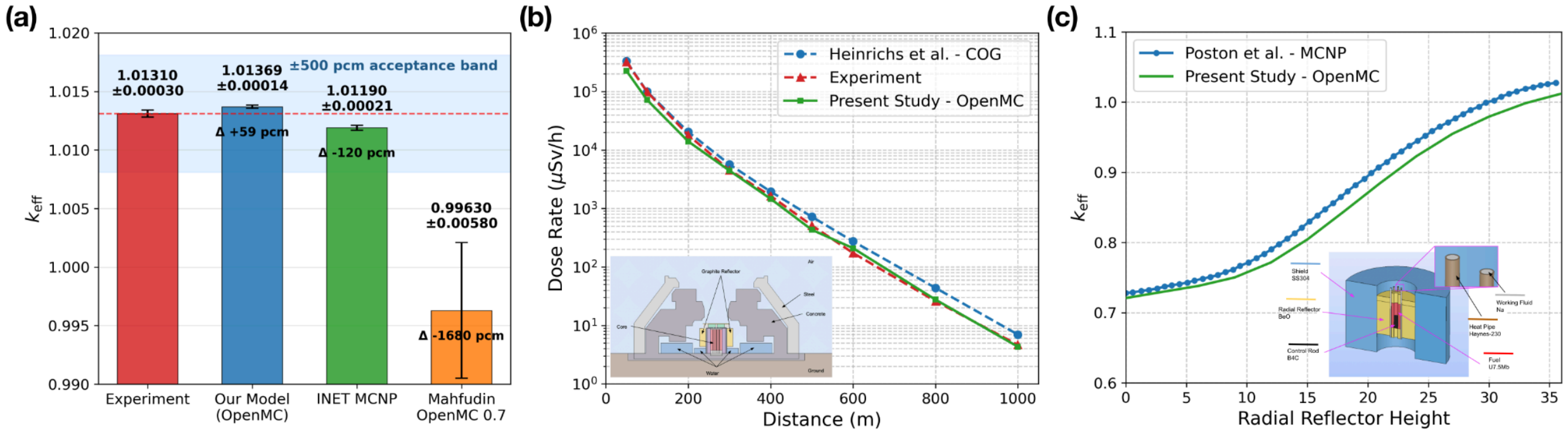

***Figure 42:*** *OpenMC simulations validated against published experimental and computational reactor datasets. (a) HTR-10 initial criticality: the OpenMC keff of 1.01369 plus or minus 0.00014 is +59 pcm from the IRPhEP benchmark and remains within the plus or minus 500 pcm acceptance band. (b) Baikal-1 reactor skyshine: OpenMC reproduces the measured dose-rate trend from 50 m to 1 km, spanning more than four orders of magnitude and consistent with the COG result of Heinrichs et al. (2022). (c) KRUSTY radial-reflector-height benchmark: OpenMC follows the published Poston et al. MCNP keff trend and captures the approach to criticality as reflector height increases.*

## 9.4 Toward the control-grade twin and supervised autonomy

Beyond characterization, the plan is to couple neutronics, thermal-fluids, and thermomechanics for a reference configuration, use the resulting high-fidelity transient data to train JEPA world models, including video and graph-structured variants that match the network of components (Skenderi et al., 2025), and produce a twin that runs faster than real time and reproduces held-out benchmark transients within validation tolerance. The very low thermal inertia of a roughly 100 kilowatt-thermal core makes faster-than-real-time prediction both necessary and achievable. The world model must predict across coupled timescales, from milliseconds for control-drum and flux transients to months and years for fuel depletion. Where the world model is uncertain, the prediction-error channel becomes the detector, which is exactly the regime, abrupt and non-stationary transients, where the identifiability guarantees of the world model are known to weaken. In that regime the trust inverts. As the live state leaves the region where the reduced control-grade physics is certified, the prediction-error signal rises, the learned proposal is demoted automatically, and the deciding tier holds to the most conservative certified bounds and drives the core toward the validated subcritical safe state. The learned layer holds the least authority it ever has at exactly the moment the validated physics holds the most. The final stage will train reinforcement-learning and model-predictive controllers against the twin and a large library of faults, wrap them in the runtime assurance gate and the validated-physics ensemble described above, and report learned-model performance and verified-system performance separately, because a controller that is accurate on average can still be correctly blocked when it proposes an action outside the verified envelope. That distinction, not the average accuracy of any model, is the heart of the safety case. This separation also answers a known limitation in digital-system reliability practice, that the probability of a systematic software fault cannot be reliably quantified (Liu and Ma, 2026), so the average accuracy of a learned controller is not admissible as the safety basis and the validated-physics gate is.

**Early evidence of the detection capability.** An unsupervised prototype already demonstrates the anomaly-detection idea on heat-pipe microreactor operating traces. Using power and average fuel and moderator temperatures from a nominal operating sequence, the prediction-error channel of a JEPA world model produces anomaly scores against warning and critical thresholds, and it is compared against standard baselines such as principal-component reconstruction, isolation forest, local outlier factor, and a one-class support vector machine. The prototypes used here, drawn from a kilowatt-class space reactor experiment (Gibson et al., 2018; Poston et al., 2020; Miao et al., 2025), a heat-pipe microreactor concept (Stauff et al., 2024), and a control-drum neutronics study, are small-scale demonstrations of the physics-anchored control, anomaly, and world-model methodology.

## 9.5 Reactor modeling and predictive control

A second microreactor prototype shows the control side of the same methodology. The goal is a compact reactor that fits within a four-foot envelope and delivers a nominal 100 kilowatts while staying controllable through startup, shutdown, and power changes. Power is regulated by rotating control drums around the core. Each drum is mostly neutron-reflecting, with one absorbing sector. Turning the reflective side toward the core returns more neutrons and raises reactivity; turning the absorbing sector toward the core lowers it. The drum angle is the main control variable.

To characterize the drums, hundreds of OpenMC neutronics runs sweep the drum angle and estimate the effective multiplication factor, the prompt neutron generation time, and the delayed-neutron fractions for each configuration. Because Monte Carlo output carries statistical noise, the raw results are fit to smooth angle-dependent curves that give the kinetic parameters the transient model needs.

During simulation, the current drum angle passes through the fitted neutronics models. The predicted multiplication factor becomes reactivity, and the kinetic parameters drive a point-kinetics model that propagates reactor power and delayed-neutron precursors. A single-temperature thermal model adds negative feedback: temperature rises with power and relaxes toward a sink, and the temperature state feeds back as negative reactivity, which represents the stabilizing effect of heating in reduced-order form.

This reduced-order model is fast enough for model-predictive control, and it is physics in reduced form, not a learned approximation of it. Its kinetic parameters are taken from the OpenMC neutronics runs and the thermal feedback is a physical relation, so the rollout the controller scores against is physics, reduced for speed, rather than a statistical stand-in for the core. In the deployed architecture this reduced physics is trusted only inside its certified domain of validity, and the validated-physics ensemble of Section 9.1 remains the deciding gate at the actuator. At each step the controller samples many future drum-motion sequences, rolls each one forward through the point-kinetics and thermal model, and scores it on tracking error, temperature and safety limits, and smoothness of motion. It applies only the first move, advances the state, and replans. Because the reactor environment runs in Torch, thousands of candidate futures are evaluated in parallel on a GPU, which makes the receding-horizon loop fast enough for interactive use. The prototype uses simplified data-driven neutronics fits and a lumped thermal model; a production controller would add larger multiphysics datasets, higher-fidelity thermal-hydraulics, uncertainty quantification, actuator and sensor models, fault handling, and validation against high-resolution simulations on systems such as Frontier or Aurora (Figures 43 and 44).

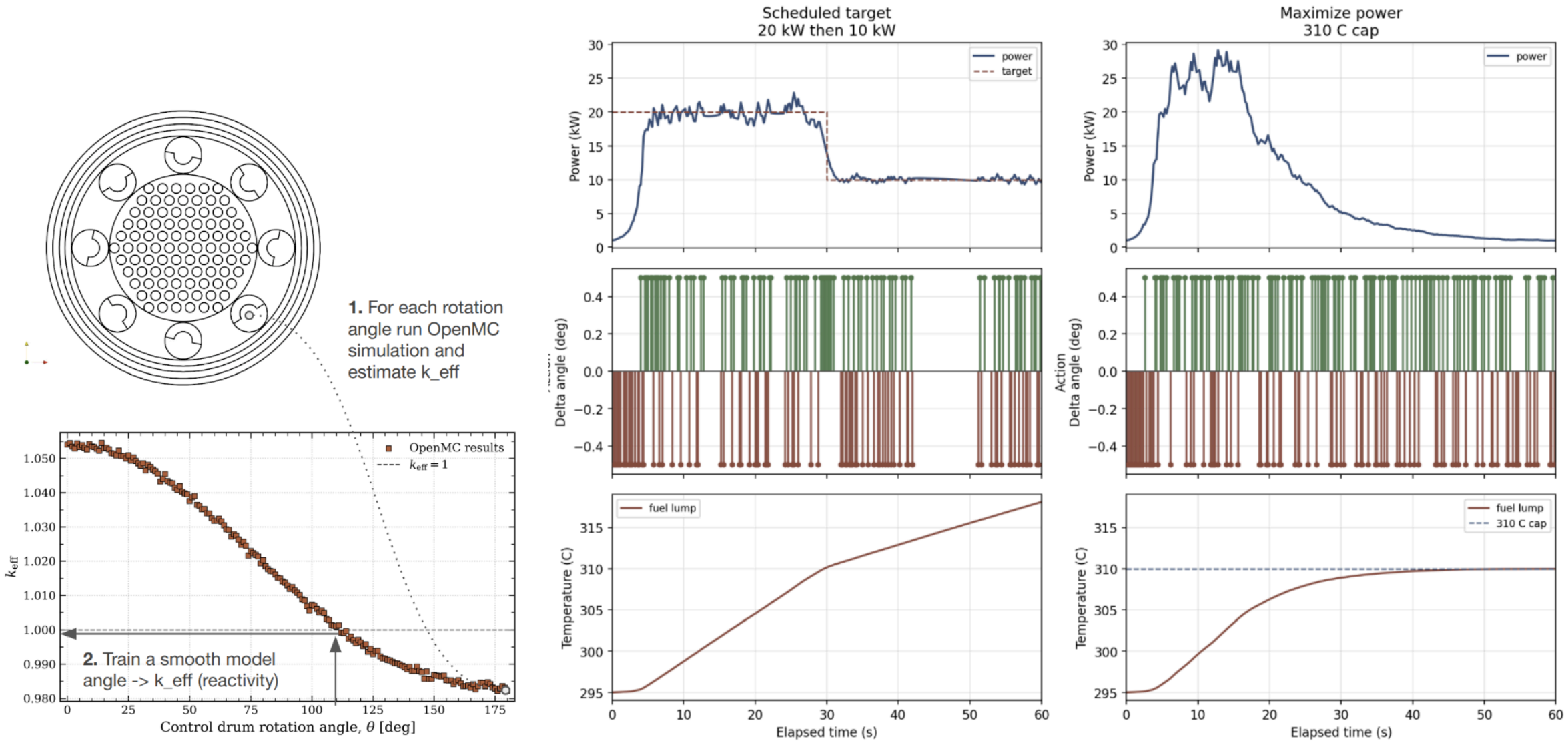

***Figure 43:*** *OpenMC drum-angle runs are fit to a smooth reactivity model used by the point-kinetics controller. The MPC then chooses drum rotations to track a scheduled power target or to maximize power while holding a fuel-temperature cap.*

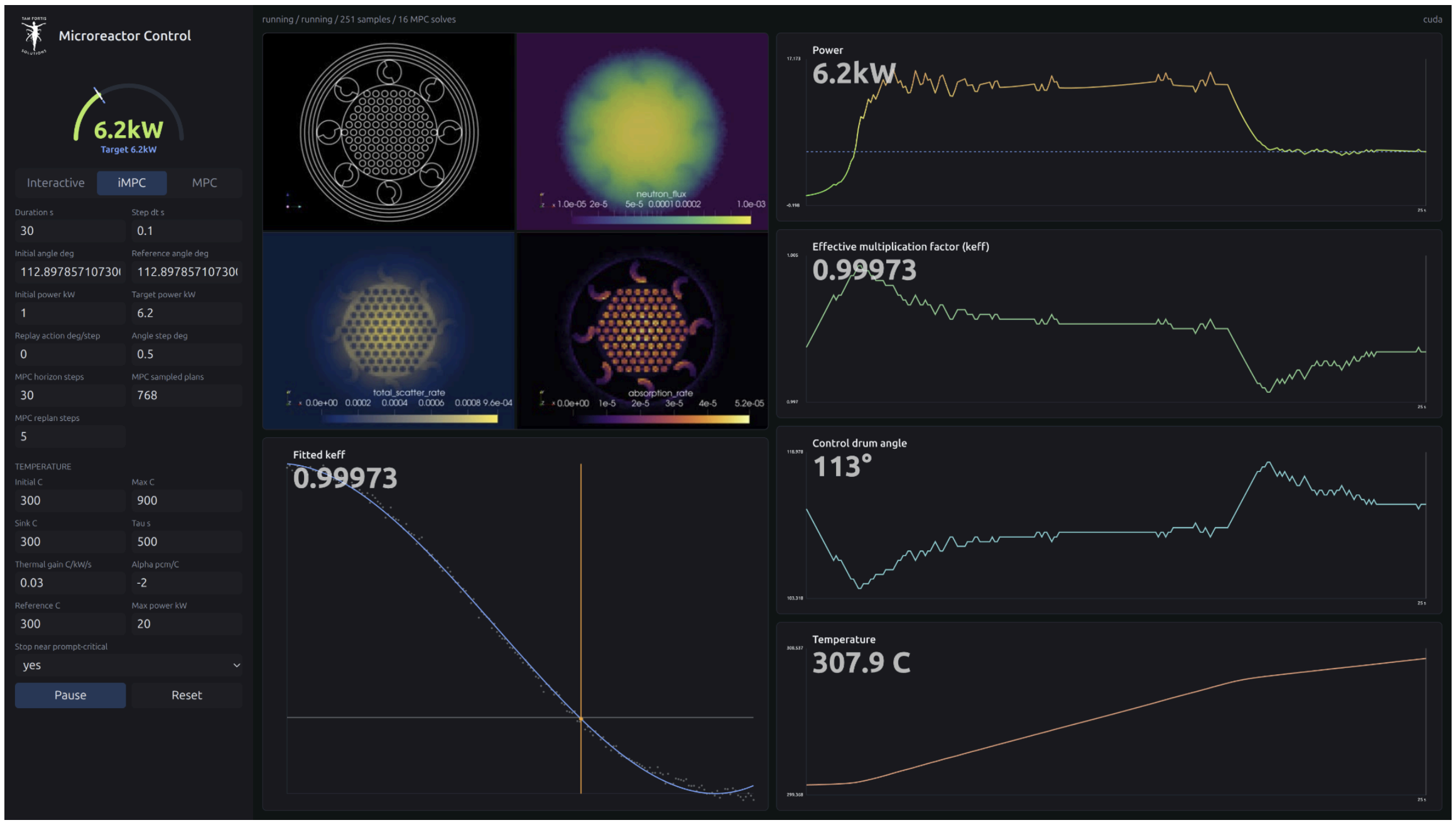

***Figure 44:*** *Interactive interface for the reactor-control simulation. The dashboard shows reactor power, the effective multiplication factor, the control-drum angle, and the fuel temperature alongside the core visualizations, and lets a user drive the controller live. Video: [https://youtu.be/w5S9YC7DB_c](https://youtu.be/w5S9YC7DB_c)*

## 9.6 The safety case in reactor control

In the reactor control setting the model-predictive controller proposes candidate control-drum motions at each step, and the point-kinetics and thermal model scores each candidate against the scheduled power target and the hard reactivity and temperature limits. The safety case is the designed behavior at the actuation gate. When the highest-ranked candidate, whether sampled by the controller or proposed by a learned policy in the full design, would drive reactivity or temperature past a hard limit, only a physics-validated candidate is eligible to reach an actuator, so an out-of-envelope proposal is discarded rather than executed, the conventional physics-based protective action is applied, and control escalates to the human supervisor. The design reports learned-model performance and verified-system constraint satisfaction separately, because a controller that is accurate on average can still be correctly blocked when it proposes an action outside the verified envelope, and that separation, not the average accuracy of any model, is the heart of the safety case. The model surfaces a candidate faster than any human could, in a place no human can reach, but a forbidden candidate is discarded rather than executed, and the validated physics decides what counts as safe. This is the designed control-grade realization of the mechanism in Section 9.1. The reduced-order prototype of Section 9.5 demonstrates the controller and the physics-scored envelope, while the distinct validated-physics rejection gate is part of the control-grade design and is reported here as a research target rather than a measured deployment result, consistent with the limitations stated in Section 12.

# 10. Results

## 10.1 The updated CFD workflow

This section reports the results that span more than one application or that require the full training-factory pipeline: the updated CFD workflow, the conjunction-risk comparison with its training-stability study, and the cross-domain reactor anomaly check. The software-control, airspace, and multi-aircraft results are reported in Sections 5.5, 7.5, and 8.4, and Section 12 gathers every result against requirements R1 to R5.

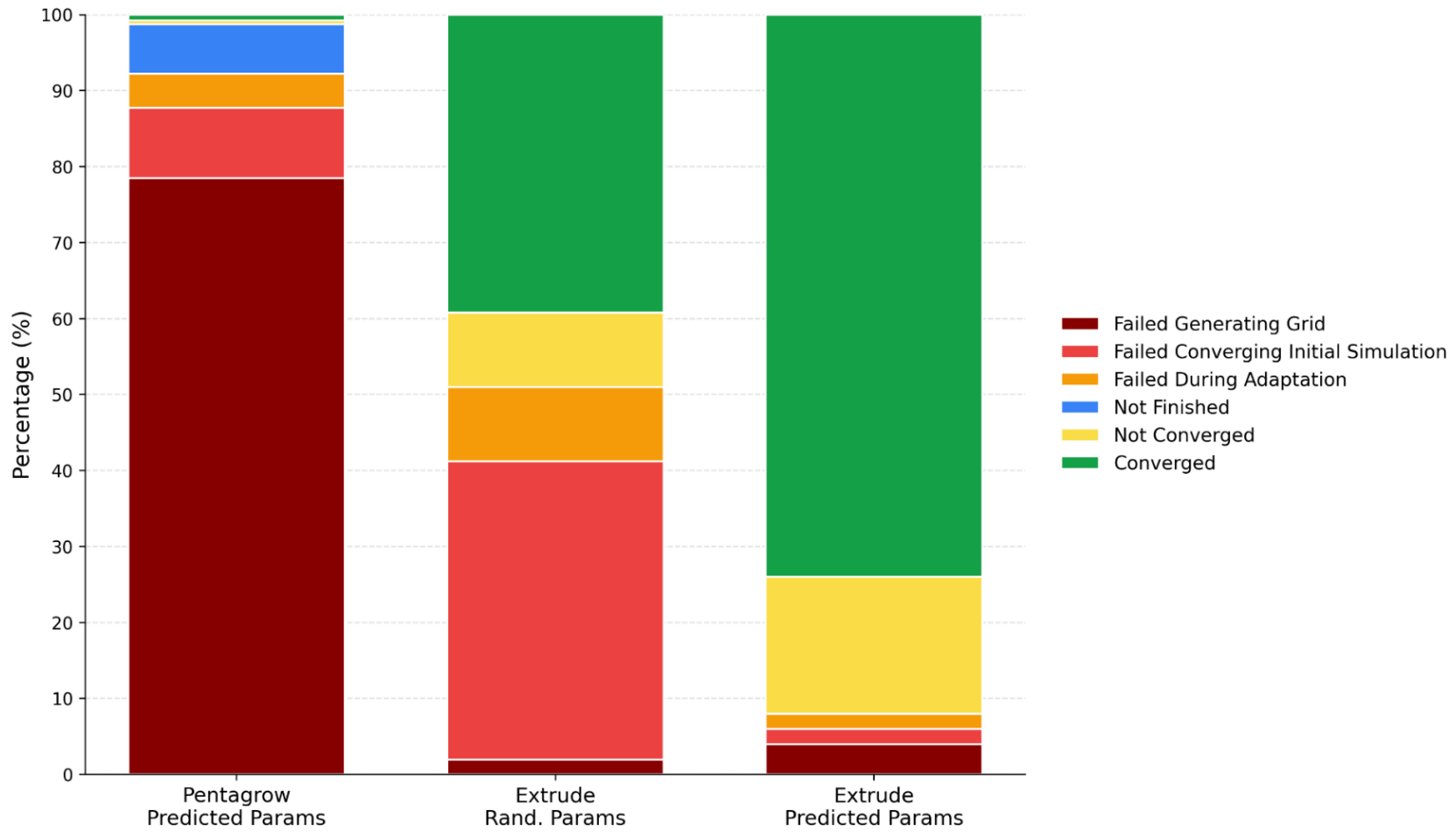

***Figure 45:*** *CFD simulation outcomes by grid generator and parameter selection. Stacked outcome shares for the earlier Pentagrow pipeline with predicted parameters, the extrusion-based generator with random parameters, and the extrusion-based generator with learned parameter selection. Outcomes are failed grid generation, failed initial solve, failed during adaptation, not finished, not converged, and converged. The rightmost configuration is the headline result in the text: 100 cases, one attempt each, of which 92 completed and 74 converged.*

The numerical results show that the updated workflow substantially improves the production of usable CFD simulations on previously unseen aircraft geometries (Figure 45). All three configurations use SU2 as the flow solver. The earlier pipeline, an AI grid generator built on Pentagrow, failed at grid generation on 78.5 percent of cases and reached a converged solution on only 0.75 percent. Replacing the grid stage with an extrusion-based generator and random parameter selection raised the converged rate to 39.22 percent, with a grid-generation failure rate of 1.96 percent. Adding the learned parameter selection on top of the extrusion generator raised the converged rate to 74 percent, with 92 percent of cases completing.

The end-to-end behavior of the final workflow is the headline result. The evaluation comprised 100 CFD cases on unseen geometries, each run once with the learned parameter selection and with no retries. Of the 100 cases, 92 ran to completion without a workflow failure and 74 converged; the remaining 18 completed cases ran through every stage without meeting the convergence criterion, and the 8 failures were distributed across the grid-generation, solve, and adaptation stages rather than concentrated in one. The workflow can take a predictor-corrector step when a case fails, in which the learned predictor proposes revised settings and the solver again decides convergence on a rerun (Section 4.3), but that step was not exercised in this measurement, so the reported rates are single-attempt rates and do not include recovery through retries. The number of distinct geometries among the 100 cases was not recorded. The geometry-preparation figures in Section 5.4 are prior peer-reviewed results, and the 13 percent convergence rate in Section 5.3 describes the earlier 15,000-case Latin-hypercube campaign. The single most important comparison is the converged rate on unseen geometries, which rose from under 1 percent with the earlier pipeline to 74 percent with the updated one.

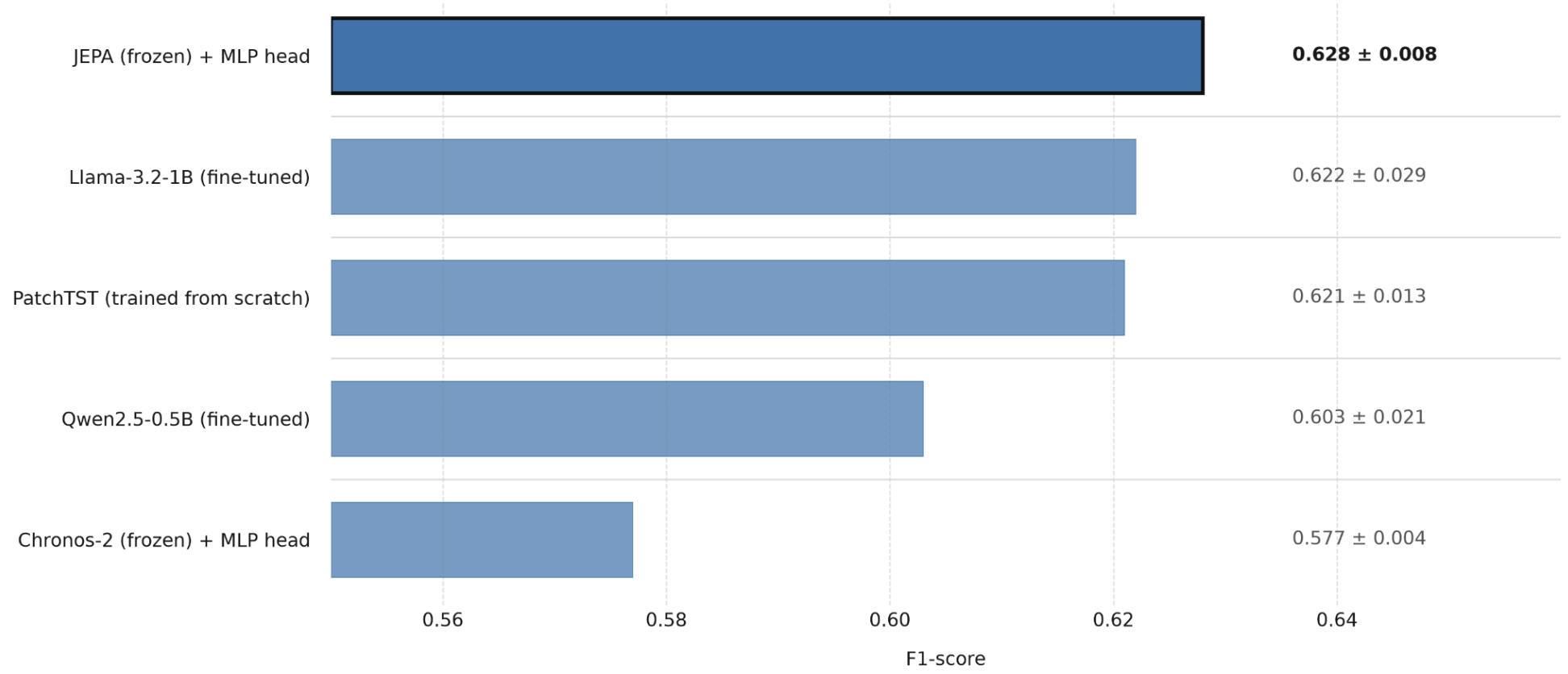

***Figure 46:*** *Frozen JEPA matches or exceeds configuration-matched baselines on conjunction risk, with seed-stable performance. Mean binary F1-score over 10 seeds on the shared test split, with every method trained under the same budget (batch size 96, 1,000 training steps); labels show mean ± standard deviation. The frozen JEPA encoder with a small MLP head scores highest, and its low seed-to-seed variance shows the result does not depend on a lucky seed, even though none of its backbone weights are trained.*

## 10.2 Frozen JEPA embeddings match or exceed configuration-matched baselines

Under a shared training budget (effective global batch size 96, 1,000 steps) and averaged over 10 random seeds, a lightweight MLP head on frozen OrbitGuard JEPA pair embeddings attained the highest mean binary F1 (0.628 ± 0.008) for emergency-reportable conjunction events in space, matching a fully fine-tuned Llama-3.2-1B (0.622 ± 0.029) and a from-scratch PatchTST (0.621 ± 0.013) within seed-to-seed variation and exceeding a fine-tuned Qwen2.5-0.5B (0.603 ± 0.021) and frozen Chronos-2 embeddings with the same head (0.577 ± 0.004); all values are mean ± standard deviation across seeds (Figure 46). Beyond holding the top mean score, the decisive result is stability across seeds: the two approaches that keep the backbone frozen and train only a small MLP head, JEPA (± 0.008) and Chronos-2 (± 0.004), are effectively seed invariant, with seed-to-seed variance several times smaller than the fully fine-tuned language models (Llama-3.2-1B ± 0.029, Qwen2.5-0.5B ± 0.021) and the end-to-end PatchTST (± 0.013).

This indicates that the frozen embedding plus MLP head recipe is a stable methodology whose performance does not depend on a lucky seed, and pairing that recipe with JEPA embeddings is what converts the stability into the leading result: JEPA delivers the best mean F1 while remaining among the most seed-robust methods, all without ever updating its backbone, whereas the language models are fully fine-tuned and PatchTST is trained end-to-end. This is consistent with recent evidence that a strong frozen predictive encoder transfers to downstream tasks without task-specific backbone training (Terver et al., 2026).

## 10.3 Training stays stable under SIGReg, where EMA targets often collapse

The best-scoring JEPA model in Figure 46 was pretrained with an exponential-moving-average (EMA) target encoder, a common heuristic for preventing trivial solutions in joint-embedding pretraining (Assran et al., 2023). We do not read this result as evidence that EMA targeting is the better choice.

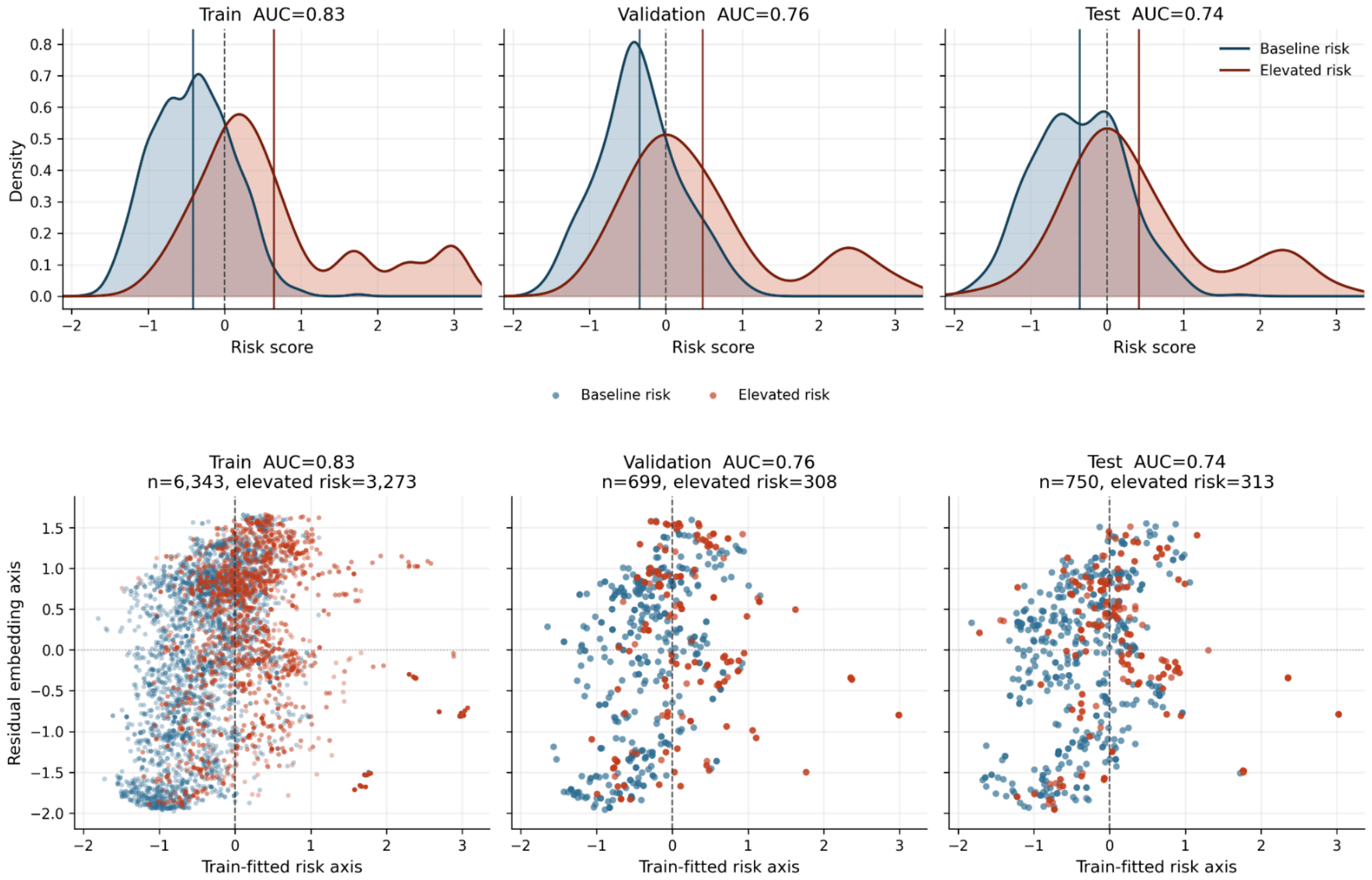

***Figure 47:*** *JEPA embeddings expose a conjunction-risk direction fit on the training split (labeled train-fitted in the panels). A class-balanced logistic probe was fit using only the training-set embeddings and then applied unchanged to validation and test pairs. (Top) Risk-score distributions for Baseline risk and Elevated risk pairs across train, validation, and test splits. (Bottom) Two-dimensional projection of the same pairs using the risk axis fit on the training split and the leading residual embedding direction after subtracting the risk-axis component. The held-out validation and test panels show that the learned risk ordering also appears for conjunctions not used to fit the probe.*

EMA stabilization is documented to be fragile, and EMA runs can still drift toward low-variance, collapsed embeddings that satisfy the prediction loss while losing discriminative structure. We therefore also trained a variant regularized with SIGReg, a penalty that checks, along many randomly chosen directions that are redrawn during training, that the batch's embeddings spread like a standard bell curve; enforcing this in enough directions keeps the whole representation spread out, so it cannot collapse (Balestriero and LeCun, 2025; Maes et al., 2026). This variant removes the separate EMA target encoder entirely and computes the target with the same trainable encoder. In our experiments it trained stably and its loss generally did not collapse, in contrast to the collapse behavior we observed under EMA-only training. We therefore report the EMA model as our strongest downstream configuration while identifying SIGReg as the more robust route to stable pretraining, consistent with recent end-to-end joint-embedding models that drop EMA and stop-gradient heuristics in favor of an explicit distributional regularizer (Maes et al., 2026).

## 10.4 A risk direction fit on training pairs carries over to held-out conjunctions

The frozen JEPA pair representation contains a linearly accessible conjunction-risk signal. To quantify this, we standardized and L2-normalized the frozen embeddings, reduced them with PCA fit only on the training split, fit a class-balanced logistic probe on the resulting training coordinates, and then applied the same trained transform and probe to validation and test pairs (Figure 47).

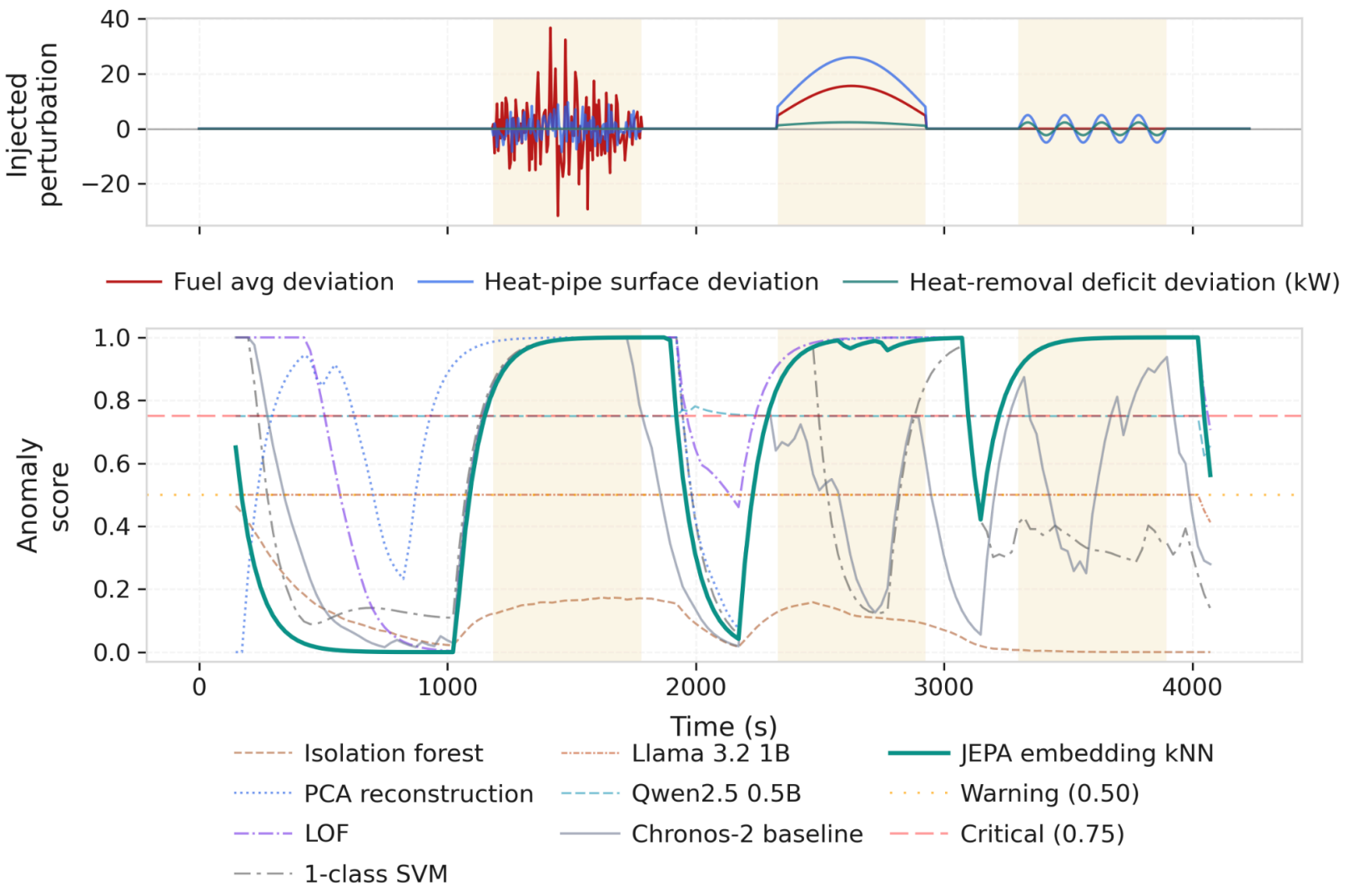

***Figure 48:*** *JEPA anomaly score tracks each injected reactor disturbance. Response of unsupervised anomaly detectors to controlled perturbations in heat-pipe microreactor data. (Top) The injected thermal and heat-removal deviations over time. (Bottom) Normalized anomaly scores from the JEPA embedding-kNN detector, classical detectors, language-model baselines, and Chronos-2, with shaded bands marking injected-disturbance windows. The JEPA score rises during each disturbance and falls back toward nominal between events.*

We label non-emergency-reportable events as Baseline risk and emergency-reportable events as Elevated risk; these labels describe the dataset classes used for this visualization. The trained probe defines a risk axis in the embedding space, and each pair's risk score is the normalized signed distance from the probe boundary, with positive values pointing toward the Elevated-risk side. In the top panels, the risk-score distributions show that Elevated-risk pairs shift toward positive values, and the same pattern remains visible in the held-out validation and test splits, whose pairs were not used to fit the preprocessing, PCA, or logistic probe. In the bottom panels, the same pairs are projected onto the risk axis fit on the training split and a residual embedding axis obtained after subtracting the risk-axis component, showing that additional representation structure remains beyond the one-dimensional risk score. This linear, out-of-sample separability echoes recent results that joint-embedding predictive models can linearly recover latent world structure (Klindt et al., 2026).

## 10.5 A sanity check: the same JEPA approach detects anomalies in reactor operating data

As a cross-domain sanity check, we applied the same self-supervised JEPA recipe to a second domain, reactor operations. We trained a JEPA encoder only on nominal multivariate windows from reference heat-pipe microreactor traces, with no labeled faults, and evaluated it by injecting controlled temperature-sensor, heat-removal, and power-oscillation disturbances. An embedding-space nearest-neighbor detector built on the frozen JEPA representation raises its anomaly score during each injected disturbance and returns toward nominal between events, giving a clear temporal signal. Classical detectors also flag many disturbances, especially when a fault directly shifts measured channels, while prompt-based language-model baselines discriminate less reliably (Figure 48). This is a sanity check that the recipe transfers, not a deployment-grade reactor result.

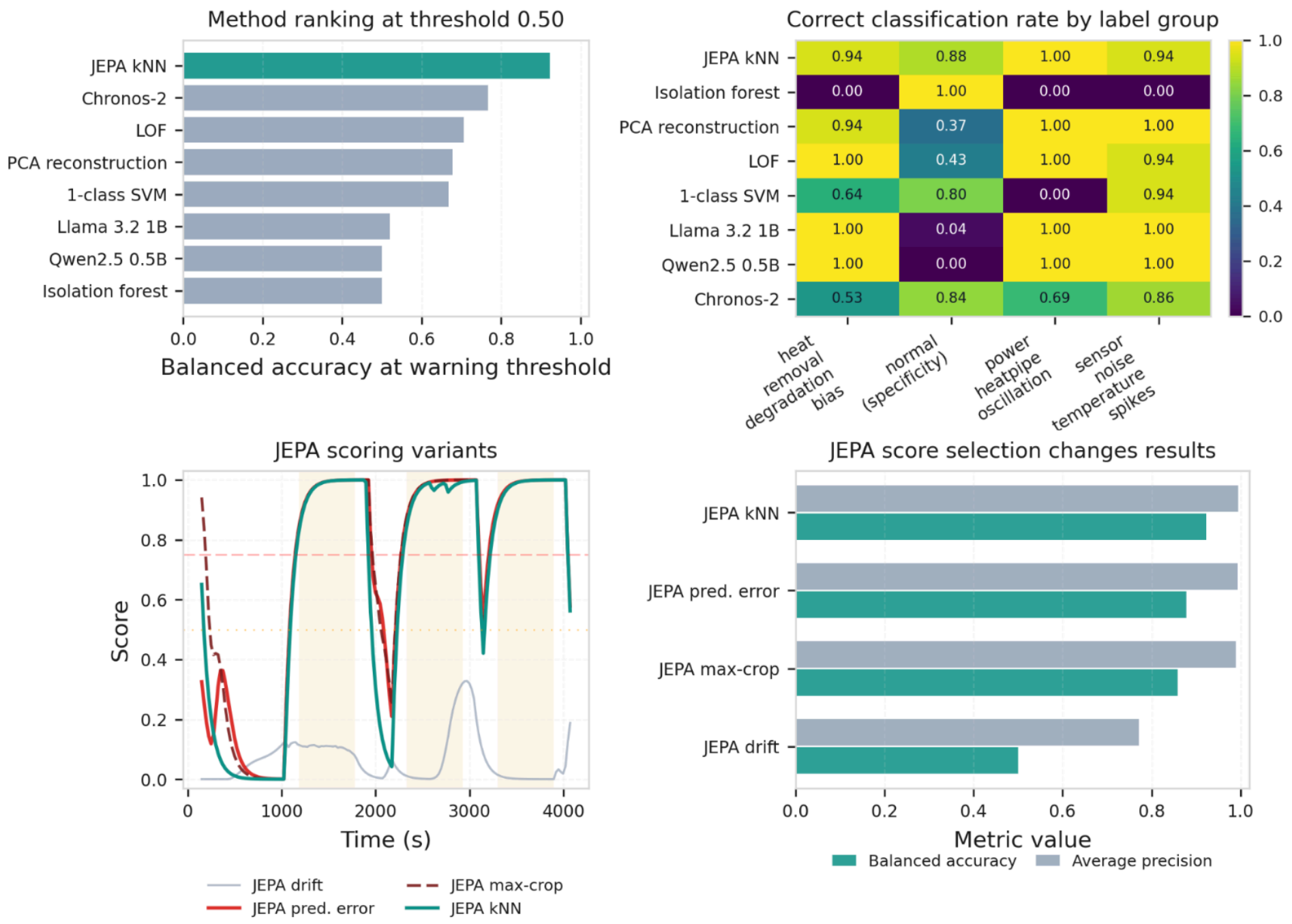

***Figure 49:*** *Quantitative comparison of anomaly-detection methods and JEPA scoring rules. (Top left) Balanced accuracy at the warning threshold by method. (Top right) Correct-classification rate per fault group. (Bottom left) Temporal anomaly scores for four JEPA scoring variants. (Bottom right) Balanced accuracy and average precision for each scoring rule. Embedding kNN gives the best and most consistent balance across fault families.*

Comparing four scoring rules built from the same encoder (embedding drift, latent prediction error, multi-crop maximum prediction error, and embedding kNN), the embedding-kNN rule gives the best balance of anomaly sensitivity and nominal specificity at the warning threshold and stays consistent across fault families (Figure 49). The key lesson is that the learned representation alone does not determine performance: the same encoder yields markedly different results depending on the scoring rule, reference set, and threshold. Reference-set construction, scoring-rule selection, and threshold calibration must therefore be treated as core parts of the JEPA methodology, echoing recent world-model findings that downstream success depends jointly on representation, prediction objective, and inference procedure rather than encoder architecture alone (Terver et al., 2026; Maes et al., 2026).

## 11. Scaling and Infrastructure

All of the results above depend on training and simulation at leadership scale, carried out on the Frontier and Aurora supercomputers under a DOE ALCC leadership-computing allocation.

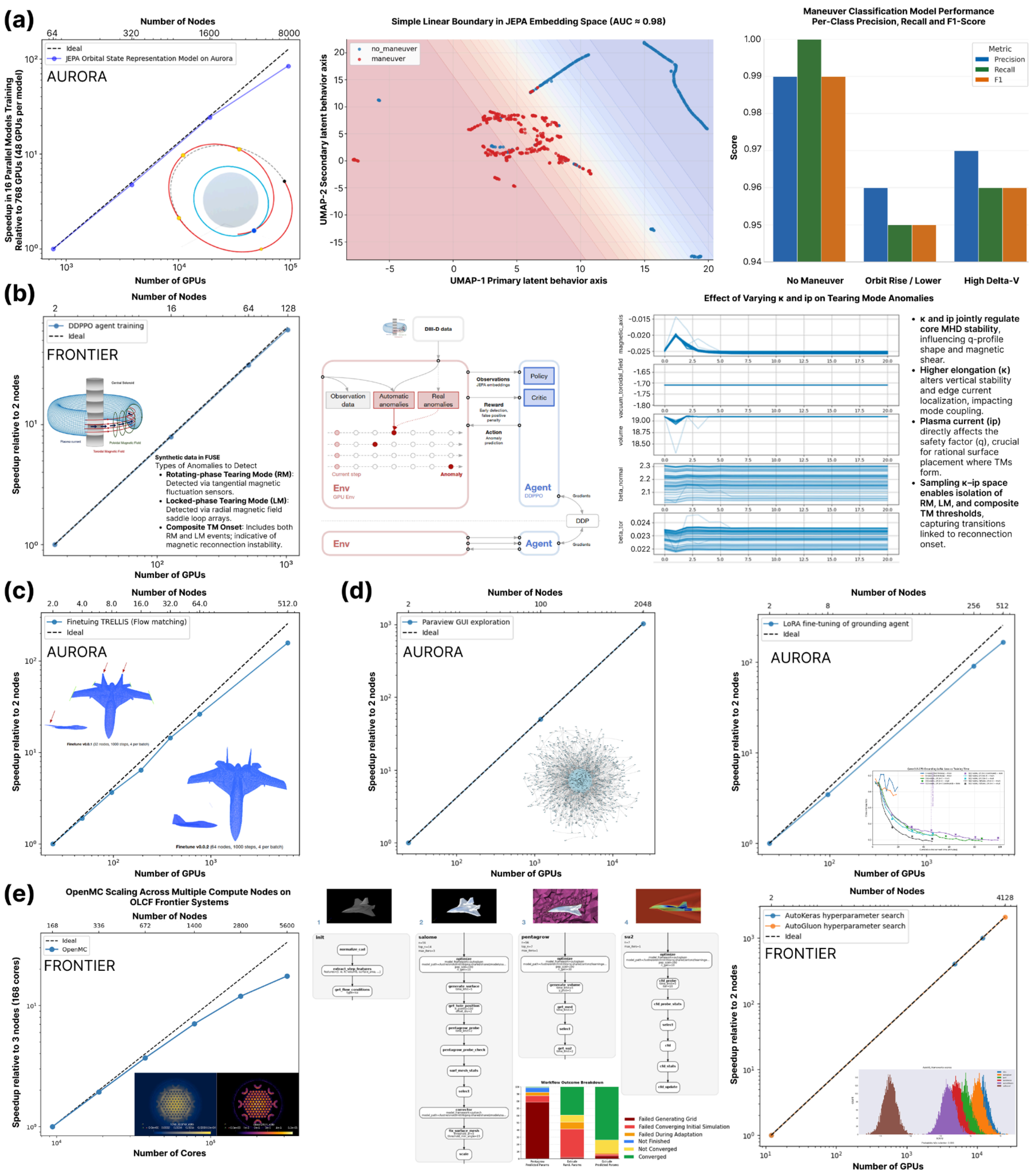

***Figure 50:*** *Scalability demonstrations across space, fusion, and engineering workflows. (a) Orbital JEPA world-model training on Aurora, run as 16 models in parallel to 8,000 nodes, shown with the maneuver embeddings and classification metrics of Figures 19 and 23c. (b) GPU-native reinforcement-learning environments with vectorized anomaly detection for fusion-facility operations, with near-linear distributed training on Frontier. (c) Geometry-synthesis model training on Aurora (512 nodes). (d) ParaView interface exploration (about 2,000 nodes) and LoRA fine-tuning for software control (512 nodes) on Aurora. (e) Scalable OpenMC, configuration-driven automated workflows, and large-scale AutoML optimization with mesh success-rate improvements on Frontier (1,000 to 4,000 nodes). Inset drawings are illustrative icons of each workload.*

The unifying engineering fact is that the same training factory scales across very different workloads. Training of the orbital JEPA world model, run as 16 models in parallel, scaled near-linearly from 64 to about 1,600 Aurora nodes and was executed at 8,000 nodes with reduced efficiency (Figure 50a). GPU-native reinforcement-learning training for a fusion-facility anomaly-detection environment, a separate application included here only for its scaling behavior, scaled near-linearly to 128 Frontier nodes (Figure 50b). Generative geometry training scaled to 512 nodes, with measured efficiencies that fall off gradually beyond that point and indicate that further gains depend more on data quality and curriculum than on adding nodes (Figure 50c). ParaView interface exploration was scaled to about 2,000 nodes, and LoRA adapter training for software control to 512 nodes (Figure 50d). The grid-quality and generation pipeline ran across as many as 1,000 nodes in parallel in the example configuration described earlier, producing 56 candidate surface meshes and 56 candidate volume-mesh configurations per case; a JEPA-based 3D geometry representation scaled to 256 nodes; and a 512-node multimodal relabeling pass converted weak endpoint labels into trajectory-level supervision. The interface-coverage gains reported in Section 4.4 come from a separate 1,024-node exploration run, which increased the diversity of collected interface states many times over at constant compute. The neutronics transport code showed near-linear strong scaling at low node counts and was driven at high throughput by running many independent jobs concurrently, which is the natural shape of a population-scale design campaign (Figure 50e).

Two engineering points support these numbers. First, we completed the portability work that heterogeneous systems require: components that originally depended on libraries specific to one accelerator vendor were adapted to run on both Intel and AMD systems, which is what allowed the same code to run across heterogeneous machines. Second, training was sized from first principles rather than by trial and error, using established analyses of large-batch training and of how to scale the learning rate and the slow target-encoder momentum with batch size, so that the reported scaling reflects genuine throughput rather than tuning artifacts (McCandlish et al., 2018; Busbridge et al., 2023). Where a deployed model must be small and fast, the efficient route is to train large on leadership hardware and then compress, rather than to train small (Li et al., 2020).

## 12. Verification, Safety, and Limitations

A system like GURU should be evaluated at two levels (Tabassi, 2023). The first level is model performance: classifier accuracy, representation separability, prediction error, policy reward, or mesh-quality prediction. The second level is workflow performance: whether the system executes the right tool, with the right inputs, under the right constraints, and produces an auditable result. The second level is often the one that matters most in deployment, because a high-scoring classifier does not guarantee a safe workflow if the wrong file, unit, or operating context was used. GURU therefore relies on replayable validation and regression testing of agent capabilities. A new agent must preserve prior capabilities or be scoped narrowly enough that its failures cannot propagate. In the interface-control setting, graph-based exploration and action logs let us compare state coverage and transition diversity. In computational fluid dynamics, deterministic grid-quality metrics provide an independent check on learned parameter selection. In space domain awareness, perturbation-based confidence scoring checks whether a detection remains stable under measurement-like uncertainty. In reactor digital engineering, validated solvers and physical constraints remain the confirmation path.

**Requirement traceability.** Table 5 traces each application against the five requirements stated in Section 1. Each entry names the mechanism and its maturity: measured (a reported result), prototype (implemented in simulation or replay), designed (specified, not yet running in the described form), or research target. Gaps remain visible on purpose.

**Table 5:** Requirement traceability across applications.

| Requirement | Mechanism in GURU | Strongest current evidence | Maturity and open gaps |
|---|---|---|---|
| R1. A validated reference decides | Separation of authority: learned components propose and rank; a domain-appropriate validated reference confirms and decides | CFD: SU2 decides convergence and grid-quality checks confirm meshes. Space: orbital propagation and physics-based screening. Airspace: NASA DAIDALUS alerting. Reactor: OpenMC validated on HTR-10, Baikal-1, and KRUSTY benchmarks | Measured for CFD and space; prototype for airspace (replay). Multi-aircraft: rule-based legality and model-based MPC, no validated vehicle model yet. Reactor validated-physics gate: research target |
| R2. Independent runtime check on every consequential action | Rules engine on typed, unit-aware Common Internal Language records; non-compensatory veto | Deterministic grid-quality library; in the published ablation cited in Section 3.5, removing the symbolic constraints lowered boundary-layer capture by 13 percentage points; about 30 perturbations per space alert; per-flight DAIDALUS evaluation | Measured for CFD and space. Multi-aircraft separation is a controller cost, not a verified constraint. Reactor actuation gate: designed |
| R3. Out-of-distribution inputs and uncertainty | Out-of-distribution gate, calibrated uncertainty, prediction-error channel, refusal path to a validated fallback | Prediction-error anomaly detection and perturbation confidence in space; confidence and evidence trails in airspace; confident CFD mispredictions caught by the solver (Figure 12h) | Prototype in several applications; the universal gate and refusal path are designed, not yet implemented on every agent |
| R4. Human retains override authority | Human sets intent and constraints, approves consequential actions, and can override | Operator approval for orbital avoidance maneuvers; advisory outputs reviewed before action, with authority retained by the FAA and air traffic control; mission-level supervision of aircraft teams | Prototype; reactor human-on-the-loop escalation is designed |
| R5. Documented lifecycle | Versioned, regression-tested, registered agents; provenance on every record; evidence references in the knowledge graph | Training factory protocols; frozen software-control protocols; fixed-seed checkpoint gates | Platform practice in place; no assurance-case artifact yet mapped to SAE ARP6983 or the NIST AI RMF (future work) |

**Evidence by application.** Table 6 answers the three questions from Section 1 for each application. Read together, the applications form an evidence ladder: strong end-to-end evidence (computational fluid dynamics and space domain awareness), transfer evidence (software control, airspace and reentry, and multi-aircraft control), and a prospective safety-critical instantiation (microreactor control) whose integrated actuation gate remains a research target.

**Table 6:** Evidence ladder: instantiation, utility, and assurance by application.

| Application | Instantiation | Utility | Assurance (R1 to R5) | Evidence tier |
|---|---|---|---|---|
| CFD | Navigation, Action, and Planning agents, rules engine, and validated solver all present | Converged rate on unseen geometries from under 1 to 74 percent | Solver and grid-quality checks decide; largely measured | 1: end-to-end |
| Space domain awareness | All roles present | Maneuver separability AUC near 0.98 (prior work); highest mean conjunction F1, 0.628 ± 0.008 | Propagation and physics-based screening decide; perturbation stress tests; Stages 3 and 4 in refinement | 1: end-to-end |
| Software control (ParaView) | Planning, Grounding, and Execution agents over the knowledge graph | Full-context success from 54.7 to 74.7 percent of attempts | Application execution with outcome scoring; development benchmark | 2: transfer |
| Airspace and reentry | Radar detection, physics-based debris propagation, and orchestration over NASA DAIDALUS, with anomaly classification | Uncertainty regions and avoidance demonstrated on a historical event and synthetic encounters | DAIDALUS decides alert levels; FAA and air traffic control retain authority; replay stage | 2: transfer |
| Multi-aircraft control | Planning, Action, and Navigation layers with typed legality rules | Blue arrival from 0.148 to 0.607 on unseen worlds | Option legality checked; separation not yet a verified constraint | 2: transfer |
| Microreactor control | Three-tier partition defined and traced to R1 to R5 | OpenMC benchmarks within band; reduced-order MPC and anomaly-detection prototypes | Integrated validated-physics gate is a research target | 3: prospective |

**Limitations.** Several limits belong in the main text. First, the applications are not equally mature. Computational fluid dynamics and space domain awareness have the strongest demonstrated evidence. The airspace and reentry work is a transfer prototype built on a replayed historical event, radar analysis of that event, and constructed synthetic encounters; it does not yet constitute a validated end-to-end live reconstruction of events. Reactor autonomy is a research target supported by design-space studies, validation benchmarks, heat-pipe-microreactor anomaly scoring, and a physics-anchored methodology, and it is not a claim of physical autonomous reactor operation. Second, synthetic data is useful, but it must be compared continuously with operational or experimental data when such data is available, which is the purpose of the validation and blind-spot analyses reported above. Third, confidence scoring is an operator triage aid and a way to expose uncertainty (R3). It is not a substitute for the validated reference (R1). Fourth, the software-control evaluation used structured application state over a small development benchmark, not remote-display-only control, and the multi-aircraft results come from simulation with synthetic communication stressors and no verified separation constraint. Fifth, the conjunction-risk F1 scores are modest in absolute terms, and differences among the strongest methods lie within seed-to-seed variation. Sixth, the CFD rates in Section 10.1 are single-attempt rates over 100 cases, and the number of distinct geometries among those cases was not recorded; the symbolic-layer ablation in Section 3.5 comes from our prior published work on an earlier grid generator and has not been repeated on the current workflow. Protocols and data are available from the corresponding author. Stating these limits plainly is part of the verification argument, because the value of the architecture is precisely that it keeps learned prediction separate from validated confirmation.

## 13. Discussion

The applications look very different on the surface, yet they share a structure. Computational fluid dynamics and its software post-processing, space domain awareness, airspace and reentry support, multi-aircraft control, and reactor digital engineering all involve stateful external environments, specialized software, hidden failure modes, and a need for human-readable evidence, and all of them benefit from learned representations and scalable training. The reusable part of GURU is the architecture that lets these pieces coexist. The same point explains the boundaries of transfer. A geometry representation model will not detect a satellite maneuver, and a maneuver detector will not validate a reactor control action. What transfers is the training factory, the agent-composition pattern, the Common Internal Language, the verification discipline, and the operational habit of keeping learned prediction separate from validated confirmation. For that reason we present this as a systems paper rather than a single-model paper. The platform is software-agnostic in exactly this sense: its interface, orchestration, Learning Engine, and Common Internal Language carry over, while trained agents are specific to their software and domain, by design.

Table 7 makes this concrete. It lists, for each application, the learned components, the planning or selection mechanism, the validated reference that decides, and the headline result with its maturity.

**Table 7:** What transfers across applications. The learned components vary by domain, while the training factory, the Common Internal Language, the agent-society composition, and the requirement that a domain-appropriate validated reference decides are shared.

| Application | Learned components | Planning and selection | Validated reference that decides (R1) | Headline result and maturity |
|---|---|---|---|---|
| CFD (digital engineering) | Geometry synthesis and setup-policy predictors trained by the shared Learning Engine | Multi-objective genetic optimization over candidate setups (Deb et al., 2002) | The CFD solver with grid-quality and convergence checks | Converged rate on unseen geometries from under 1 percent to 74 percent, 92 percent completing (measured) |
| Software control (ParaView) | Role-specific Planning and Grounding adapters with knowledge-graph retrieval | Planning agent decomposes goals; Grounding agent selects executable operations | Application execution with per-goal outcome scoring | Full-context success from 54.7 to 74.7 percent of attempts (measured, development benchmark) |
| Space domain awareness | Frozen self-supervised JEPA encoder and predictor; conjunction risk head | Staged screening; fusion and ranking of evidence (designed) | Orbital propagation (SGP4 and high-fidelity) with physics-based screening | Maneuver separability AUC near 0.98 (prior work); highest mean conjunction F1, 0.628 ± 0.008 over 10 seeds, matching fine-tuned language models within seed variability (measured) |
| Airspace and reentry | Anomaly classifier with confidence and evidence trail | Event-level orchestration and traffic-wide aggregation | NASA DAIDALUS detect-and-avoid alerting | In a historical replay with recorded traffic, 21 of 42 selected flights intersected the modeled hazard bands, with 8 accepted route alternatives and 13 flagged for review; decision-support consequences, no performance metric claimed (prototype) |
| Multi-aircraft control | Parameter-shared PPO policy, course-conditioned | Hierarchical MPC and local MPPI | Typed legality rules and model-based MPC; separation not yet a verified constraint | Blue arrival from 0.148 to 0.607 on 256 unseen worlds (prototype, simulation) |
| Microreactor control | Learned screening models for design-space search; reactor anomaly detector | Model-predictive control over candidate control-drum motions, scored by a physics-based reduced-order model | Validated-physics ensemble at the actuation gate with hard reactivity and temperature limits | OpenMC within the ±500 pcm band on HTR-10 (measured); reduced-order MPC tracks power targets under limits (prototype); validated-physics gate (research target) |

**What the table claims, and what it does not.** The invariants that transfer across every row are the training factory, the Common Internal Language, the agent-society composition, and the requirement that a domain-appropriate validated reference decides at the actuation gate. The reference itself differs by domain: a CFD solver with grid-quality checks, application execution with outcome scoring, orbital propagation with physics-based screening, NASA DAIDALUS, rule-based legality with model-based control, and validated reactor physics. The learned components also vary by domain and are stated per row. The claim the table supports is that the same factory and the same verification discipline carry across domains, not that a single encoder or a single verifier does.

This reuse extends past engineering simulation and physical operations into document-heavy reasoning: the software-control work in Sections 4.4 and 5.5 turns manuals, tutorials, and exploration into a provenance-preserving knowledge graph expressed in the same Common Internal Language. A single platform that spans engineering design, software operation, physical operations, and decision support is the clearest evidence that capability is added by training and registering agents over a shared representation, not by rebuilding the system for each use case.

**Why a better harness is not enough.** In engineering and operations the hard part is not producing a plausible action but knowing that an action is physically valid before it executes. Three properties of these domains make a model-centered design insufficient, however good its harness. There is often no test oracle at run time, so correctness must come from a validated reference rather than from agreement among models (U.S. Nuclear Regulatory Commission, 2024). A learned component and a learned checker trained on related data can fail together, the systematic common-cause pattern that reliability practice defends against with diversity (Liu and Ma, 2026). And the costs are asymmetric: an action that is right on average but occasionally violates a hard limit is not admissible. The authorities cited in Section 1 do not ask whether a model is accurate on average. They ask whether a safety-significant action remains inside a validated envelope when conditions are off-normal, and whether they can see why it was allowed. This is also the answer to the practical question of why a purpose-built system is needed rather than a frontier language model at the center. A general model with a harness has no explicit, inspectable state, no separation between what it knows and how it reasons, and no account of its decisions beyond the text it generates (Section 2.3); GURU has all three by construction, in the Common Internal Language, the knowledge graph, and the rules engine with its validated reference. The evidence in this paper bears on the comparison directly. Where we compared like for like, a frozen self-supervised world model with a small head scored highest of the five methods compared at conjunction-risk classification, including fully fine-tuned language models, with several-times-lower seed variance (Section 10.2); in the published ablation, the symbolic layer accounted for a measurable 13 points of boundary-layer capture on its own (Section 3.5); and a setup predictor that reported 0.90 confidence on a case that then diverged (Figure 12h) was caught by the solver, not by any model. Language models are used in GURU where they help, translating intent into the Common Internal Language and, as role-specific adapters on an open-weight model, driving engineering software (Section 5.5), and never where they decide. The competitive position is therefore not a better model but a different center of authority, which is the property the regulators and the operator of the national airspace now require (Section 1).

What the breadth taught us. Working across five applications with one system changed the system, and the changes are visible in the evidence. The largest CFD gain came from replacing the validated tool before learning on top of it: the grid-generator change alone raised the converged rate from under 1 percent to 39 percent, and learned parameter selection raised it to 74 percent (Section 10.1), which is why the factory now treats a successful, error-free run as an explicit optimization target rather than a filter applied afterward (Section 5.4). A setup predictor that reported 0.90 confidence on a case that then diverged (Figure 12h) is why the protective pattern of Section 3.5, an out-of-distribution gate, a calibrated uncertainty estimate, and a refusal path to a validated fallback, is specified for every agent rather than chosen per application; Table 5 records where it is implemented and where it is still designed. In software control, execution without an error turned out not to be success, which is why exploration outcomes are relabeled by the state actually reached and why explicit procedures and API contracts retrieved from the knowledge graph, with role-specific adapters, produced the gain from 55 to 75 percent (Sections 4.4 and 5.5). In space domain awareness, a frozen self-supervised encoder with a small head proved more stable across seeds than fully fine-tuned language models, and SIGReg trained where EMA targets collapsed (Sections 10.2 and 10.3); that recipe, with the prediction-error channel as the anomaly signal, is the one the paper carries forward, and it transferred unchanged to reactor operating data (Section 10.5). In airspace, an all-time hazard outline discards encounter timing, and screening candidate maneuvers against fragment forecasts before the detect-and-avoid check avoided violations that reactive guidance did not (Sections 7.4 and 7.5); hazard representations therefore carry time. In multi-aircraft control, separation handled as a controller cost is not a verified constraint, and Section 12 says so rather than letting a typed legality rule stand in for a validated vehicle model. In reactor control, the deciding tier is physics in reduced, control-grade form and never a learned proxy, because the moment the learned layer is least reliable is the moment the validated physics must hold the most (Section 9). Each of these is the same lesson: where a learned component was wrong, the architecture caught it only because something validated and independent stood between the proposal and the action, and every revision strengthened that separation rather than the model.

The architecture also changes how leadership computing is justified. In a traditional campaign, the machine runs a large number of physics cases. In GURU, the machine also generates the data needed to train agents, explores software interfaces, creates synthetic operating scenarios, tests confidence under perturbation, and runs validation ensembles. The output is not only a dataset or a curve. It is a trained and tested capability that can operate inside a governed workflow. The next research step is to turn the current evidence into a standardized evaluation suite in which each domain reports task success, agent-level metrics, workflow-level metrics, scaling behavior, failure modes, and verification coverage.

# 14. Conclusions

GURU scales across domains because it treats mission-critical automation as a reusable agentic platform rather than a collection of one-off models. The same hierarchy of Navigation, Action, and Planning agents operates over very different environments once domain data, tools, constraints, ontologies, and validation rules are registered. Computational fluid dynamics and software post-processing, space domain awareness, airspace decision support, multi-aircraft control, and reactor engineering differ in their physics, their software, and their evidence requirements, but not in the training factory. Each domain contributes assets that become reusable agent capabilities, and each is held to the same standard: learned models provide adaptability, while symbolic constraints, provenance, verification, and validated physics provide the reliability that high-consequence work demands. That standard is a purpose-built answer to a long-standing problem. Symbolic methods are verifiable but brittle, learned methods are adaptable but do not reliably obey hard constraints, and the recognized response is to combine them (Garcez et al., 2019). What we contribute is a specific and safety-grade form of that combination: a hierarchy that does not merge learned and validated computation but assigns them different authority, so the learned tier proposes, ranks, and accelerates while the validated tier decides. The answer is specific. Five mechanisms recur across the applications: a separation of authority in which learned components propose and rank while a validated reference decides (R1); a symbolic verification layer that checks every consequential action against typed, unit-aware hard limits and holds a non-compensatory veto (R2); out-of-distribution gating and prediction-error monitoring that demote learned proposals as conditions leave the validated envelope (R3); a human who sets intent and retains override authority (R4); and a documented lifecycle of versioned, regression-tested, registered agents with provenance on every record (R5). Their maturity differs by application, and Section 12 reports it application by application.

The results support this claim concretely. In computational fluid dynamics, the updated workflow raised the converged-simulation rate on previously unseen aircraft geometries from under 1 percent to 74 percent, with 92 percent of cases completing, and, in one live demonstration, drove 54 simulations across the full speed range on three facilities in a single run. In software control, knowledge-graph retrieval and role-specific adapters raised ParaView task success from 54.7 to 74.7 percent of attempts under full context. In space domain awareness, where a language model cannot be trusted to decide and validated orbital propagation holds that seat, a frozen self-supervised world model scored highest of the five methods compared at conjunction-risk classification, including fully fine-tuned language models, with seed variance several times lower than theirs, and the same world-model recipe transferred to anomaly detection in reactor operating data: the trustworthy design gave up nothing in accuracy and gained stability. In simulated multi-aircraft control, the same hierarchy raised target arrival on unseen worlds from 0.148 to 0.607 while keeping tactical options under typed legality rules. Across these applications, training scaled near-linearly to more than a thousand Aurora nodes, with a world-model run executed at 8,000 nodes, and the same factory drove campaigns of independent jobs across thousands of Frontier nodes. Working across these five applications also changed the system itself, in ways Section 13 records: each revision made after a learned component was wrong strengthened the validated, independent check between proposal and action rather than the model. The same approach now extends toward a control-grade microreactor digital twin and supervised autonomous operation, where the defining design choice is the same partition of authority. That choice is what makes the system both useful at scale and trustworthy where failure is unacceptable.

This paper does not claim that every application has reached the same maturity. It claims something more fundamental: across materially different engineering and operations settings, the same architectural partition remains useful and intact. The strongest end-to-end evidence comes from computational fluid dynamics and space domain awareness; newer applications provide progressively earlier evidence of transfer, from measured software-control and multi-aircraft results to replayed airspace decision support and a reactor assurance architecture whose integrated actuation gate remains a research target. The applications therefore constitute a staged test of one hypothesis rather than claims of completed deployment. Across the evidence accumulated so far, the hypothesis has held: learned models are valuable for perception, prediction, adaptation, and search, but high-consequence autonomy requires a separate authority for validity, a symbolic and domain-validated layer capable of refusing the learned system. No evaluated application has required a learned component to assume the validating authority reserved for independent symbolic or physics-based mechanisms. The deciding reference differs by domain (a CFD solver with grid-quality and convergence checks, orbital propagation with physics-based screening, NASA DAIDALUS, typed legality rules with model-based control, and validated reactor physics), but it is never a learned approximation. What generalizes is not a particular model. What generalizes is the allocation of authority. That is also the answer to the question of language models and their harnesses: a better harness improves a model-centered system, whereas GURU changes the center of authority, and the measured comparisons in this paper, a frozen world model scoring highest against fully fine-tuned language models with several-times-lower seed variance, a symbolic layer worth 13 points of boundary-layer capture on its own, and a confident misprediction caught by the solver rather than by any model, are what that change buys.

This is the level of assurance that high-consequence industrial operation now demands, and it is what the emerging requirements describe. The Nuclear Regulatory Commission expects physics-based models in safety-significant roles and, in its proposed Part 57 rule, anticipates autonomous microreactor operation; national-security guidance sets the bar at reliable, robust, steerable, and controllable under rigorous test, evaluation, validation, and verification; the civil-aviation and space communities place every learned component behind an independent monitor with a human retaining override; and the demand side has arrived with the same condition attached, in a national policy that calls for more than 1,000 launches and reentries a year by 2030 with launch and reentry management integrated into air traffic control modernization, and in a Federal Aviation Administration that fielded its first artificial-intelligence airspace system in 2026 while stating that controllers remain responsible for separation. The five requirements of Section 1 are the common content of those documents, and Table 5 traces every application against them, gaps included. GURU was purpose-built to that bar rather than retrofitted to it. The partition it enforces, learned models accelerate and rank while validated physics confirms and decides, is what makes human-on-the-loop autonomy auditable and licensable. It is the contribution we intend the field to carry forward: for anything mission-critical, the path to trustworthy autonomy is not a better approximation in the deciding seat, but a validated authority that an approximation can inform and never replace.

## Acknowledgments

An award of computer time was provided by the ASCR Leadership Computing Challenge (ALCC) program. This research used resources of the Argonne Leadership Computing Facility, which is a U.S. Department of Energy Office of Science User Facility operated under contract DE-AC02-06CH11357. This research used resources of the Oak Ridge Leadership Computing Facility at the Oak Ridge National Laboratory, which is supported by the Advanced Scientific Computing Research programs in the Office of Science of the U.S. Department of Energy under Contract No. DE-AC05-00OR22725. We thank Tam Fortis Solutions for collaboration on the open reference microreactor research presented in Section 9.

## Data and Code Availability

Public data sources are cited in the text, including Space-Track catalog and conjunction data, published reactor-physics and shielding benchmarks, and open solvers. The GURU software is proprietary to MSBAI. Curated datasets and trained-model artifacts that underlie specific reported results are available from the corresponding author. Requests for access to data supporting specific results may be directed to the corresponding author.

## References

Adams, J. A., Hamell, J., & Walker, P. (2023). Can a single human supervise a swarm of 100 heterogeneous robots? arXiv:2308.00102. https://arxiv.org/abs/2308.00102

Alarcon-Rodriguez, J. R., Martinez-Fadrique, F. M., & Klinkrad, H. (2002). Collision risk assessment with a smart sieve method. In Proceedings of the Joint ESA-NASA Space-Flight Safety Conference (ESA SP-486, pp. 159-164). European Space Agency. https://www.researchgate.net/publication/234420237_Collision_Risk_Assessment_with_a_Smart_Sieve'_Method

Assran, M., Bardes, A., Fan, D., Garrido, Q., Howes, R., Komeili, M., Muckley, M., Rizvi, A., Roberts, C., Sinha, K., Zholus, A., Arnaud, S., Gejji, A., Martin, A., Robert Hogan, F., Dugas, D., Bojanowski, P., Khalidov, V., Labatut, P., ... Ballas, N. (2025). V-JEPA 2: Self-supervised video models enable understanding, prediction and planning. arXiv:2506.09985. https://arxiv.org/abs/2506.09985

Assran, M., Duval, Q., Misra, I., Bojanowski, P., Vincent, P., Rabbat, M., LeCun, Y., & Ballas, N. (2023). Self-supervised learning from images with a joint-embedding predictive architecture. In Proceedings of the IEEE/CVF Conference on Computer Vision and Pattern Recognition (pp. 15619-15629). IEEE. https://openaccess.thecvf.com/content/CVPR2023/html/Assran_Self-Supervised_Learning_From_Images_With_a_Joint-Embedding_Predictive_Architecture_CVPR_2023_paper.html

Baars, B. J. (1988). A cognitive theory of consciousness. Cambridge University Press.

Balaprakash, P., Salim, M., Uram, T. D., Vishwanath, V., & Wild, S. M. (2018). DeepHyper: Asynchronous hyperparameter search for deep neural networks. In 2018 IEEE 25th International Conference on High Performance Computing (HiPC) (pp. 42-51). IEEE. https://www.researchgate.net/publication/331039792_DeepHyper_Asynchronous_Hyperparameter_Search_for_Deep_Neural_Networks

Balestriero, R., & LeCun, Y. (2025). LeJEPA: Provable and scalable self-supervised learning without the heuristics. arXiv:2511.08544. https://arxiv.org/abs/2511.08544

Bardes, A., Garrido, Q., Ponce, J., Chen, X., Rabbat, M., LeCun, Y., Assran, M., & Ballas, N. (2024). Revisiting feature prediction for learning visual representations from video. Transactions on Machine Learning Research. https://arxiv.org/abs/2404.08471

Breiman, L. (2001). Random forests. Machine Learning, 45(1), 5-32. https://doi.org/10.1023/A:1010933404324

Brown, D. A., Chadwick, M. B., Capote, R., Kahler, A. C., Trkov, A., Herman, M. W., Sonzogni, A. A., Danon, Y., Carlson, A. D., Dunn, M., Smith, D. L., Hale, G. M., Arbanas, G., Arcilla, R., Bates, C. R., Beck, B., Becker, B., Brown, F., Casperson, R. J., ... Zhu, Y. (2018). ENDF/B-VIII.0: The 8th major release of the nuclear reaction data library with CIELO-project cross sections, new standards and thermal scattering data. Nuclear Data Sheets, 148, 1-142. https://www.sciencedirect.com/science/article/pii/S0090375218300206

Busbridge, D., Ramapuram, J., Ablin, P., Likhomanenko, T., Dhekane, E. G., Suau, X., & Webb, R. (2023). How to scale your EMA. In Advances in Neural Information Processing Systems 36. arXiv:2307.13813. https://papers.nips.cc/paper_files/paper/2023/file/e7681dd6fe16052433ab68cd1555bdc9-Paper-Conference.pdf

Chen, L., Lu, K., Rajeswaran, A., Lee, K., Grover, A., Laskin, M., Abbeel, P., Srinivas, A., & Mordatch, I. (2021). Decision Transformer: Reinforcement learning via sequence modeling. In Advances in Neural Information Processing Systems 34. https://arxiv.org/abs/2106.01345

Chen, T., & Guestrin, C. (2016). XGBoost: A scalable tree boosting system. In Proceedings of the 22nd ACM SIGKDD International Conference on Knowledge Discovery and Data Mining (pp. 785-794). ACM. https://dl.acm.org/doi/10.1145/2939672.2939785

de Moura, L., & Bjørner, N. (2008). Z3: An efficient SMT solver. In Tools and Algorithms for the Construction and Analysis of Systems (Lecture Notes in Computer Science Vol. 4963, pp. 337-340). Springer. https://link.springer.com/chapter/10.1007/978-3-540-78800-3_24

Deb, K., Pratap, A., Agarwal, S., & Meyarivan, T. (2002). A fast and elitist multiobjective genetic algorithm: NSGA-II. IEEE Transactions on Evolutionary Computation, 6(2), 182-197. https://ieeexplore.ieee.org/document/996017

Defense Advanced Research Projects Agency. (2019, March 22). In the sky and on the ground, collaboration vital to DARPA's CODE for success. https://www.darpa.mil/news/2019/code-success

Defense Advanced Research Projects Agency. (n.d.). OFFensive Swarm-Enabled Tactics (OFFSET). https://www.darpa.mil/research/programs/offensive-swarm-enabled-tactics

Department of Defense. (2023). DoD Instruction 5000.97: Digital engineering. Office of the Under Secretary of Defense for Research and Engineering.

Dowell, D. C., Alexander, C. R., James, E. P., Weygandt, S. S., Benjamin, S. G., Manikin, G. S., Blake, B. T., Brown, J. M., Olson, J. B., Hu, M., Smirnova, T. G., Ladwig, T., Kenyon, J. S., Ahmadov, R., Turner, D. D., Duda, J. D., & Alcott, T. I. (2022). The High-Resolution Rapid Refresh (HRRR): An hourly updating convection-allowing forecast model. Part I: Motivation and system description. Weather and Forecasting, 37(8), 1371-1395. https://doi.org/10.1175/WAF-D-21-0151.1

Dunlap, K., Bennett, K., van Wijk, D., Hamilton, N., & Hobbs, K. (2024). Run time assured reinforcement learning for six degree-of-freedom spacecraft inspection. In AIAA AVIATION Forum and ASCEND 2024. American Institute of Aeronautics and Astronautics. https://doi.org/10.2514/6.2024-4935 (arXiv:2406.11795)

Economon, T. D., Palacios, F., Copeland, S. R., Lukaczyk, T. W., & Alonso, J. J. (2016). SU2: An open-source suite for multiphysics simulation and design. AIAA Journal, 54(3), 828-846. https://arc.aiaa.org/doi/10.2514/1.J053813

Ennadir, S., Golkar, S., & Sarra, L. (2025). Joint embeddings go temporal. arXiv:2509.25449. https://arxiv.org/abs/2509.25449

Erickson, N., Mueller, J., Shirkov, A., Zhang, H., Larroy, P., Li, M., & Smola, A. (2020). AutoGluon-Tabular: Robust and accurate AutoML for structured data. arXiv:2003.06505. https://arxiv.org/abs/2003.06505

European Union Aviation Safety Agency. (2024). EASA artificial intelligence concept paper, Issue 2: Guidance for Level 1 and Level 2 machine-learning applications. https://www.easa.europa.eu/en/document-library/general-publications/easa-artificial-intelligence-concept-paper-issue-2

European Union Aviation Safety Agency. (2026). EASA artificial intelligence concept paper, proposed Issue 03: Extension to reinforcement learning and symbolic AI. https://www.easa.europa.eu/en/newsroom-and-events/news/easa-releases-latest-issue-its-concept-paper-artificial-intelligence

Federal Aviation Administration. (2024). Roadmap for artificial intelligence safety assurance, Version 1. U.S. Department of Transportation. https://www.faa.gov/media/82891

National Security Presidential Memorandum 17. (2026, August 20). The national space transportation policy. The White House. https://www.whitehouse.gov/presidential-actions/2026/08/national-security-presidential-memorandum-nspm-17/

Federal Aviation Administration. (2026a, January 8). Airspace management considerations for space launch activities (Safety Alert for Operators No. 26001). U.S. Department of Transportation, Flight Standards Service. https://www.faa.gov/sites/faa.gov/files/SAFO26001.pdf

Federal Aviation Administration. (2026b). Strategic Management of Airspace, Routes, and Trajectories (SMART) [Fact sheet]. U.S. Department of Transportation. https://www.faa.gov/newsroom/SMART_One-Pager.pdf

Feurer, M., Klein, A., Eggensperger, K., Springenberg, J., Blum, M., & Hutter, F. (2015). Efficient and robust automated machine learning. In Advances in Neural Information Processing Systems 28 (pp. 2962-2970). Curran Associates. https://papers.nips.cc/paper_files/paper/2015/hash/11d0e6287202fced83f79975ec59a3a6-Abstract.html

Fischer, P., Kerkemeier, S., Min, M., Lan, Y.-H., Phillips, M., Rathnayake, T., Merzari, E., Tomboulides, A., Karakus, A., Chalmers, N., & Warburton, T. (2022). NekRS, a GPU-accelerated spectral element Navier-Stokes solver. Parallel Computing, 114, 102982. https://doi.org/10.1016/j.parco.2022.102982

Fries, M., & Fries, J. (2010). Doppler weather radar as a meteorite recovery tool. Meteoritics & Planetary Science, 45(9), 1476-1487. https://doi.org/10.1111/j.1945-5100.2010.01115.x

Fries, M., & Fries, J. (2025). Weather radar detection and analysis of bolides in the day or night. Advances in Astronomy, 2025(1), Article 3041760. https://doi.org/10.1155/aa/3041760

Garcez, A. S. d'Avila, Gori, M., Lamb, L. C., Serafini, L., Spranger, M., & Tran, S. N. (2019). Neural-symbolic computing: An effective methodology for principled integration of machine learning and reasoning. Journal of Applied Logics, 6(4), 611-632. https://openaccess.city.ac.uk/id/eprint/22984/

Garnelo, M., & Shanahan, M. (2019). Reconciling deep learning with symbolic artificial intelligence: Representing objects and relations. Current Opinion in Behavioral Sciences, 29, 17-23. https://doi.org/10.1016/j.cobeha.2018.12.010

Geirhos, R., Jacobsen, J.-H., Michaelis, C., Zemel, R., Brendel, W., Bethge, M., & Wichmann, F. A. (2020). Shortcut learning in deep neural networks. Nature Machine Intelligence, 2(11), 665-673. https://doi.org/10.1038/s42256-020-00257-z

Gibson, M. A., Poston, D. I., McClure, P. R., Godfroy, T. J., Briggs, M. H., & Sanzi, J. L. (2018). The Kilopower Reactor Using Stirling Technology (KRUSTY) nuclear ground test results and lessons learned (NASA/TM-2018-219941). National Aeronautics and Space Administration. https://ntrs.nasa.gov/citations/20180007389

Graepel, T. (2026, October 2). Don't be fooled: LLMs don't reason. MIT Technology Review. https://www.technologyreview.com/2026/10/02/1145639/dont-be-fooled-llms-dont-reason/

Grosvenor, A. (2026, May 6). Virtual range architecture: Hybrid AI for space simulation [Conference presentation]. 2026 Department of the Air Force Modeling, Simulation and Analytics Summit, Colorado Springs, CO, United States. https://www.dafmss.org/agenda

Grosvenor, A., Wahab, A., Bohachov, K., Zemlyansky, A., Adams, R., & Deighan, D. (2025). Hierarchical neuro-symbolic AI for autonomous spacecraft maneuvering and anomaly detection [Conference paper]. Advanced Maui Optical and Space Surveillance Technologies (AMOS) Conference. https://amostech.com/TechnicalPapers/2025/Poster/Grosvenor.pdf

Grosvenor, A., Zemlyansky, A., Deighan, D., & Sysko, D. (2022). Simulation workflows in minutes, at scale for next-generation HPC. In Smoky Mountains Computational Sciences and Engineering Conference (SMC 2022) (Communications in Computer and Information Science Vol. 1690, pp. 288-300). Springer. https://www.researchgate.net/publication/367232982_Simulation_Workflows_in_Minutes_at_Scale_for_Next-Generation_HPC

Grosvenor, A., Zemlyansky, A., Wahab, A., Bohachov, K., Dogan, A., & Deighan, D. (2025). Hybrid intelligence systems for reliable automation: Advancing knowledge work and autonomous operations with scalable AI architectures. Frontiers in Robotics and AI, 12, Article 1566623. https://doi.org/10.3389/frobt.2025.1566623

Healy, L. M. (1995). Close conjunction detection on parallel computer. Journal of Guidance, Control, and Dynamics, 18(4), 824-829. https://doi.org/10.2514/3.21465

Heinrichs, D., Lee, C., & Lent, E. (2022). Baikal-1 skyshine experiment (LLNL-TR-843853). Lawrence Livermore National Laboratory. https://cog.llnl.gov/sites/cog/files/2023-01/baikal-1.pdf

Hobbs, K. L., Phillips, S., Simon, M., Lyons, J. B., Culbertson, J., Clouse, H. S., Hamilton, N., Dunlap, K., Lippay, Z. S., Aurand, J., Bell, Z. I., Hammack, T., Ayres, D., & Lim, R. (2025). The Safe Trusted Autonomy for Responsible Space program. arXiv:2501.05984. https://arxiv.org/abs/2501.05984

Hu, E. J., Shen, Y., Wallis, P., Allen-Zhu, Z., Li, Y., Wang, S., Wang, L., & Chen, W. (2022). LoRA: Low-rank adaptation of large language models. In International Conference on Learning Representations. https://arxiv.org/abs/2106.09685

Jin, H., Song, Q., & Hu, X. (2019). Auto-Keras: An efficient neural architecture search system. In Proceedings of the 25th ACM SIGKDD International Conference on Knowledge Discovery and Data Mining (pp. 1946-1956). ACM. https://doi.org/10.1145/3292500.3330648

Kambhampati, S., Valmeekam, K., Guan, L., Verma, M., Stechly, K., Bhambri, S., Saldyt, L., & Murthy, A. (2024). LLMs can't plan, but can help planning in LLM-Modulo frameworks. In Proceedings of the 41st International Conference on Machine Learning. https://proceedings.mlr.press/v235/kambhampati24a.html

Ke, G., Meng, Q., Finley, T., Wang, T., Chen, W., Ma, W., Ye, Q., & Liu, T.-Y. (2017). LightGBM: A highly efficient gradient boosting decision tree. In Advances in Neural Information Processing Systems 30 (pp. 3146-3154). Curran Associates.

Kelso, T. S., & Alfano, S. (2005). Satellite orbital conjunction reports assessing threatening encounters in space (SOCRATES) (Paper AAS 05-124). American Astronautical Society. https://celestrak.org/publications/AAS/05-124/AAS-05-124.pdf

Klindt, D., LeCun, Y., & Balestriero, R. (2026). When does LeJEPA learn a world model? arXiv:2605.26379. https://arxiv.org/abs/2605.26379

LeCun, Y. (2022). A path towards autonomous machine intelligence (Version 0.9.2). OpenReview. https://openreview.net/pdf?id=BZ5a1r-kVsf

LeDell, E., & Poirier, S. (2020). H2O AutoML: Scalable automatic machine learning. In 7th ICML Workshop on Automated Machine Learning (AutoML).

Li, L., Xue, H., Song, Y., & Salim, F. D. (2024). T-JEPA: A joint-embedding predictive architecture for trajectory similarity computation. In Proceedings of the 32nd ACM SIGSPATIAL International Conference on Advances in Geographic Information Systems (pp. 569-572). ACM. https://dl.acm.org/doi/10.1145/3678717.3691271

Li, Z., Wallace, E., Shen, S., Lin, K., Keutzer, K., Klein, D., & Gonzalez, J. E. (2020). Train large, then compress: Rethinking model size for efficient training and inference of transformers. In Proceedings of the 37th International Conference on Machine Learning (Vol. 119, pp. 5958-5968). PMLR. https://proceedings.mlr.press/v119/li20m.html

Liu, T., & Ma, Z. (2026). Operating experience data analysis for digital instrumentation and control system reliability and risk assessment (NUREG/CR-7313, INL/RPT-24-82553). U.S. Nuclear Regulatory Commission, Office of Nuclear Regulatory Research. https://www.nrc.gov/docs/ML2617/ML26175A375.pdf

Lowe, R., Wu, Y., Tamar, A., Harb, J., Abbeel, P., & Mordatch, I. (2017). Multi-agent actor-critic for mixed cooperative-competitive environments. In Advances in Neural Information Processing Systems 30. https://arxiv.org/abs/1706.02275

Lundberg, S. M., & Lee, S.-I. (2017). A unified approach to interpreting model predictions. In Advances in Neural Information Processing Systems 30 (pp. 4765-4774). Curran Associates.

Maes, L., Le Lidec, Q., Scieur, D., LeCun, Y., & Balestriero, R. (2026). LeWorldModel: Stable end-to-end joint-embedding predictive architecture from pixels. arXiv:2603.19312. https://arxiv.org/abs/2603.19312

Maisonobe, L., Pommier, V., & Parraud, P. (2010, May 3-6). Orekit: An open source library for operational flight dynamics applications [Conference paper]. 4th International Conference on Astrodynamics Tools and Techniques (ICATT), Madrid, Spain. https://www.orekit.org

McCandlish, S., Kaplan, J., Amodei, D., & OpenAI Dota Team. (2018). An empirical model of large-batch training. arXiv:1812.06162. https://arxiv.org/abs/1812.06162

McDonald, R. A., & Gloudemans, J. R. (2022). Open Vehicle Sketch Pad: An open-source parametric geometry and analysis tool for conceptual aircraft design. In AIAA SciTech 2022 Forum. https://arc.aiaa.org/doi/10.2514/6.2022-0004

McInnes, L., Healy, J., & Melville, J. (2018). UMAP: Uniform manifold approximation and projection for dimension reduction. arXiv:1802.03426. https://arxiv.org/abs/1802.03426

McKay, M. D., Beckman, R. J., & Conover, W. J. (1979). A comparison of three methods for selecting values of input variables in the analysis of output from a computer code. Technometrics, 21(2), 239-245. https://www.jstor.org/stable/1268522

Merzari, E., Hamilton, S., Evans, T., Min, M., Fischer, P., Kerkemeier, S., Fang, J., Romano, P., Lan, Y.-H., Phillips, M., Rathnayake, T., Biondo, E., Royston, K., Chalmers, N., & Warburton, T. (2023). Exascale multiphysics nuclear reactor simulations for advanced designs. In SC '23: Proceedings of the International Conference for High Performance Computing, Networking, Storage and Analysis (Article 3). Association for Computing Machinery. https://doi.org/10.1145/3581784.3627038

Merzari, E., Yuan, H., Min, M., Shaver, D., Rahaman, R., Shriwise, P., Romano, P., Talamo, A., Lan, Y.-H., Gaston, D., Martineau, R., Fischer, P., & Hassan, Y. (2021). Cardinal: A lower-length-scale multiphysics simulator for pebble-bed reactors. Nuclear Technology, 207(7), 1118-1141. https://www.ans.org/pubs/journals/nt/article-49885/

Miao, Y., Cao, Y., Mo, K., Stauff, N. E., Lee, C., Matthews, C., & Shemon, E. R. (2025). Multiphysics simulation of KRUSTY warm critical experiments using MOOSE tools. Nuclear Science and Engineering. Advance online publication. https://www.osti.gov/servlets/purl/3007284

Minsky, M. (1986). The society of mind. Simon & Schuster.

Minsky, M. L. (1991). Logical versus analogical or symbolic versus connectionist or neat versus scruffy. AI Magazine, 12(2), 34-51.

Muñoz, C., Narkawicz, A., Hagen, G., Upchurch, J., Dutle, A., Consiglio, M., & Chamberlain, J. (2015). DAIDALUS: Detect and avoid alerting logic for unmanned systems. In 2015 IEEE/AIAA 34th Digital Avionics Systems Conference (DASC) (pp. 5A1-1 to 5A1-12). IEEE. https://doi.org/10.1109/DASC.2015.7311421

National Aeronautics and Space Administration. (2023). NASA spacecraft conjunction assessment and collision avoidance best practices handbook (NASA/SP-20230002470, Rev. 1). https://nodis3.gsfc.nasa.gov/OCE_docs/OCE_51.pdf

National Security Presidential Memorandum 11. (2026, June 5). Artificial intelligence in the national security enterprise. The White House. https://www.whitehouse.gov/presidential-actions/2026/06/national-security-presidential-memorandum-nspm-11/

Nie, Y., Nguyen, N. H., Sinthong, P., & Kalagnanam, J. (2023). A time series is worth 64 words: Long-term forecasting with transformers. In International Conference on Learning Representations. https://openreview.net/forum?id=Jbdc0vTOcol

Olson, R. S., Bartley, N., Urbanowicz, R. J., & Moore, J. H. (2016). Evaluation of a tree-based pipeline optimization tool for automating data science. In Proceedings of the Genetic and Evolutionary Computation Conference 2016 (pp. 485-492). ACM. https://arxiv.org/abs/1603.06212

Operation of Multiple Small Unmanned Aircraft, 14 C.F.R. § 107.35 (2026). https://www.ecfr.gov/current/title-14/chapter-I/subchapter-F/part-107/subpart-B/section-107.35

Oquab, M., Darcet, T., Moutakanni, T., Vo, H. V., Szafraniec, M., Khalidov, V., Fernandez, P., Haziza, D., Massa, F., El-Nouby, A., Assran, M., Ballas, N., Galuba, W., Howes, R., Huang, P.-Y., Li, S.-W., Misra, I., Rabbat, M., Sharma, V., … Bojanowski, P. (2024). DINOv2: Learning robust visual features without supervision. Transactions on Machine Learning Research. https://openreview.net/forum?id=a68SUt6zFt

Ovadia, Y., Fertig, E., Ren, J., Nado, Z., Sculley, D., Nowozin, S., Dillon, J. V., Lakshminarayanan, B., & Snoek, J. (2019). Can you trust your model's uncertainty? Evaluating predictive uncertainty under dataset shift. In Advances in Neural Information Processing Systems 32. https://arxiv.org/abs/1906.02530

Peddakotla, S. A., Morgado, F., Thillaithevan, D., O'Driscoll, D., Santer, M., Maddock, C., Vasile, M., & Fossati, M. (2022, September 18-22). Multi-fidelity and multi-disciplinary approach for the accurate simulation of atmospheric re-entry (Paper IAC-22,A6,IPB,2,x73592) [Conference paper]. 73rd International Astronautical Congress (IAC 2022), Paris, France. https://strathprints.strath.ac.uk/82459/

Permann, C. J., Gaston, D. R., Andrs, D., Carlsen, R. W., Kong, F., Lindsay, A. D., Miller, J. M., Peterson, J. W., Slaughter, A. E., Stogner, R. H., & Martineau, R. C. (2020). MOOSE: Enabling massively parallel multiphysics simulation. SoftwareX, 11, 100430. https://www.sciencedirect.com/science/article/pii/S2352711019302973

Poston, D. I., Gibson, M. A., Godfroy, T., & McClure, P. R. (2020). KRUSTY reactor design. Nuclear Technology, 206(sup1), S13-S30. https://www.osti.gov/biblio/1699487

Prokhorenkova, L., Gusev, G., Vorobev, A., Dorogush, A. V., & Gulin, A. (2018). CatBoost: Unbiased boosting with categorical features. In Advances in Neural Information Processing Systems 31. https://arxiv.org/abs/1706.09516

Qwen Team. (2026). Qwen3.8-27B [Large language model]. Hugging Face. https://huggingface.co/Qwen/Qwen3.8-27B

Radford, A., Kim, J. W., Hallacy, C., Ramesh, A., Goh, G., Agarwal, S., Sastry, G., Askell, A., Mishkin, P., Clark, J., Krueger, G., & Sutskever, I. (2021). Learning transferable visual models from natural language supervision. In Proceedings of the 38th International Conference on Machine Learning (pp. 8748-8763). https://proceedings.mlr.press/v139/radford21a.html

Romano, P. K., Horelik, N. E., Herman, B. R., Nelson, A. G., Forget, B., & Smith, K. (2015). OpenMC: A state-of-the-art Monte Carlo code for research and development. Annals of Nuclear Energy, 82, 90-97. https://www.sciencedirect.com/science/article/abs/pii/S0306454914003752

SAE International, & EUROCAE. (n.d.). ARP6983 / ED-324: Process standard for development and certification or approval of aeronautical safety-related products implementing AI [Standard in development; SAE G-34 and EUROCAE WG-114, W-shaped lifecycle]. https://standict.eu/standards-repository/eurocae-wg114-sae-g34-artificial-intelligence-aviation-as6983-process-standard

Schick, T., Dwivedi-Yu, J., Dessì, R., Raileanu, R., Lomeli, M., Hambro, E., Zettlemoyer, L., Cancedda, N., & Scialom, T. (2023). Toolformer: Language models can teach themselves to use tools. In Advances in Neural Information Processing Systems 36. https://arxiv.org/abs/2302.04761

Schierman, J. D., DeVore, M. D., Richards, N. D., & Clark, M. A. (2020). Runtime assurance for autonomous aerospace systems. Journal of Guidance, Control, and Dynamics, 43(12), 2205-2217. https://doi.org/10.2514/1.G004862

Schulman, J., Wolski, F., Dhariwal, P., Radford, A., & Klimov, O. (2017). Proximal policy optimization algorithms. arXiv:1707.06347. https://arxiv.org/abs/1707.06347

Sha, L. (2001). Using simplicity to control complexity. IEEE Software, 18(4), 20-28. https://doi.org/10.1109/MS.2001.936213

Shanahan, M. (2006). A cognitive architecture that combines internal simulation with a global workspace. Consciousness and Cognition, 15(2), 433-449. https://www.sciencedirect.com/science/article/abs/pii/S1053810005001510

Skenderi, G., Li, H., Tang, J., & Cristani, M. (2025). Graph-level representation learning with joint-embedding predictive architectures. Transactions on Machine Learning Research. https://openreview.net/forum?id=v47f4DwYZb

Slotnick, J., Khodadoust, A., Alonso, J., Darmofal, D., Gropp, W., Lurie, E., & Mavriplis, D. (2014). CFD Vision 2030 study: A path to revolutionary computational aerosciences (NASA/CR-2014-218178). National Aeronautics and Space Administration.

Smeresky, B., Abell, P., Fries, M., & Hankey, M. (2021). Bolide fragment detection in Doppler weather radar data using artificial intelligence/machine learning. Meteoritics & Planetary Science, 56(8), 1585-1596. https://doi.org/10.1111/maps.13718

Stauff, N. E., Miao, Y., Cao, Y., Mo, K., Abdelhameed, A. A. E., Ibarra, L., Matthews, C., & Shemon, E. R. (2024). High-fidelity multiphysics modeling of a heat pipe microreactor using BlueCrab. Nuclear Science and Engineering. Advance online publication. https://www.osti.gov/pages/biblio/2569789

Stevenson, E., Rodriguez-Fernandez, V., Urrutxua, H., & Camacho, D. (2023). Benchmarking deep learning approaches for all-vs-all conjunction screening. Advances in Space Research, 72(7), 2660-2675. https://www.sciencedirect.com/science/article/pii/S0273117723000601

Studt, M., & Schildbach, G. (2025). Hierarchical reinforcement learning with low-level MPC for multi-agent control. arXiv:2509.15799. https://arxiv.org/abs/2509.15799

Tabassi, E. (2023). Artificial intelligence risk management framework (AI RMF 1.0) (NIST AI 100-1). National Institute of Standards and Technology. https://nvlpubs.nist.gov/nistpubs/ai/NIST.AI.100-1.pdf

Terver, B., Yang, T.-Y., Ponce, J., Bardes, A., & LeCun, Y. (2026). What drives success in physical planning with joint-embedding predictive world models? Transactions on Machine Learning Research. https://openreview.net/forum?id=cHZn5Gdh8e (arXiv:2512.24497v3)

Towner, M. C., Jansen-Sturgeon, T., Cupak, M., Sansom, E. K., Devillepoix, H. A. R., Bland, P. A., Howie, R. M., Paxman, J. P., Benedix, G. K., & Hartig, B. A. D. (2022). Dark-flight estimates of meteorite fall positions: Issues and a case study using the Murrili meteorite fall. The Planetary Science Journal, 3(2), Article 44. https://doi.org/10.3847/PSJ/ac3df5

Tramm, J. R., Romano, P. K., Shriwise, P., Lund, A., Doerfert, J., Steinbrecher, P., Siegel, A., & Ridley, G. (2024). Performance portable Monte Carlo particle transport on Intel, NVIDIA, and AMD GPUs. EPJ Web of Conferences, 302, 04010. https://doi.org/10.1051/epjconf/202430204010

U.S. Nuclear Regulatory Commission. (2024). Regulatory framework gap assessment for the use of artificial intelligence in nuclear applications (ADAMS Accession No. ML24290A059). https://www.nrc.gov/docs/ML2429/ML24290A059.pdf

U.S. Nuclear Regulatory Commission. (2026). Licensing requirements for microreactors and other reactors with comparable risk profiles. Federal Register, 91, 23628-23766. https://www.federalregister.gov/documents/2026/05/01/2026-08550/licensing-requirements-for-microreactors-and-other-reactors-with-comparable-risk-profiles

U.S. Space Command. (2026a). Conjunction Data Messages (CDMs) [Data set]. Space-Track.org. Retrieved May 31, 2026, from https://www.space-track.org/

U.S. Space Command. (2026b). General perturbations element sets [Data set]. Space-Track.org. https://www.space-track.org/

Vallado, D. A., Crawford, P., Hujsak, R., & Kelso, T. S. (2006). Revisiting Spacetrack Report #3. In AIAA/AAS Astrodynamics Specialist Conference and Exhibit (Paper AIAA 2006-6753). https://celestrak.org/publications/AIAA/2006-6753/

van der Maaten, L., & Hinton, G. (2008). Visualizing data using t-SNE. Journal of Machine Learning Research, 9, 2579-2605.

Vaswani, A., Shazeer, N., Parmar, N., Uszkoreit, J., Jones, L., Gomez, A. N., Kaiser, L., & Polosukhin, I. (2017). Attention is all you need. In Advances in Neural Information Processing Systems 30 (pp. 5998-6008). https://papers.nips.cc/paper_files/paper/2017/hash/3f5ee243547dee91fbd053c1c4a845aa-Abstract.html

Virtanen, P., Gommers, R., Oliphant, T. E., Haberland, M., Reddy, T., Cournapeau, D., Burovski, E., Peterson, P., Weckesser, W., Bright, J., van der Walt, S. J., Brett, M., Wilson, J., Millman, K. J., Mayorov, N., Nelson, A. R. J., Jones, E., Kern, R., Larson, E., . . . SciPy 1.0 Contributors. (2020). SciPy 1.0: Fundamental algorithms for scientific computing in Python. Nature Methods, 17(3), 261-272. https://doi.org/10.1038/s41592-019-0686-2

Weller, H. G., Tabor, G., Jasak, H., & Fureby, C. (1998). A tensorial approach to computational continuum mechanics using object-oriented techniques. Computers in Physics, 12(6), 620-631. https://pubs.aip.org/aip/cip/article/12/6/620/510187/A-tensorial-approach-to-computational-continuum

Wijmans, E., Kadian, A., Morcos, A., Lee, S., Essa, I., Parikh, D., Savva, M., & Batra, D. (2020). DD-PPO: Learning near-perfect PointGoal navigators from 2.5 billion frames. In International Conference on Learning Representations. https://arxiv.org/abs/1911.00357

Williams, G., Aldrich, A., & Theodorou, E. A. (2017). Model predictive path integral control: From theory to parallel computation. Journal of Guidance, Control, and Dynamics, 40(2), 344-357. https://doi.org/10.2514/1.G001921

Williamson, T., Bi, S., Vasile, M., & Fossati, M. (2025). Gaussian mixture-unscented transform propagation during re-entry fragmentation for impact uncertainty quantification (Paper 221). In S. Lemmens, T. Flohrer, & F. Schmitz (Eds.), Proceedings of the 9th European Conference on Space Debris. ESA Space Debris Office. https://conference.sdo.esoc.esa.int/proceedings/sdc9/paper/221

Wissink, A., Cary, A., Georgiadis, N., Smith, B., Wyman, N., Wang, Z. J., Gropp, B., Duraisamy, K., & Wukie, N. (2026). CFD Vision 2030 Roadmap: Progress and projections in 2026. In AIAA AVIATION

Forum (AIAA Paper 2026-4020). American Institute of Aeronautics and Astronautics. https://doi.org/10.2514/6.2026-4020

Wright, E., Boley, A., & Byers, M. (2025). Airspace closures due to reentering space objects. Scientific Reports, 15(1), Article 2966. https://doi.org/10.1038/s41598-024-84001-2

Xiang, J., Lv, Z., Xu, S., Deng, Y., Wang, R., Zhang, B., Chen, D., Tong, X., & Yang, J. (2025). Structured 3D latents for scalable and versatile 3D generation. In Proceedings of the IEEE/CVF Conference on Computer Vision and Pattern Recognition (pp. 21469-21480). https://arxiv.org/abs/2412.01506

Xie, T., Zhang, D., Chen, J., Li, X., Zhao, S., Cao, R., Hua, T. J., Cheng, Z., Shin, D., Lei, F., Liu, Y., Xu, Y., Zhou, S., Savarese, S., Xiong, C., Zhong, V., & Yu, T. (2024). OSWorld: Benchmarking multimodal agents for open-ended tasks in real computer environments. In Advances in Neural Information Processing Systems 37 (Datasets and Benchmarks Track). https://arxiv.org/abs/2404.07972

Yao, S., Zhao, J., Yu, D., Du, N., Shafran, I., Narasimhan, K., & Cao, Y. (2023). ReAct: Synergizing reasoning and acting in language models. In International Conference on Learning Representations. https://arxiv.org/abs/2210.03629